\documentclass[twoside]{article}
\usepackage[accepted]{aistats2023}

\usepackage{amsmath,amsfonts,bm}

\def\eqref#1{equation~\ref{#1}}

\def\1{\bm{1}}

\DeclareMathAlphabet{\mathsfit}{\encodingdefault}{\sfdefault}{m}{sl}
\SetMathAlphabet{\mathsfit}{bold}{\encodingdefault}{\sfdefault}{bx}{n}

\usepackage{hyperref}
\usepackage{url}
\usepackage{booktabs}
\usepackage{pifont}
\usepackage{xcolor}
\usepackage{adjustbox}
\usepackage{caption}
\usepackage{amsfonts}
\usepackage{transparent}    
\usepackage{pifont}
\usepackage{multirow}
\usepackage{makecell}
\usepackage{mathtools}
\usepackage{algorithm}
\usepackage{algpseudocode}

\newcommand{\cmark}{\ding{51}}

\newcommand{\xmark}{\transparent{0.4} \ding{55}}%
\newcommand{\quotes}[1]{``#1''}

\usepackage[round]{natbib}

\setcitestyle{authoryear,round,citesep={;},aysep={,},yysep={;}}

\begin{document}

\runningauthor{Samuel Holt, Alihan Hüyük, Zhaozhi Qian, Hao Sun, Mihaela van der Schaar}

\twocolumn[
\aistatstitle{Neural Laplace Control for Continuous-time Delayed Systems}

\aistatsauthor{ 
Samuel Holt \\ University of Cambridge \\ \href{mailto:sih31@cam.ac.uk}{sih31@cam.ac.uk}
\And 
Alihan Hüyük \\ University of Cambridge \\ \href{mailto:ah2075@cam.ac.uk}{ah2075@cam.ac.uk}
\And
Zhaozhi Qian \\ University of Cambridge \\ \href{mailto:zq224@maths.cam.ac.uk}{zq224@maths.cam.ac.uk}}
\vspace{0.1in}
\aistatsauthor{
Hao Sun \\ University of Cambridge \\ \href{mailto:hs789@cam.ac.uk}{hs789@cam.ac.uk}
\And
Mihaela van der Schaar \\ University of Cambridge \& The Alan Turing Institute \\ \href{mailto:mv472@cam.ac.uk}{mv472@cam.ac.uk}}
\vspace{0.20in}]

\begin{abstract}
Many real-world offline reinforcement learning (RL) problems involve continuous-time environments with delays.
Such environments are characterized by two distinctive features: firstly, the state $x(t)$ is observed at irregular time intervals, and secondly, the current action $a(t)$ only affects the future state $x(t + \tau)$ with an \textit{unknown} delay $\tau > 0$.
A prime example of such an environment is satellite control where the communication link between earth and a satellite causes irregular observations and delays. 
Existing offline RL algorithms have achieved success in environments with irregularly observed states in time or known delays.
However, environments involving both irregular observations in time and unknown delays remains an open and challenging problem.
To this end, we propose \textit{Neural Laplace Control}, a continuous-time model-based offline RL method that combines a Neural Laplace dynamics model with a model predictive control (MPC) planner---and is able to learn from an offline dataset sampled with irregular time intervals from an environment that has a inherent unknown constant delay.
We show experimentally on continuous-time delayed environments it is able to achieve near expert policy performance.
\end{abstract}

\section{INTRODUCTION}

Online Reinforcement learning methods struggle to be applied to many real-world environments, for example in healthcare, business, and autonomous driving environments. RL methods rely on costly trial-and-error approaches performed either online or in a realistic simulator of the environment---which are not often readily available.
In contrast, \quotes{offline} model-based reinforcement learning learns the environment dynamics, from a previously collected dataset of state-action trajectories, which is often readily available.
It then controls the system to a desired goal using any suitable planning method, such as training a policy \citep{fujimoto2018addressing} or using a model predictive controller (MPC) \citep{williams2016aggressive}.

In practice, real-world environments are continuous in time by nature and possess constant delays $\tau$ whereby either actions are not executed instantaneously $\bm{a}(t+\tau)$, or states are not observed instantaneously $\bm{x}(t+\tau)$ (we formally define these later, in Section \ref{problemformulation}).
For instance, in healthcare, observing a treatment effect $\bm{a}(t)$ from giving a patient a medication is not observable instantaneously and is instead \textit{delayed} $\bm{x}(t+\tau)$, whilst measurements may be measured at \textit{irregular} time intervals, $\bm{x}(t + \Delta_i)$, $\Delta_i \neq \Delta_j$---as is common where the frequency of observations is indicative of the patient's medical status \citep{goldberger2000physiobank}.
Similarly in autonomous driving it can take more than $\tau=0.4$s for a hydraulic automotive braking  system to generate a desired deceleration, therefore accounting for the delayed environment dynamics is \textit{crucial} for correct safe control of the vehicle \citep{bayan2010brake}.
All together these environment dynamics, can often be described through sets of \textit{delay differential equations} (DDEs) \citep{lynch2017modern}, however are often \textit{unknown}.

Prior work has shown the success of model-based RL to learn from offline datasets consisting separately of either (1) irregularly-sampled data with no environment delays $\Delta_i \neq \Delta_j, \tau=0$ \citep{yildiz2021continuous} with \textbf{continuous-time methods}, or (2) regularly-sampled data with environment delays $\Delta_i = \Delta_j, \tau > 0$ \citep{chen2021delay} with \textbf{discrete-time delay methods}.
However, performing offline model-based RL with both delays $\tau > 0$ and irregularly-sampled data $\Delta_i \neq \Delta_j$ for continuous-time control tasks is a largely understudied problem, yet an important problem setting.

The existing two individual approaches are inherently incompatible with each other. On one hand, \textbf{continuous-time methods} use a continuous-time model, such as neural \textit{ordinary differential equation} (ODE), to  learn from irregularly-sampled observations $\Delta_i \neq \Delta_j, \tau=0$  \citep{yildiz2021continuous, du2020model}. 
However, ODEs cannot model environments with unknown delays $\tau > 0$ \citep{holt2022neural}. 
Furthermore, neural-ODE based models suffer from poor computational efficiency, when scaling to longer time horizons in a given trajectory (as shown in Section \ref{insights}).

On the other hand, the existing methods for handling delays only considers a discrete-time $\Delta_i = \Delta_j, \tau > 0$ environment. 
Where these methods assume the delay $\tau$ is either \textit{known} a priori \citep{firoiu2018human} or is implicitly learned from a discrete buffer of previously executed actions $\bm{\bar{a}}_{i-1}=\{\bm{a}_1,\ldots,\bm{a}_{i-1}\}$ (or states) \citep{chen2021delay}.
A simple approach for handling environments with delays $\tau > 0$ is to greatly increase the time interval between actions performed, so as to synchronize the agent's actions with its delayed observations.
However, such an approach would lead to a \quotes{waiting agent} that is clearly sub optimal in most environments.

Hence, the following two properties are highly desirable for offline model-based RL to perform in more real-world environments. \\
\textbf{(P1) Learn from irregular samples}: able to learn from irregularly-sampled $\Delta_i \neq \Delta_j$ in time offline datasets resulting from a continuous-time environment. \\
\textbf{(P2) Learn delayed dynamics}: can learn the delayed dynamics of the environment, implicitly modeling any \textit{unknown} delays $\tau > 0$. 

To fulfill P1 and P2, we propose \textbf{Neural Laplace Control} (NLC), a continuous-time model-based RL method. 
Rather than describing the environment dynamics with a (neural) ODE, it uses Neural Laplace to learn implicit \textit{delay differential equation} dynamics, which simultaneously accounts for unknown delays $\tau > 0$ and continuous-time dynamics $\Delta_i \neq \Delta_j$. 
This brings two immediate advantages.
First, many continuous-time control problems involving delay DEs can easily be represented and solved in the Laplace domain \citep{schiff1999laplace, aastrom2021feedback, yi2008controllability}.
Secondly, Neural Laplace Control bypasses the standard numerical ODE solvers and uses an inverse Laplace transform algorithm \citep{holt2022neural} to reconstruct any future state $\bm{x}(t + \Delta_i)$ of the dynamics model with the same amount of compute.
This makes employing more principled planning strategies such as MPC feasible for continuous-time domains over expensive numerical step wise ODE based models.

Specifically, Neural Laplace Control is able to tackle the novel continuous-control problem formulated of having both \textit{states observed at irregular time intervals $\Delta_i \neq \Delta_j$} and an \textit{unknown fixed delay $\tau > 0$ in the environment}. We motivate Neural Laplace Control as a principled approach for this problem and demonstrate the empirical effectiveness in experiments.

\paragraph{Contributions}
Our contributions are two-fold:
\textbf{\textcircled{\raisebox{-0.9pt}{1}}} In Section \ref{methodsection}, we formulate and motivate the novel Neural Laplace Control method, that can learn a dynamics model that can encode an environments unknown delay differential equation dynamics from irregularly-sampled in time state-action trajectories (P1,P2).
\textbf{\textcircled{\raisebox{-0.9pt}{2}}} In section \ref{main_results}, we benchmark Neural Laplace Control against the existing continuous-time model-based approaches on standard continuous-time delayed environments.
Specifically, we demonstrate that Neural Laplace Control is able to achieve near expert policy performance, significantly achieving a higher episode reward than the other competing continuous-time model-based baseline methods.
We also gain insight and understanding of how Neural Laplace Control works in Section \ref{insights}, of how it can correctly extrapolate to longer time horizons for the dynamics model and is computationally more efficient for predicting the next state at a longer time horizon, thereby making model predictive control feasible for longer time horizons for a fixed compute budget. All together, we learn such a model from irregularly-sampled states and actions in time $\Delta_i \neq \Delta_j$ (P1) and environments that possess a delay that is unknown $\tau>0$ (P2).

A PyTorch \citep{NEURIPS2019_9015} implementation of the code is at \href{https://github.com/samholt/NeuralLaplaceControl}{https://github.com/samholt/NeuralLaplaceControl}, and have a broader research group codebase at \href{https://github.com/vanderschaarlab/NeuralLaplaceControl}{https://github.com/vanderschaarlab/NeuralLaplaceControl}.
\section{RELATED WORK}
\begin{table*}
    \centering
    \caption{Comparison with related model-based approaches to RL. \textbf{(P1) Learn from irregular samples}---can it learn from an offline dataset sampled at irregular times, $\Delta_i \neq \Delta_j$? \textbf{(P2) Learn delayed dynamics}---can it learn environments that contain a delay $\tau > 0$? Neural Laplace Control is the only method that can both learn from irregular samples (P1) as well as learn environments that contain a delay (P2).}
    \label{table:main_related_work_main}
    \small
    \resizebox{\linewidth}{!}{%
    \begin{tabular}{@{}lcccccc@{}}
        \toprule
        \bf Approach & \bf True Dynamics & \bf Data Available & \bf Reference \ & \bf Model & \bf (P1) $\Delta_i \neq \Delta_j$ & \bf (P2) $\tau>0$ \\
        \midrule
        Conventional model-based RL & $\bm{x}_{t+1}\sim f(\bm{x}_t,\bm{a}_t)$ & $\mathcal{D}=\{(\bm{x}_i,\bm{a}_i)\}_{i=0}^n$ & \citet{williams2017information} & MDP~/~Neural Network & \xmark & \xmark \\
        Discrete-time delay methods & $\bm{x}_{t+1}\sim f(\bm{x}_t,\bm{a}_{t-\tau})$ & $\mathcal{D}=\{(\bm{x}_i,\bm{a}_i)\}_{i=0}^n$ & \citet{chen2021delay} & DA-MDP / RNN & \xmark & \cmark \\
        \multirow{2}{*}{Continuous-time methods} & \multirow{2}{*}{$\bm{\dot{x}}(t)=f(\bm{x}(t),\bm{a}(t))$} & \multirow{2}{*}[-1.2pt]{\makecell{$\mathcal{D}=\{(\bm{x}(t_i),\bm{a}(t_i))\}_{i=0}^n$\\[-2.4pt]\scriptsize s.t. $\exists i,\!j: t_{i+1}\!-\!t_i\neq t_{j+1}\!-\!t_j$}} & \citet{yildiz2021continuous} & Neural ODE & \cmark & \xmark \\
        & & & \citet{du2020model} & Latent ODE & \cmark & \xmark \\
        \midrule
        \bf Neural Laplace Control & $\bm{\dot{x}}(t)=f(\bm{x}(t),\bm{a}(t-\tau))$ & \makecell{$\mathcal{D}=\{(\bm{x}(t_i),\bm{a}(t_i))\}_{i=0}^n$\\[-2.4pt]\scriptsize s.t. $\exists i,\!j: t_{i+1}\!-\!t_i\neq t_{j+1}\!-\!t_j$} & \bf (Ours) & Neural Laplace Control & \cmark & \cmark \\
        \bottomrule
    \end{tabular}}
\end{table*}

In offline reinforcement learning, an agent learns from a fixed replay buffer and is not permitted to interact with the environment \citep{wu2019behavior}. While both model-free \citep{kumar2019stabilizing,kumar2020conservative,fujimoto2021minimalist} and model-based \citep{kidambi2020morel,wang2021offline} approaches have been proposed for offline RL, in general, model-based methods have been shown to be more sample efficient than model-free methods \citep{moerland2020model}. The main challenge in model-based RL is known as \quotes{extrapolation error} \citep{fujimoto2019off}, whereby the learnt dynamics model inaccuracies compound for a larger number of future predicted time steps. Hence, it is crucial in model-based RL to learn an appropriate dynamics model that is capable of accurately capturing the unique characteristics of an environment. However, despite the fact that many environments operate in continuous-time by nature and contain action or observation delays, almost all of the existing approaches to model-based RL consider dynamics models only suited to the conventional discrete-time $\Delta_i = \Delta_j$ setting with no delays $\tau = 0$. We review here some of the few approaches that go beyond the conventional setting, namely (i) \textit{discrete-time delay methods} and (ii) \textit{continuous-time methods}.

\paragraph{Discrete-time Delay Methods}
Modeling environments with either delayed observations $\bm{x}(t+\tau)$ or delayed actions $\bm{a}(t+\tau)$ are equivalent in form \citep{katsikopoulos2003markov}. Prior work models regular sampled $\Delta_i=\Delta_j$ (discrete time) environments with constant time delays $\tau>0$, and provides the agent with the current state $\bm{x}(t)$, and a history of past actions performed in the environment $\bm{\bar{a}}_{i-1}=\{\bm{a}_1,\ldots,\bm{a}_{i-1}\}$, whereby the history action window is larger than or equal to the observation or action delay in the environment \citep{walsh2009learning,firoiu2018human,bouteiller2020reinforcement, liotet2021learning, agarwal2021blind}.
Recently, \citet{chen2021delay} proposed delay-aware Markov decision processes (MDPs) that are capable of modeling delayed dynamics in discrete-time based on regularly sampled data, with an RNN encoding the history of past actions and the current state. Moreover, \citet{derman2020acting} proposes a discrete-time \textit{known} delay method---whereby, \citet{derman2020acting}'s App. D.2 benchmarks against a model-free A2C baseline that only uses the current state-action fed into a RNN, that is \textit{unable} to learn the delay.

\paragraph{Continuous-time Methods}
Real world data is often sampled irregularly $\Delta_i \neq \Delta_j$, as such \citep{yildiz2021continuous} propose to use Neural ODEs \citep{chen2018neural} as their continuous-time dynamics model that can model irregularly-sampled environments with no delays $\tau=0$. Similarly, the work of \citet{du2020model} uses a Latent ODE model when planning policies.
Moreover, the work of \citet{seedat2022continuous} uses a controlled differential equation \citep{kidger2020neural} to model counterfactual outcomes. 
However, these existing approaches are limiting, as an ODE-based model by definition cannot handle a delay differential equation, necessitating the need for a model that can learn and model more diverse classes of differential equations. Recent models, of modeling diverse classes of differential equations is made possible with the work of Neural Laplace \citep{holt2022neural} by representing them in the Laplace domain. These Laplace-based models have been shown to be able to model such systems, be more accurate and scale better with increasing time horizons in time complexity. Our approach, namely Neural Laplace Control, essentially extends Neural Laplace to the setting of controlled systems---that is systems that evolve based on an action signal $\bm{a}(t)$---so that it can be used in planning policies in a RL setting.
Furthermore, \citet{bruder2007impulse} provides theory for the continuous time specific setting where action is an impulse, rather than a multivariate continuous input and the environment dynamics are a diffusion process.

We summarize the key related work in Table \ref{table:main_related_work_main} and provide an extended discussion of additional related works, including a review of the benefits of using model-based RL and using model predictive control, which happens to be our preferred strategy for planning policies, in Appendix \ref{extendedrelatedwork}.
\section{PROBLEM FORMULATION}
\label{problemformulation}

\paragraph{States \& Actions}
For a system with \textit{state} space $\mathcal{X}=\mathbb{R}^{d_\mathcal{X}}$ and \textit{action} space $\mathcal{A}=\mathbb{R}^{d_\mathcal{A}}$, the state at time $t\in\mathbb{R}$ is denoted as $\bm{x}(t)=[x_1(t),\ldots,x_{d_\mathcal{X}}(t)]\in\mathcal{X}$ and the action at time $t\in\mathbb{R}$ is denoted as \mbox{$\bm{a}(t)=[a_1(t),\ldots,a_{d_\mathcal{A}}(t)]\in\mathcal{A}$}. We elaborate that \textit{state trajectory} $\bm{x}:\mathbb{R}\to\mathcal{X}$ and \textit{action trajectory} $\bm{a}:\mathbb{R}\to\mathcal{A}$ are both functions of time, where an individual state $\bm{x}(t)\in\mathcal{X}$ or an individual action~$\bm{a}(t)\in\mathcal{A}$ are points on these trajectories. Given a time interval~$\mathcal{I}\subseteq \mathbb{R}$, $\bm{x}_{\mathcal{I}}\in\mathcal{X}^{\mathcal{I}}$ and $\bm{a}_{\mathcal{I}}\in\mathcal{A}^{\mathcal{I}}$ we denote the partial state and action trajectories on that interval such that $\bm{x}_{\mathcal{I}}(t)=\bm{x}(t)$ and $\bm{a}_{\mathcal{I}}(t)=\bm{a}(t)$ for $t\in\mathcal{I}$. Finally, we also note that action values are usually bounded by an actuator's limits hence we also restrict the action space to $\mathcal{A}=[\bm{a}_{\min}, \bm{a}_{\max}]$, i.e., a box in Euclidean space.

\begin{figure*}[!t]
  \centering
\includegraphics[width=\textwidth]{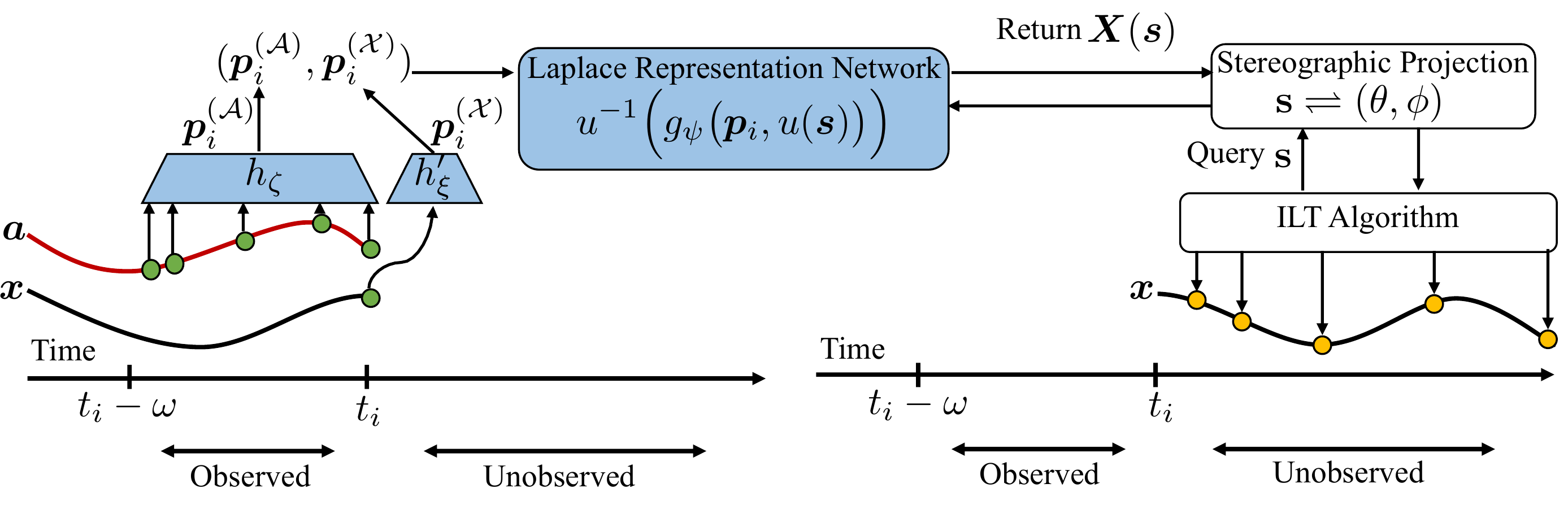}
\caption{Block diagram of Neural Laplace Control. The query points $\bm{s}$ are given by the inverse Laplace transform (ILT) algorithm based on the time points of the future state trajectory to predict. The gradients can be back-propagated through the ILT algorithm and stereographic projection to train networks $h_\zeta, h'_\xi, g_{\psi}$.}
\label{Fig:blockdiag}
\end{figure*}

\paragraph{Environment Dynamics}
Dynamics of the system are described by a non-autonomous non-linear controlled delay \textit{differential equation} with a constant action delay $\tau\in\mathbb{R}_+$:
\begin{align}
    \bm{\dot{x}}(t) = \frac{d\bm{x}(t)}{dt} = f(\bm{x}(t),\bm{a}(t-\tau))
    \label{maindedynamics}
\end{align}
where function~$f:\mathcal{X}\times\mathcal{A}\to\mathbb{R}^{d_\mathcal{X}}$ maps the current state and delayed action pair $\bm{x}(t),\bm{a}(t-\tau)$ to a state derivative~$\bm{\dot{x}}(t)$.
We note that this setting of \textit{Continuous-time Control} \citep{kwakernaak1972linear, aastrom2021feedback}, assumes deterministic environments with no observation noise---however, also consider observation noise in Appendix \ref{additonalexperiments}, and find NLC still performant.
Furthermore, we note that an environment that has either a constant action delay or a constant observation state delay are both equivalent and refer to a constant action delay throughout \citep{katsikopoulos2003markov}.
This is evident as an agent that interacts in an environment that \textit{only} has an action delay of $\tau$ will only observe the effect of its selected action $a(t)$ $\tau$ seconds later in the state observation---therefore the agent has to decide an action based on a state observation that is delayed by $\tau$.
Therefore, given an initial state $\bm{x}(0)$ as well as an action trajectory $\bm{a}_{(-\tau,t-\tau)}$, the state at time $t\in(0,\infty)$ can be written as
\begin{align}
    \bm{x}(t) = \bm{x}(0) + \int_0^t f(\bm{x}(t'), \bm{a}(t'-\tau)) dt'
\end{align}
Consider some policy $\pi:\cup_{t\in\mathbb{R}}\mathcal{X}^{(-\infty,t]}\to\mathcal{A}$ that controls the system by mapping state trajectories $\bm{x}_{(-\infty,t]}$ to actions $a(t)=\pi(\bm{x}_{(-\infty,t]})$. Then similarly, given an initial state trajectory~$\bm{x}_{(-\infty,0]}$, the state at time $t\in(0,\infty)$ for when the system is controlled by policy $\pi$ can be written as $\bm{x}^{(\pi)}(t)=\bm{x}(0)+\int_0^tf(\bm{x}(t'), \pi(\bm{x}_{(-\infty,t'-\tau]})) dt'$.

\paragraph{Offline Dataset}
We consider the case where the environment dynamics $f$ including the action delay $\tau$ are unknown. Instead, we observe irregularly-sampled in time state-action trajectories: $\mathcal{D}=\{(\bm{x}^{(k)}(t_i^{(k)}),\bm{a}^{(k)}(t_i^{(k)}))\}_{i=0}^{n^{(j)}}$ where $\bm{x}^{(k)},\bm{a}^{(k)}$ denotes the $k$-th state-action trajectory sampled at irregular times $\{t_0^{(k)},t_1^{(k)},\ldots,t_{n^{(k)}}^{(k)}\}$; we drop the trajectory index~$k$ unless explicitly needed. Letting $\Delta_i=t_i-t_{i-1}$ denote the time interval between two consecutive samples, having irregular samples entails that $\Delta_i\neq\Delta_j$ for some $i,j\in\{1,\ldots,n\}$. We also denote with $\bm{x}_i=\bm{x}(t_i)$ as the $i$-th state observation and with $\bm{a}_i=\bm{a}(t_i)$ as the $i$-th action observation.

\paragraph{Control Objective}
Our overall objective is to control the environment to a given goal state $\bm{x}^*\in\mathcal{X}$. This is achieved by defining an instantaneous reward function $r:\mathcal{X}\to\mathbb{R}$ of the current state. A common reward function in continuous-time control is the exponential of the negative distance from the current state to the goal state, that is $r(\bm{x}(t))=e^{-||\bm{x}(t) - \bm{x}^*||_2}$---so that the instantaneous reward is maximized when $\bm{x}(t)=\bm{x}^*$. Consequently, to achieve our objective, we seek to find the \textit{optimal policy} $\pi^*$ within a feasible set of policies~$\Pi$ that maximizes the reward integral for a given final time $T\in\mathbb{R}_+$:
\begin{align}
    \pi^* &= \arg\!\max{}_{\pi\in\Pi}\bigg( R^{(\pi)}\doteq\int_0^{T} r(\bm{x}^{(\pi)}(t'))dt' \bigg)
\end{align}
given the offline dataset~$\mathcal{D}$ but without access to the environment dynamics~$f$ or the action delay~$\tau$.

In practice, policies cannot observe continuous state trajectories in their entirety and hence have to work with point-wise samples instead. As such, we restrict our search space~$\Pi$ to practical policies that only update their actions whenever a state observation is made. Given sampling times $\{t_0,t_1,\ldots\}$, these are policies of the form 
\begin{align}
    \pi(\bm{x}_{(-\infty,t)}) = \begin{dcases}
        \bm{0} & \text{if}~ t\in(-\infty,t_0) \\
        \bm{a}_i & \text{if}~ t\in[t_i,t_{i+1}) \\
    \end{dcases}
\end{align}
where actions $\bm{a}_i$ are generated recursively given $\bm{x}_{(-\infty,t_i]}$ by an auxiliary policy $\pi':\mathcal{X}\times\cup_{j\in\mathbb{Z}_+}\mathcal{A}^j\to\mathcal{A}$ such that $\bm{a}_i=\pi'(\bm{x}_i,\bm{\bar{a}}_{i-1}=\{\bm{a}_1,\ldots,\bm{a}_{i-1}\})$. Note that it is sufficient for the auxiliary policy to keep track of only the most recent state observation $\bm{x}_i$ due to the Markovianity of environment dynamics $f$, but it needs to keep track of all the past actions due to the (unknown) action delay~$\tau$---keeping consistent with our convention of modeling action delays, however, note that this is equivalent for state delays.
However, in the case of a having a fixed previous time window $\omega \in \mathbb{R}_+$, the implicit learnable delay $\tau$ is bounded by $\omega$, i.e., $\tau < \omega$.

\section{NEURAL LAPLACE CONTROL}
\label{methodsection}
We follow the standard model-based framework setup \citep{lutter2021learning}. First, we learn a dynamics model as outlined in Section~\ref{sec:learning}. Then in Section~\ref{sec:planning}, we use this learnt model to plan a policy via model predictive control.
\subsection{Learning the Dynamics Model}
\label{sec:learning}
In the following we propose a way to incorporate \textit{actions} into the Neural Laplace \citep{holt2022neural} model for modeling diverse DE systems, also detailed in Appendix \ref{problemandbackgroundappendix}. We build on the Neural Laplace framework, which was originally designed to only model the state differential equation evolution. Specifically, Neural Laplace Control involves three main components: \textbf{(1)} an encoder that learns to infer and represent the initial representation of the current state-action trajectory up to time $t$, \textbf{(2)} a Laplace representation network that learns to represent the solutions of the state trajectory in the Laplace domain conditioned on the input state-action trajectories, and \textbf{(3)} an inverse Laplace transform (ILT) algorithm that converts the Laplace representation back to the time domain. We also note that Neural Laplace Control is preferable for non-linear dynamics as nonlinear delay DE's can be solved in the Laplace domain, through the Laplace Adomian decomposition method \citep{yousef2018application}.
We provide a block diagram in Figure \ref{Fig:blockdiag} and now discuss each component in detail.

\paragraph{(1) Learning to Represent Initial Conditions at Time $t$}
The future state trajectory solution depends on the \textit{initial condition} of the state-action trajectories.
To model the delay differential equation environment dynamics fully, we seek to encode this initial condition, whereby the dynamics implicitly depend on the \textit{past} state-action histories. Therefore, Neural Laplace Control uses an encoder network to learn a \textit{representation} of the current initial condition at time $t$ by encoding the recent state-action history of the trajectory up to a fixed previous time window $\omega\in \mathbb{R}+$. Note that ideally, we want $\omega>\tau$ since previous actions at least up to a time window of $\tau$ affect how states evolve in the future.

For a sample observed at time $t_i$, instead of encoding both state and action histories, we note that we only need to encode one history and follow the convention of \citet{walsh2009learning} to only encode the current state $\bm{x}_i=\bm{x}(t_i)$, and the action history $\mathcal{H}_i=\{(\bm{a}_j,t_j-t_i):t_j\in[t_i-\omega,t_i]\}$ up to the previous time window~$\omega$. We highlight that the actions encoded can be at irregular times.
As the action history varies in time, we encode it with a recurrent neural network, that of a reverse time gated recurrent unit \citep{DBLP:journals/corr/abs-1806-07366, holt2022neural}---denoted as $h_{\zeta}$ with parameters $\zeta$---and encode the current state with a linear neural network layer---denoted as $h'_{\xi}$ with parameters $\xi$---and concatenate both into a latent dimension representing the initial condition of the state-action trajectory:
\begin{align}
    \bm{p}_i=(\bm{p}_i^{(\mathcal{A})} \doteq h_\zeta(\mathcal{H}_i),~ \bm{p}_i^{(\mathcal{X})} \doteq h'_\xi(\bm{x}_i)) \label{encoders_eq}
\end{align}
The vector $\bm{p}_i \in\mathcal{P}=\mathbb{R}^{d_{\mathcal{P}}}$ is the learned initial condition representation, where $d_{\mathcal{P}} \ge d_{\mathcal{X}}$ is a hyper-parameter. The encoders $h_\zeta, h'_\xi$ have trainable weights $\zeta,\xi$ respectively. Neural Laplace Control is agnostic to the exact choice of encoder architecture.

\paragraph{(2) Learning DE Solutions in the Laplace Domain} 
Given an initial condition representation $\bm{p}\in\mathcal{P}$, we need to learn a function ${l} : \mathcal{P}\times \mathbb{C}^{d_{\mathcal{S}}} \to \mathbb{C}^{d_{\mathcal{X}}}$ that models the Laplace representation of the delay DE solution, i.e., when we take the inverse Laplace transform (ILT) of $\bm{X}(\bm{s}) = l(\bm{p}, \bm{s})$, it approximates $\bm{x}(t)$ well for future $t$.
Here, $d_{\mathcal{S}} \in \mathbb{N}_+$ denotes the number of reconstruction terms per time point and is specific to the ILT algorithm (Appendix \ref{problemandbackgroundappendix}).
However, the Laplace representation $\bm{X}(\bm{s})$ often involves singularities \citep{schiff1999laplace}, which are difficult for neural networks to approximate or represent \citep{DBLP:journals/rc/BakerP98}. Therefore, we use the proposed stereographic projection onto a Riemann sphere to mitigate this \citep{holt2022neural}.
With the stereographic projection, we introduce a feed-forward neural network $g$ to learn the Laplace representation of the dynamics model solution:
\begin{equation}
\label{fs_h}
    \bm{X}(\bm{s}) =  u^{-1} \Big(g_{\psi} \big(\bm{p}, u(\bm{s})\big) \Big),
\end{equation}
where $u$ is the stereographic projection and $u^{-1}$ is the inverse stereographic projection (Appendix \ref{problemandbackgroundappendix}), the vector $\bm{p}$ is the output of the encoders (Equation \ref{encoders_eq}), and $\psi$ is the trainable weights.
Here the neural network's inputs and outputs are the coordinates on the Riemann Sphere $(\theta, \phi) \in \mathcal{D} = (-{\pi}, {\pi}) \times (-\frac{\pi}{2}, \frac{\pi}{2})$, which are bounded and free from singularities \citep{holt2022neural}.

\paragraph{(3) Inverse Laplace transform}
After obtaining the Laplace representation $\bm{X}(\bm{s})$, we compute the predicted or reconstructed state values $\bm{\hat{x}}(t)=\mathcal{L}^{-1}\{\bm{X}\}(t)$ for future $t$, where $\mathcal{L}^{-1}$ is the inverse Laplace transform, using a numerical inverse Laplace transform algorithm.
We highlight that we can evaluate $\hat{\bm{x}}(t)$ \textit{at any future time} $t\in \mathbb{R}_+$ as the Laplace representation is independent of time once learnt.
In practice, we use the well-known ILT Fourier series inverse algorithm (ILT-FSI), which can obtain the most general time solutions whilst remaining numerically stable \citep{10.1145/321439.321446, deHoog:1982:IMN, kuhlman2012, holt2022neural}.

\paragraph{Loss Function} 
Neural Laplace Control trains its dynamics model end-to-end using the mean squared error loss of the next step ahead prediction error,
\begin{align}\label{lossfunc}
    \mathcal{J}(\zeta,\xi,\psi) &=  \sum\nolimits_{t\in \{t_{i+1},\ldots,t_n\}}  \left\| \bm{\hat{x}}(t) - \bm{x}(t) \right\|_2^2 \\
    \text{where}\quad \bm{\hat{x}}(t)&=\mathcal{L}^{-1}\{\bm{X}(\cdot)=u^{-1}(g_{\psi}(\bm{p}_i,u(\cdot))\}(t)
\end{align}
We minimize the above loss function $\mathcal{J}$ to learn the encoders $h_\zeta,h_\xi$ and the Laplace representation network $g_\psi$. This training is summarized in Appendices \ref{problemandbackgroundappendix} and \ref{datasetgeneationandmodeltraining}.
\subsection{Planning with the Learnt Dynamics Model}
\label{sec:planning}
Once we have learnt the dynamics model, any arbitrary model-based reinforcement learning framework can be used for planning. The straightforward choice would be to perform deep Q-learning using the dynamics model as a simulator. Ideally, we want to pursue a more principled approach that can accommodate the Laplace-domain representation of the dynamics more readily.
Specifically, this Laplace-domain representation provides us with the important benefit of being able to simulate state trajectories for arbitrarily long control signals without requiring any additional compute. This is \textit{not} the case for neural ODEs \citep{holt2022neural}, as shown in Section \ref{insights}.

\paragraph{Laplace-model with MPC}
We opt to use the model predictive controller of Model Predictive Path Integral (MPPI) \citep{williams2017information}. This uses a zeroth order particle-based trajectory optimizer method with our learned Laplace dynamics model.
Specifically, this computes a discrete action sequence up to a fixed time horizon of $H \in \mathbb{R}_+$, and then executes the first element in the planned action sequence.
We note that our continuous-time Laplace-based dynamics model can be used to reconstruct a trajectory at any future time horizon up to $H$ seconds, however it requires that the corresponding control input up to that future time horizon is input into the dynamics model, i.e., $\bm{a}_{[t,t+H)}$. To simplify the planning problem, we assume a state, and action history tuple is given to the Laplace-based dynamics model, along with the time interval to predict the dynamics model next future state at, i.e., an input of $(\bm{x}_t, \bm{a}_{[t-\omega:t]}, \delta)$, to predict $\bm{x}_{t+\delta}$.
Where we denote $\delta\in\mathbb{R}_+$ as the observation time interval, that is the time between two consecutive state observations \footnote{We note that the observation state time interval $\delta$ is the same as the time interval between the executed actions.}.
It is natural for online control problems to be controlled at discrete-time steps of $\delta$, where $\delta$ can be varied. 
Therefore, the MPPI plans actions at discrete-time steps $\delta$, up to a fixed horizon $H$ by planning ahead $N \in \mathbb{Z}_+$ steps into the future, thus the planning horizon is determined by $H = \delta\cdot N$.
This leverages a number of parallel roll-outs $M \in \mathbb{Z}_+$, a hyper parameter, which can be tuned.
As MPPI is a Monte Carlo based sampler, increasing the number of roll-outs improves the input trajectory optimization, however scales the run-time complexity as $\mathcal{O}(NM)$.
Although planning benefits from having a longer time horizon to use when optimizing the next action trajectory, it becomes computationally infeasible to do so for a large $N$.
Naturally with the Neural Laplace Control dynamics model representation we can change the observation interval $\delta$ time step, to increase it to enable planning at a longer time horizon.
Clearly, however, planning at a longer horizon can compound model inaccuracies \citep{williams2017information}---which can become significant, rendering the model uncontrollable within that planning regime, and is further explored in Section \ref{insights}. 
Additionally, we also detail the MPC MPPI pseudocode and planner implementation in Appendix \ref{mppipseudocode}.
\section{EXPERIMENTS AND EVALUATION}
\label{mainexperimentssection}
\paragraph{Benchmark Environments}
We use the continuous-time control environments from the ODE-RL suite \citep{yildiz2021continuous}, as they provide true irregular samples in time of state observations and are fully continuous in time, unlike discrete environments \citep{brockman2016openai}. We adapt these to incorporate an arbitrary fixed delayed action time, turning the ODE environments into delay DE environments.
This ODE-RL suite consists of three environments of the Pendulum, Cartpole and Acrobot. The starting state for all tasks is hanging down and the goal is to swing up and stabilize the pole(s) upright \citep{yildiz2021continuous}. Here, each environment uses the reward function of the exponential of the negative distance from the current state to the goal state $\bm{x}^*$, whilst also penalizing the magnitude of action, and we assume we are given this reward function when planning. We detail all environments in Appendix \ref{EnviromentSelectionanddetails}.

\paragraph{Benchmark Dynamic Models}
We select benchmark dynamics models for our specific setting of having both \textit{states observed at irregular time intervals $\Delta_i \neq \Delta_j$} and an \textit{unknown fixed delay $\tau > 0$ in the environment}.
We benchmark against a \textit{discrete-delay method} of a RNN over the action buffer and current state \citep{chen2021delay}, and adapt it to model continuous-time with a new input of the time increment to predict the next state for (\textbf{$\Delta t-$RNN}) \footnote{We note to adapt discrete-time models to continuous-time we add an additional input parameter, that of the time difference between the current time and the next state observation to predict, i.e., $\delta$, e.g., $\bm{x}_{i+1}=\bm{x}_i+\bm{f}(\bm{x}_i,\bm{a}_i,\delta)$.
\citep{yildiz2021continuous}}. We also compare with the true environment dynamics (\textbf{Oracle}), an augmented Neural-ODE (\textbf{NODE}) \citep{chen2018neural}, Latent-ODE (\textbf{Latent-ODE}) \citep{rubanova2019latent} and our Neural Laplace Control (\textbf{NLC}) model. We plan all dynamics models with the MPC MPPI method \citep{williams2017information}, and further compare against a random policy (\textbf{Random}). We provide further details of model selection, hyperparameter selection and implementation details in Appendix \ref{benchmarkmethodimplementationdetails}.

\begin{table*}[!tb]
  \caption[]{Normalized scores $\mathcal{R}$ of the offline model-based agents, where the irregularly-sampled (P1) offline dataset consists of an action delay (P2) of $\{1,2,3\}$ multiples of the environments observation interval time step $\bar{\Delta} = 0.05$ seconds. Averaged over 20 random seeds, with $\pm$ standard deviations.
  Scores are un-discounted cumulative rewards normalized to be between 0 and 100, where 0 corresponds to the Random agent and 100 corresponds to the expert with the \textit{known} world model (Oracle+MPC). Negative normalized scores, i.e., worse than random are set to zero. Full results are included in the Appendix \ref{rawresults}.}
  \resizebox{\textwidth}{!}{
  \begin{tabular}{@{}l|ccc|ccc|ccc}
  \toprule
                                  & \multicolumn{3}{c|}{Action Delay~$\tau=\bar{\Delta}$}      & \multicolumn{3}{c|}{Action Delay~$\tau=2\bar{\Delta}$}      &  \multicolumn{3}{c}{Action Delay~$\tau=3\bar{\Delta}$}               \\
          Dynamics Model                  & Cartpole & Pendulum & Acrobot                   & Cartpole & Pendulum & Acrobot                     & Cartpole & Pendulum & Acrobot \\        
  \midrule
  Random              &     0.0$\pm$0.0 &      0.0$\pm$0.0 &      0.0$\pm$0.0 &      0.0$\pm$0.0 &        0.0$\pm$0.0 &       0.0$\pm$0.0 &     0.0$\pm$0.0 &       0.0$\pm$0.0 &      0.0$\pm$0.0 \\
  Oracle              &     100.0$\pm$0.15 &     100.0$\pm$3.14 &    100.0$\pm$2.19 &     100.0$\pm$0.04 &     100.0$\pm$2.57 &    100.0$\pm$1.79 &     100.0$\pm$0.08 &     100.0$\pm$2.57 &    100.0$\pm$1.26 \\
  $\Delta t-$RNN      &      95.28$\pm$0.4 &      1.14$\pm$6.31 &     18.95$\pm$7.6 &     97.01$\pm$0.31 &      9.94$\pm$2.48 &    28.39$\pm$9.73 &      97.8$\pm$0.25 &    11.81$\pm$11.93 &     3.89$\pm$6.72 \\
  Latent-ODE          &        0.0$\pm$0.0 &        0.0$\pm$0.0 &       0.0$\pm$0.0 &        0.0$\pm$0.0 &     1.24$\pm$20.67 &    8.91$\pm$13.62 &    41.56$\pm$47.07 &     3.26$\pm$12.24 &     9.19$\pm$9.08 \\
  NODE                &     85.09$\pm$7.95 &      0.63$\pm$5.16 &    23.07$\pm$6.94 &     90.75$\pm$1.34 &        0.0$\pm$0.0 &   10.92$\pm$10.09 &     94.55$\pm$1.08 &      1.97$\pm$4.01 &    11.78$\pm$8.33 \\ 
  \midrule
  \bf NLC \textbf{(Ours)} &     \textbf{99.83$\pm$0.19} &     \textbf{98.31$\pm$3.51} &     \textbf{99.12$\pm$1.7} &      \textbf{99.88$\pm$0.1} &     \textbf{93.28$\pm$4.96} &   \textbf{100.44$\pm$2.13} &     \textbf{99.92$\pm$0.12} &     \textbf{98.98$\pm$1.32} &    \textbf{99.46$\pm$1.88} \\
  \bottomrule
  \end{tabular}
  }
  \label{table:main_normalized_scores}
\end{table*}

\paragraph{Offline Dataset Generation}
For each environment we generate an offline state-action trajectory dataset by using an agent that uses an oracle dynamics model combined with MPC and has additional noise added to the agents selected action, $\bar{\pi}(t) = \pi(t) + \epsilon, \epsilon \sim \mathcal{N}(0,\bm{a}_{\max})$. This \quotes{noisy expert} agent interacts with the environment and observes observations at irregular unknown times, where we sample the time interval to the next observation from an exponential distribution, i.e.,  $\Delta \sim \text{Exp}(\bar{\Delta})$, with a mean of $\bar{\Delta}=0.05$ seconds \footnote{We note other irregular sampling types are possible, however \citet{yildiz2021continuous} has shown they are approximately equivalent.
} \citep{yildiz2021continuous}.
We assume a fixed action delay $\tau$, and evaluate discrete multiples of this delay of the mean sampling time $\bar{\Delta}$, i.e., $\tau=\bar{\Delta}$ for one step delay, $\tau=2\bar{\Delta}$ for two step delay etc.
We enforce the observed action history buffer that includes past actions back to $\omega = 4\bar{\Delta}$ seconds. We provide further details on the dataset generation and model training in Appendix \ref{datasetgeneationandmodeltraining}.

\paragraph{Evaluation}
For each environment, with a different delay setting (described above) we collect an offline dataset of irregularly-sampled trajectories, consisting of 1e6 samples from the \quotes{noisy expert} agent interacting within that environment. For each benchmark dynamics model, we follow the same two step evaluation process of, firstly, training the dynamics model on that environment's collected offline dataset using a MSE error loss for the next step ahead state prediction $\hat{\bm{x}}(t_{i+1})$. Then, secondly, taking the same pre-trained model and freezing the weights, and only using it for planning with the MPPI (MPC) planner at run-time in an environment episode, that lasts for $10$ seconds.
In total, we evaluate our model-based control algorithms online in the same environment, running each one for a fixed observation interval of $\delta=\bar{\Delta} = 0.05$ seconds (as is the nominal value for these environments \citep{yildiz2021continuous, brockman2016openai}), and take the cumulative reward value after running one episode of the planner (policy) and repeat this for 20 random seed runs for each result.
We quote the normalized score $\mathcal{R}$ \citep{yu2020mopo} of the policy in the environment, averaged over the 20 random seed run episodes, with standard deviations throughout.
The scores are un-discounted cumulative rewards normalized to lie roughly between 0 and 100, where a score of 0 corresponds to a random policy, and 100 corresponds to an expert (oracle with a MPC planner). 
We further detail our evaluation metrics and experimental setup in Appendix \ref{EvaluationMetrics}.
\subsection{Main results}
\label{main_results}
We compared all the benchmark methods against each environment, which consists of a continuous-time environment with a specific delay---with normalized scores $\mathcal{R}$ are tabulated in Table \ref{table:main_normalized_scores}.
Neural Laplace Control achieves high normalized scores (high episode rewards) on all the environments.
Specifically, NLC is able to model naturally a variety of different delay environment dynamics (P2), whereas existing \textit{delay methods} adapted to continuous-time ($\Delta t-$RNN) struggle to learn appropriate dynamics models for a range of different challenging environments.
Importantly, NLC performs well by learning a $\textit{good}$ dynamics model from the irregularly-sampled offline datasets, whereas the standard \textit{continuous-time methods} (NODE, Latent-ODE) struggle to learn such a model from environments that have an inherent delay.
We also observe similar patterns from additional experiments in Appendix \ref{additonalexperiments}.
\subsection{Insight and Understanding of How Neural Laplace Control Works}
\label{insights}
In this section we seek to gain further insight into \textit{how} Neural Laplace Control outperforms the benchmarks. In the following we seek to understand if NLC is able to learn from irregularly-sampled state-action offline datasets (P1), whilst learning the delayed dynamics of the environment (P2). Furthermore, we also explore the benefits of the NLC approach for planning at longer time horizons with a fixed amount of compute and being sample efficient.

\begin{figure}[!tb]
  \centering
\includegraphics[width=\columnwidth]{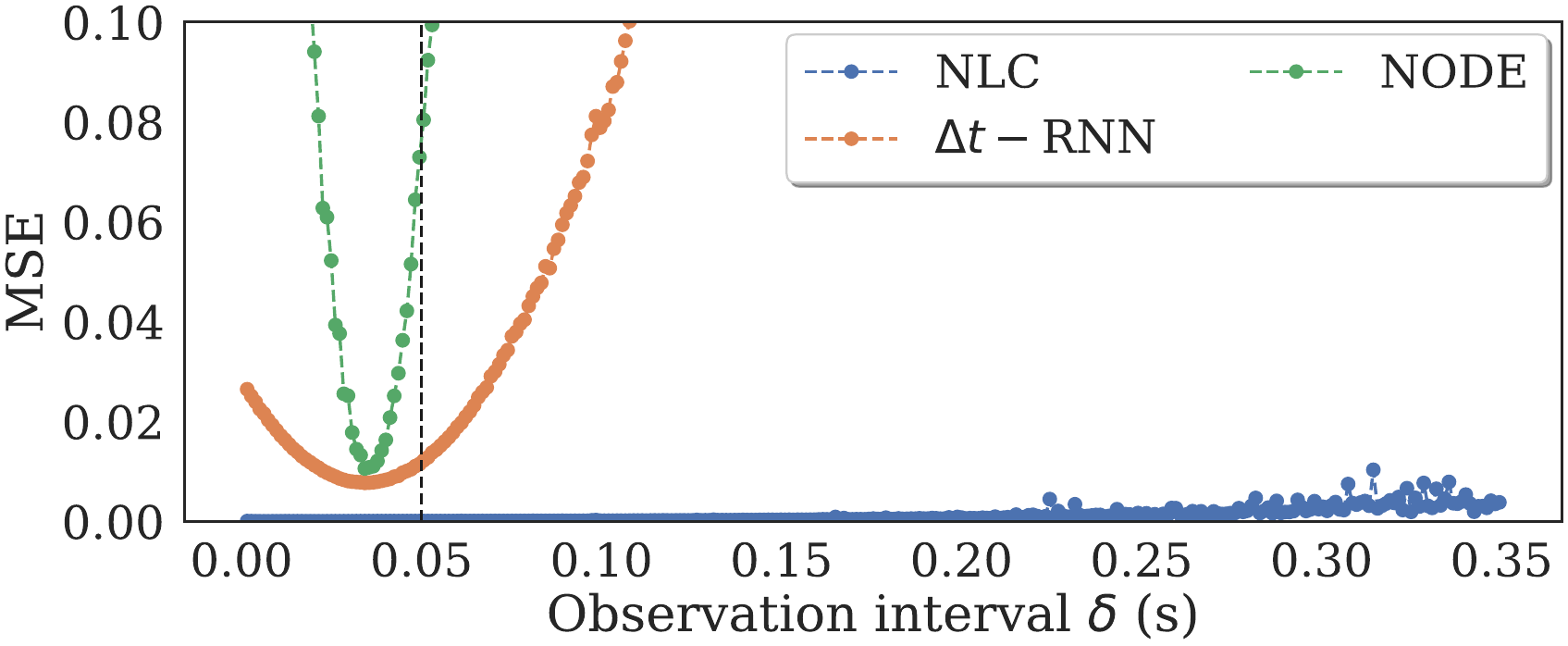}
\caption{Next step ahead validation error (MSE) at a variable time step of an observation interval $\delta$ of the learnt baseline dynamics models, for the irregularly-sampled Cartpole environment with a fixed action delay of $\tau=\bar{\Delta}$. The black dotted line indicates the environments run-time observation interval $\delta=\bar{\Delta}=0.05$ s. Here, we observe Neural Laplace Control learns a good dynamics model over a wide range of observation intervals $\delta$, correctly learning from the irregularly-sampled offline dataset (P1).}
\label{Fig:NextStepAheadError1}
\end{figure}

\paragraph{Can NLC Learn a Good Dynamics Model?}
To explore if NLC is able to learn a suitable dynamics model, we plot the trained models next step ahead prediction error with that of the ground truth for a varying observation interval $\delta$ for the Cartpole environment with a delay of $\bar{\Delta}$, as shown in Figure \ref{Fig:NextStepAheadError1}. Empirically we observe that NLC using its Laplace-based dynamics model is able to better approximate a wider range of observation intervals $\delta$ and achieve a good \textit{global} approximation compared to the recurrent neural network and ODE based models.
We note that due to the offline dataset being sampled with trajectories that have irregular sampling times (P1), where the sampling times are defined by an exponential distribution with a mean of $\bar{\Delta}=0.05$ seconds; the other competing methods seem to over-fit purely to the median sample time of the exponential distribution, i.e., $0.05 \cdot \ln(2) = 0.034$ s.
Other works have shown a more accurate next step prediction model correlates to a higher environment episode reward \citep{williams2017information}. 

\begin{figure}[!tb]
  \centering
\includegraphics[width=\columnwidth]{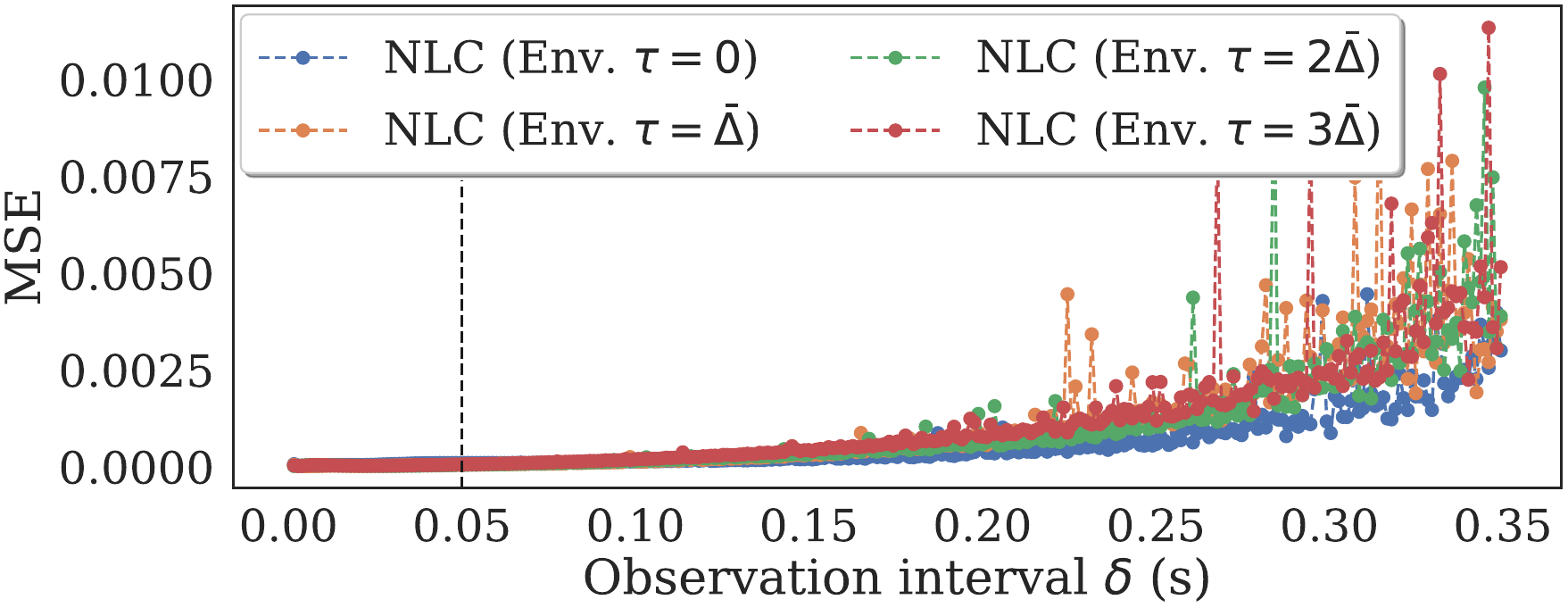}
\caption{Next step ahead validation error (MSE) at a variable time step of an observation interval $\delta$ of the learnt Neural Laplace Control dynamics models, for each delayed environment versions $\tau=\{0,\bar{\Delta},2\bar{\Delta},3\bar{\Delta}\}$ of the specific Cartpole environment. The black dotted line indicates the environments run-time observation interval $\delta=\bar{\Delta}=0.05$ s. Here, Neural Laplace Control is able to correctly learn and capture the delayed dynamics (P2), as the forward MSE errors are low and similar---whereas neural-ODE methods have a greater increasing forward MSE, Appendix \ref{insightexperiments}.}
\label{Fig:NextStepAheadError2}
\end{figure}

\paragraph{Can NLC Learn Delay Environment Dynamics?}
To investigate this, we similarly plot the trained NLC dynamics models next step ahead prediction error with that of the ground truth for a varying observation interval $\delta$, for each of the delayed environment versions of the specific Cartpole environment, as show in Figure \ref{Fig:NextStepAheadError2}. Empirically we observe that the NLC dynamics models correctly learnt the delay dynamics (P2) of each individual environment, as they each have a similar low forward MSE error for the varying levels of inherent delay. In contrast, neural-ODE models are unable to model the delay dynamics correctly, and we observe that they have a higher rate of increasing forward MSE error, that can also increase for an increasing environment delay and is shown further in Appendix \ref{insightexperiments}.

\begin{figure}[!tb]
  \centering
\includegraphics[width=\columnwidth]{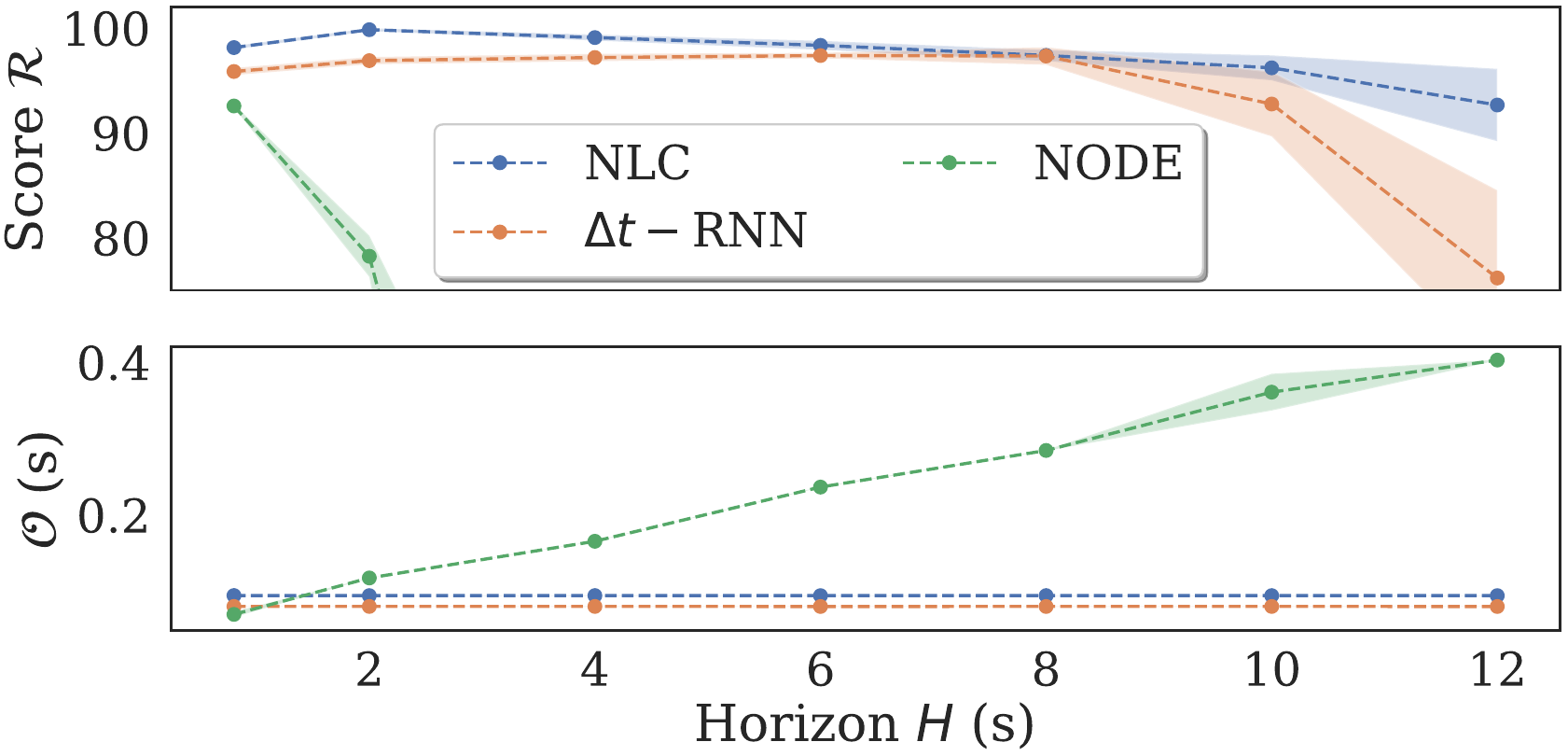}
\caption{Normalized score $\mathcal{R}$ of the baseline methods on the Cartpole environment with an action delay of $\tau=\bar{\Delta}=0.05$ seconds, plotted against an increasing time horizon $H$, by increasing the observation interval $\delta$. NLC, maintains a high performing policy at a longer time horizon---whilst using the same amount of \textit{constant} planning time per action $\mathcal{O}$ as a $\Delta t-$RNN.}
\label{Fig:fixed_amount_of_compute}
\end{figure}

\paragraph{Can NLC Plan with a Longer Time Horizon Using a Fixed Amount of Compute?}
We investigate this by planning with a MPC planner, increasing the observation interval $\delta$ and keeping $N$ fixed, therefore the time horizon $H$ increases---as shown in Figure \ref{Fig:fixed_amount_of_compute}. Here we measure the total planning time taken to plan the next action as $\mathcal{O}$ seconds \footnote{We perform all results using a Intel Core i9-12900K CPU @ 3.20GHz, 64GB RAM with a Nvidia RTX3090 GPU 24GB.} and observe that planning with the NLC dynamics model takes the same amount of planning time, and hence a \textit{fixed amount compute} for planning at a greater time horizon $H$---which is the same as a $\Delta t-$RNN.
This is achieved by the Laplace-based dynamics model that can predict a future state at \textit{any} future time interval using the same number of forward model evaluations, and hence the same amount of compute.
In contrast, this is \textit{not} readily achievable with neural-ODE continuous-time methods that use a larger number of numerical forward steps with a numerical ODE step-wise solver for a increasing time horizon---leading to an increasing planning time for an increasing time horizon, i.e., $\mathcal{O} \propto H$. Furthermore, we highlight, that there exists a trade-off of the time horizon $H$ to plan at---as we wish to use a large \quotes{enough} horizon that captures sufficient future dynamics, whilst minimizing compounded model inaccuracies at a larger planning time horizon. Therefore these two opposing factors, give rise to the maxima of the normalized score $\mathcal{R}$ at a time horizon $H=2$ seconds, as seen in Figure \ref{Fig:fixed_amount_of_compute}.

\begin{figure}[!tb]
  \centering
\includegraphics[width=\columnwidth]{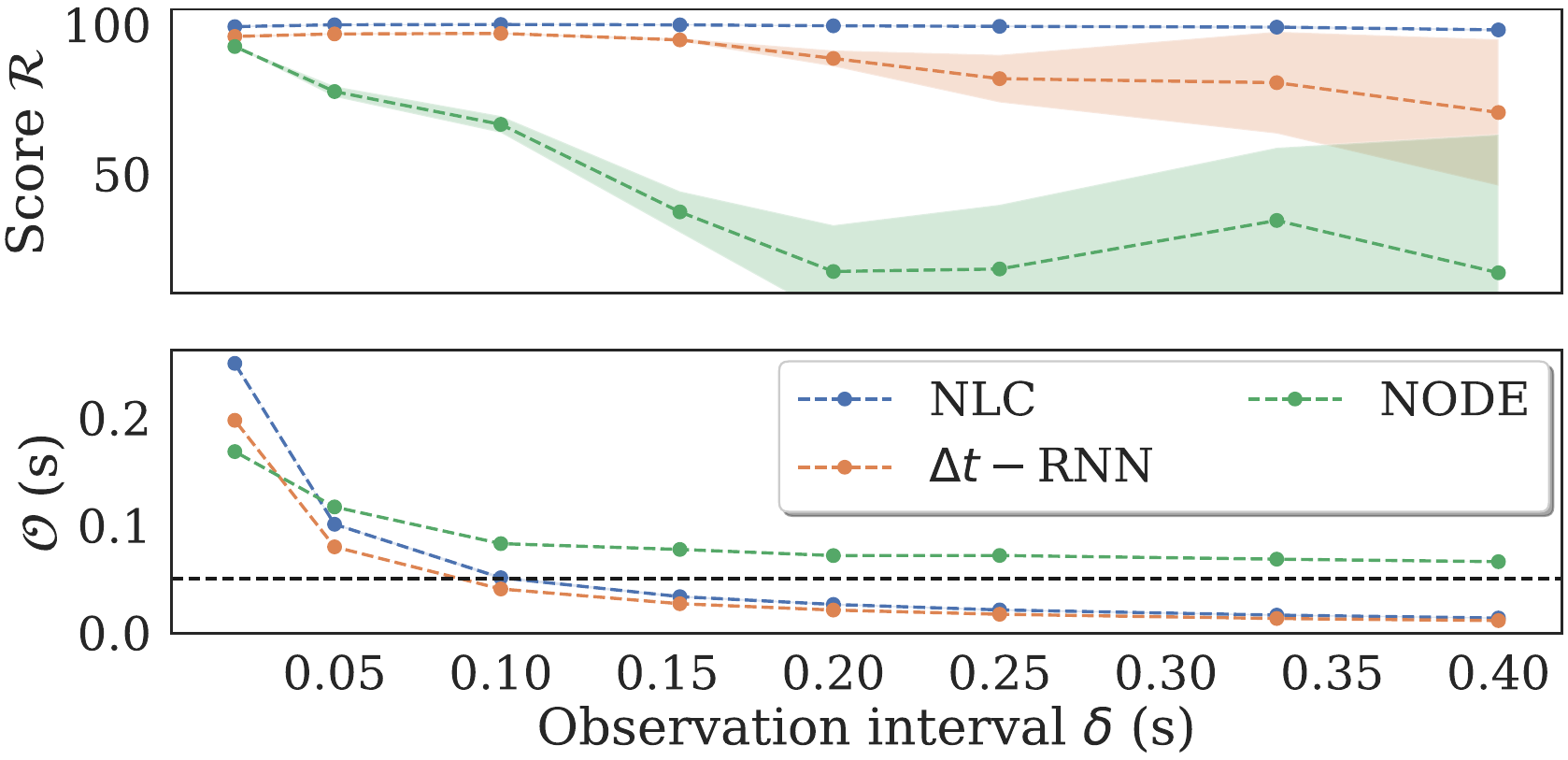}
\caption{Normalized score $\mathcal{R}$ of the baseline methods on the Cartpole environment with an action delay of $\tau=\bar{\Delta}=0.05$s, plotted against an increasing observation interval $\delta$. Here, the time horizon is fixed at $H=2$s, thus increasing the observation interval $\delta$ decreases the number of MPC forward planning steps needed (i.e., $N=\frac{H}{\delta}$). The black dotted line indicates the environments run-time observation interval $\delta=\bar{\Delta}=0.05$ s. NLC demonstrates that it can still outperform the baselines, achieving a near optimal policy---whilst reducing the planning time taken $\mathcal{O}$ needed to generate the next action.}
\label{Fig:DifferentFreqs}
\end{figure}

We further investigate an alternative setup in Figure \ref{Fig:DifferentFreqs}, and keep the time horizon fixed at $H=2$ seconds and increase the observation interval $\delta$---allowing us to reduce $N$ the number of MPC forward planning steps (i.e., $N=\frac{H}{\delta}$). Importantly, this \textit{reduces the planning time} $\mathcal{O}$ needed to generate the next action, enabling a method to use a higher frequency of executing actions to control the dynamics---whilst still planning at the \textit{same fixed time horizon} $H$. NLC is able to still outperform the baselines, achieving a high performing policy---even when using a lesser amount of planning compute per action.
The numeric values, along with those for other environments are provided in Appendix \ref{insightexperiments}.

\begin{figure}[!tb]
  \centering
\includegraphics[width=\columnwidth]{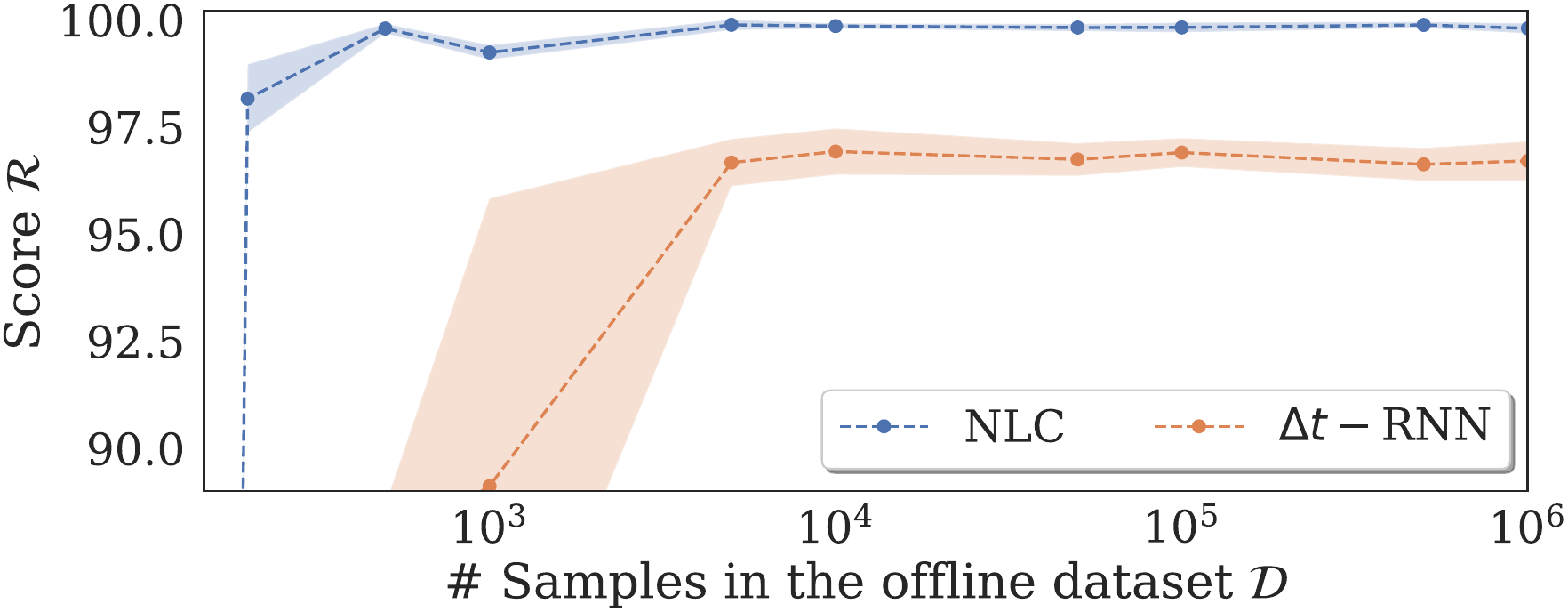}
\caption{Normalized score $\mathcal{R}$ of the baseline methods on the Cartpole environment with an action delay of $\tau=\bar{\Delta}=0.05$s, plotted against the number of samples in the irregularly-sampled offline dataset used to train the dynamics model of each method. The two closest highest performing baselines are plotted here, and refer to Appendix \ref{insightexperiments} for others. NLC can maintain a high performing policy---even from the challenging case of only learning a dynamics model from 200 samples from an irregularly-sampled in time offline dataset $\mathcal{D}$.
}
\label{Fig:Sampleefficiency}
\end{figure}

\paragraph{Is NLC Sample Efficient?}
We observe in Figure \ref{Fig:Sampleefficiency} that NLC can still learn a suitable dynamics model, and perform well on the Cartpole environment with a delay of $\tau=\bar{\Delta}=0.05$ seconds, when trained with an offline irregularly-sampled in time dataset that contains only 200 random samples---which corresponds 10 seconds of interaction time of a noisy expert (expert with random action noise) agent from the delayed environment.
We further detail full results, including other environment results in Appendix \ref{insightexperiments}.

\paragraph{Can NLC Incorporate Adaptive State-based Constraints}
Using an MPC planner, NLC can naturally handle unseen state-based constraints, and we show this in Appendix \ref{insightexperiments}.

\section{DISCUSSION AND FUTURE WORK}

\paragraph{Discussion}
In this work, we have proposed and validated a novel model-based offline RL method, which combines a Neural Laplace dynamics model with a MPC planner. This novel method performs RL in continuous-time, training only on offline irregularly-sampled data that has an inherent delay.
We have shown experimentally that these Laplace-domain models outperform their neural-ODE and recurrent neural-network based counterparts on irregularly-sampled datasets, and make model predictive control feasible for longer time horizons, with a fixed compute budget.

\paragraph{Future Works} 
Our current focus is on continuous-time environments with unknown \textit{fixed} delays. We consider learning in environments with unknown and \textit{variable} delays an important area for future work. Furthermore, in the current work, we solely explored using only one instance of the dynamics model. However, we note that we can extend NLC trivially to create an \textit{ensemble} of Neural Laplace Control models, thereby providing uncertainty estimation (epistemic uncertainty) of the future state prediction.

\paragraph{Societal Impact}
We envisage Neural Laplace Control as a tool to perform offline RL in realistic continuous-control settings, although emphasize that the dynamics action control trajectory proposed would need to be further verified by a human expert or via experimentation.

\subsubsection*{Acknowledgements}
SH would like to acknowledge and thank AstraZeneca for funding. This work was additionally supported by the Office of Naval Research (ONR) and the NSF (Grant number: 1722516). Moreover, we would like to warmly thank all the anonymous reviewers, alongside research group members of the van der Scaar lab, for their valuable input, comments and suggestions as the paper was developed---where all these inputs ultimately improved the paper.

\bibliography{aistats2023}
\bibliographystyle{iclr}

\onecolumn
\appendix

Contents of supplementary materials:
\begin{enumerate}
  \item Appendix \ref{extendedrelatedwork}: Extended Related Work
  \item Appendix \ref{problemandbackgroundappendix}: Problem and Background
  \item Appendix \ref{mppipseudocode}: MPC MPPI Pseudocode and Planner Implementation Details
  \item Appendix \ref{EnviromentSelectionanddetails}: Environment Selection and Details
  \item Appendix \ref{benchmarkmethodimplementationdetails}: Benchmark Method Implementation Details
  \item Appendix \ref{EvaluationMetrics}: Evaluation Metrics
  \item Appendix \ref{datasetgeneationandmodeltraining}: Dataset Generation and Model Training
  \item Appendix \ref{rawresults}: Raw Results
  \item Appendix \ref{insightexperiments}: Insight Experiments
  \item Appendix \ref{additonalexperiments}: Additional Experiments
\end{enumerate}

\paragraph{Code}
We have released a PyTorch implementation \citep{NEURIPS2019_9015} at \href{https://github.com/samholt/NeuralLaplaceControl}{https://github.com/samholt/NeuralLaplaceControl}.
Additionally, we have a research group codebase, available at \href{https://github.com/vanderschaarlab/NeuralLaplaceControl}{https://github.com/vanderschaarlab/NeuralLaplaceControl}.

\textbf{AISTATS 2022 Checklist}
For all models and algorithms presented, check if you include:
\begin{enumerate}
    \item A clear description of the mathematical setting, assumptions, algorithm, and/or model. (Yes, see Section 3.)
    \item An analysis of the properties and complexity (time, space, sample size) of any algorithm. (Yes, see Section 5.2 and Appendix \ref{insightexperiments}.)
    \item (Optional) Anonymized source code, with specification of all dependencies, including external libraries. (\href{https://github.com/samholt/NeuralLaplaceControl}{https://github.com/samholt/NeuralLaplaceControl}.)
\end{enumerate}

For any theoretical claim, check if you include:
\begin{enumerate}
    \item A statement of the result. (Not applicable.)
    \item A clear explanation of any assumptions. (Not applicable.)
    \item A complete proof of the claim. (Not applicable.)
\end{enumerate}

For all figures and tables that present empirical results, check if you include:
\begin{enumerate}
    \item A complete description of the data collection process, including sample size. (Yes, see Section 5 and Appendix \ref{datasetgeneationandmodeltraining}.)
    \item A link to a downloadable version of the dataset or simulation environment. (Yes, see Appendix \ref{EnviromentSelectionanddetails}.)
    \item An explanation of any data that were excluded, description of any pre-processing step. (Yes, see Appendix \ref{datasetgeneationandmodeltraining}.)
    \item An explanation of how samples were allocated for training / validation / testing. (Yes, see Appendix \ref{datasetgeneationandmodeltraining}.)
    \item The range of hyper-parameters considered, method to select the best hyper-parameter configuration, and specification of all hyper-parameters used to generate results. (Yes, see Appendix \ref{benchmarkmethodimplementationdetails}.)
    \item The exact number of evaluation runs. (Yes, see Section 5.)
    \item A description of how experiments were run. (Yes, see Section 5 and Appendix \ref{EvaluationMetrics}.)
    \item A clear definition of the specific measure or statistics used to report results. (Yes, see Section 5 and Appendix \ref{EvaluationMetrics}.)
    \item Clearly defined error bars. (Yes, see Section 5 and Appendix \ref{EvaluationMetrics}.)
    \item A description of results with central tendency (e.g., mean) \& variation (e.g., stddev). (Yes, see Section 5, and results throughout.)
    \item A description of the computing infrastructure used. (Yes, see Section 5.2. and Appendix \ref{EvaluationMetrics}.)
\end{enumerate}

\section{EXTENDED RELATED WORK}
\label{extendedrelatedwork}

\begin{table*}[!htb]
    \centering
    \caption{Comparison with related model-based approaches to RL. \textbf{(P1) Learn from irregular samples}---can it learn from an offline dataset sampled at irregular times, $\Delta_i \neq \Delta_j$? \textbf{(P2) Learn delayed dynamics}---can it learn environments that contain a delay $\tau > 0$? Neural Laplace Control is the only method that can both learn from irregular samples (P1) as well as learn environments that contain a delay (P2).}
    \label{table:main_related_work_main_extended}
    \small
    \resizebox{\linewidth}{!}{%
    \begin{tabular}{@{}lcccccc@{}}
        \toprule
        \bf Approach & \bf True Dynamics & \bf Data Available & \bf Reference \ & \bf Model & \bf (P1) $\Delta_i \neq \Delta_j$ & \bf (P2) $\tau>0$ \\
        \midrule
        Conventional model-based RL & $\bm{x}_{t+1}\sim f(\bm{x}_t,\bm{a}_t)$ & $\mathcal{D}=\{(\bm{x}_i,\bm{a}_i)\}_{i=0}^n$ & \citet{williams2017information} & MDP~/~Neural Network & \xmark & \xmark \\
        Discrete-time delay methods & $\bm{x}_{t+1}\sim f(\bm{x}_t,\bm{a}_{t-\tau})$ & $\mathcal{D}=\{(\bm{x}_i,\bm{a}_i)\}_{i=0}^n$ & \citet{chen2021delay} & DA-MDP / RNN & \xmark & \cmark \\
        \multirow{2}{*}{Continuous-time methods} & \multirow{2}{*}{$\bm{\dot{x}}(t)=f(\bm{x}(t),\bm{a}(t))$} & \multirow{2}{*}[-1.2pt]{\makecell{$\mathcal{D}=\{(\bm{x}(t_i),\bm{a}(t_i))\}_{i=0}^n$\\[-2.4pt]\scriptsize s.t. $\exists i,\!j: t_{i+1}\!-\!t_i\neq t_{j+1}\!-\!t_j$}} & \citet{yildiz2021continuous} & Neural ODE & \cmark & \xmark \\
        & & & \citet{du2020model} & Latent ODE & \cmark & \xmark \\
        \midrule
        \bf Neural Laplace Control & $\bm{\dot{x}}(t)=f(\bm{x}(t),\bm{a}(t-\tau))$ & \makecell{$\mathcal{D}=\{(\bm{x}(t_i),\bm{a}(t_i))\}_{i=0}^n$\\[-2.4pt]\scriptsize s.t. $\exists i,\!j: t_{i+1}\!-\!t_i\neq t_{j+1}\!-\!t_j$} & \bf (Ours) & Neural Laplace Control & \cmark & \cmark \\
        \bottomrule
    \end{tabular}}
\end{table*}

In the following we summarize the key related work in Table \ref{table:main_related_work_main_extended} and provide an extended discussion of additional related works including a review of the benefits of using model-based RL and using model predictive control, which happens to be our preferred strategy for planning policies.

\paragraph{Why model-based reinforcement learning?}
Model-based RL holds the promise to enable creating policies for real world tasks in the offline setting where the true environment dynamics are \textit{unknown}, and instead we require to learn a dynamics model from a dataset of demonstrations from agents acting within the environment.
Although existing model-free RL approaches have shown effective performance \citep{fujimoto2018addressing}, they have the inherent disadvantages of being less sample efficient \citep{lutter2021learning} where they require interacting either online or with the \textit{known} environment dynamics, and often require millions or billions of interactions with the environment to learn a good policy. Furthermore, learning a dynamics model of the environment, allows planning over that dynamics model to optimize actions (MPC) rather than learning a specific policy---by planning, a model can easily adapt to different goals or tasks at run-time. Whereas a policy trained for a specific task or goal would often have to be re-trained for a new task or goal, making it difficult for a policy to adapt to multi-task settings \citep{lutter2021learning}.
Naturally in continuous-control settings the dynamics model (e.g., the physics of the environment) is independent of the reward function (e.g., the goal state to reach), therefore changing tasks are straightforward by changing the reward function arbitrarily. 
An important property of model-based reinforcement learning is that in general it is more sample-efficient than model-free methods in conventional control tasks \citep{wang2019benchmarking,moerland2020model}.
While model-free methods learn to master challenging tasks \citep{mnih2015human, lillicrap2015continuous} and improves learning efficiency in high-dimensional continuous control tasks \citep{fujimoto2018addressing, haarnoja2018soft}, it was later shown in \citet{ha2018world, janner2019trust} that model-based methods have much higher sample efficiency once properly tuned. Furthermore, \citet{hafner2019dream, sharma2019dynamics} propose to learn dynamics in a latent space, and similar insights have been applied to further improve the model-based reinforcement learning performance \citep{hansen2022temporal}.

In offline reinforcement learning, an agent learns from a fixed replay buffer and is not permitted to interact with the environment \citep{wu2019behavior}. While both model-free \citep{kumar2019stabilizing,kumar2020conservative,fujimoto2021minimalist} and model-based \citep{kidambi2020morel,wang2021offline} approaches have been proposed for offline RL, in general, model-based methods have been shown to be more sample efficient than model-free methods \citep{moerland2020model}. The main challenge in model-based RL is known as \quotes{extrapolation error} \citep{fujimoto2019off}, whereby the learnt dynamics model inaccuracies compound for a larger number of future predicted time steps. Hence, it is crucial in model-based RL to learn an appropriate dynamics model that is capable of accurately capturing the unique characteristics of an environment. However, even though many environments operate in continuous-time by nature and contain action or observation delays, almost all the existing approaches to model-based RL consider dynamics models only suited to the conventional discrete-time $\Delta_i = \Delta_j$ setting with no delays $\tau = 0$. We review here some of the few approaches that go beyond the conventional setting, namely (i) \textit{discrete-time delay methods} and (ii) \textit{continuous-time methods}.

\paragraph{Discrete-time delay methods} 
One approach for handling environment observation delays is to increase the time step till the next action is performed, that is to synchronize an agent's actions with its delayed observations. However, such an approach is infeasible in most environments (for example dynamics involving momentum), and even when it is feasible, such a \quotes{wait agent} is often sub optimal, as it is possible to perform better by acting before receiving the most recent observation \citep{walsh2009learning}.
We note that modeling environments with either delayed observations $\bm{x}(t+\tau)$ or delayed actions $\bm{a}(t+\tau)$ are equivalent in form \citep{katsikopoulos2003markov}. Prior work models regular sampled $\Delta_i=\Delta_j$ (discrete time) environments with constant time delays $\tau>0$, and provides the agent with the current state $\bm{x}(t)$, and a history of past actions performed in the environment $\bm{\bar{a}}_{i-1}=\{\bm{a}_1,\ldots,\bm{a}_{i-1}\}$, whereby the history action window is larger than or equal to the observation or action delay in the environment \citep{walsh2009learning,firoiu2018human,bouteiller2020reinforcement}. Recently, \citet{chen2021delay} proposed delay-aware Markov decision processes (MDPs) that are capable of modeling delayed dynamics in discrete-time based on regularly sampled data, with an RNN encoding the history of past actions and the current state.

\paragraph{Continuous-time methods}
A standard approach for applying model-based RL to irregular sampled time series is to divide the timeline into equally sized intervals and impute or aggregate state and action tuples using averages \citep{rubanova2019latent}.
Thus, turning a continuous-time environment into a discrete-time environment approximation; however, such pre-processing destroys information, particularly about the timing of measurements and the specific underlying environment dynamics.
Real world data is often sampled irregularly $\Delta_i \neq \Delta_j$, as such \citep{yildiz2021continuous} propose to use Neural-ODEs \citep{chen2018neural} as their continuous-time dynamics model can model irregularly-sampled environments with no delays $\tau=0$. Similarly, the work of \citet{du2020model} uses an a Latent-ODE model when planning policies. However, these existing approaches are limiting, as an ODE-based model by definition cannot handle a delay differential equation, necessitating the need for a model that can learn and model more diverse classes of differential equations. Recent models, of modeling diverse classes of differential equations is made possible with the work of Neural Laplace \citep{holt2022neural} by representing them in the Laplace domain. These Laplace-based models have been shown to be able to model such systems, be more accurate and scale better with increasing time horizons in time complexity. Our approach, namely Neural Laplace Control, essentially extends Neural Laplace to the setting of controlled systems---that is systems that evolve based on an action signal $\bm{a}(t)$---so that it can be used in planning policies in a RL setting.

\paragraph{Model predictive control (MPC)}
Principally relies on a good dynamics model, historically using simple first principle \textit{known} dynamic models \citep{richalet1978model, salzmann2022neural}. Recently, MPC Model Predictive Path Integral (MPPI) \citep{williams2017information} is a zeroth order particle-based trajectory optimizer method that is capable of handling complex cost criteria and general nonlinear dynamics. Specifically, \citet{williams2017information} showed it could be used with a neural network learned dynamics model and used to drive a toy vehicle on a dirt track. MPC, and hence MPPI is often computationally infeasible for long time horizons, therefore often only being run for a fixed receding time horizon optimization into the future with a dynamics model. Using an MPC planner benefits from being able to handle arbitrary state constraints, and changing goals. Here MPPI is the state-of-the-art for MPC with a learned dynamics model, improving upon the previous cross-entropy method (CEM) \citep{kobilarov2012cross} MPC method. These naturally can incorporate new state-based constraints at run time.

\paragraph{Hybrid MPC}
Various ways of combining a powerful MPC planner with an accurate dynamics model is another fruitful thread \citep{argenson2020model}. All existing hybrid works, work only on discrete domains where demonstration data is collected on regular time intervals. These include MBOP \citep{argenson2020model}, TD-MPC \citep{hansen2022temporal} and DADS \citep{sharma2019dynamics}. We highlight that hybrid methods that plan in the latent space, i.e., TD-MPC and DADS are unable to incorporate state-based constraints. However, these methods still struggle with scaling MPC computational complexity forwards for longer time horizons.

\paragraph{Control literature}
We perform full system identification, i.e., learning the nonlinear dynamics model that has an \textit{unknown} inherent delay. 
Whereas the existing control literature provides control algorithms for known forms (often linear) dynamics models for a \textit{known} delay \citep{kwon2003simple, raff2007model}---therefore are not comparable. Moreover, there exists a wealth of orthogonal related work on stability analysis in Control \citep{aastrom2010feedback}.

\paragraph{Learning from noisy demonstrations}
It is preferable to learn a dynamics model on state-action trajectories that come from a \quotes{noisy} expert. As a noisy expert can provide better trajectories than an expert as it shows how to recover from \quotes{bad} states \citep{laskey2017dart}---specifically we assume the true expert is \textit{unknown}. Moreover, as we only have access to trajectories from a noisy expert, performing imitation learning \citep{pmlr-v162-liotet22a} would propagate the noisy behavior, achieving a poor performance.

\section{NEURAL LAPLACE BACKGROUND}
\label{problemandbackgroundappendix}

In the following we provide a brief Laplace background, specifically from that of the Neural Laplace \citep{holt2022neural} model for modeling diverse differential equation (DE) systems---in the context of Neural Laplace Control. We defer the reader to the work of \citet{holt2022neural} for a full comprehensive explanation of the original Neural Laplace model.

\paragraph{States \& actions}
For a system with \textit{state} space $\mathcal{X}=\mathbb{R}^{d_\mathcal{X}}$ and \textit{action} space $\mathcal{A}=\mathbb{R}^{d_\mathcal{A}}$, the state at time $t\in\mathbb{R}$ is denoted as $\bm{x}(t)=[x_1(t),\ldots,x_{d_\mathcal{X}}(t)]\in\mathcal{X}$ and the action at time $t\in\mathbb{R}$ is denoted as \mbox{$\bm{a}(t)=[a_1(t),\ldots,a_{d_\mathcal{A}}(t)]\in\mathcal{A}$}. We elaborate that \textit{state trajectory} $\bm{x}:\mathbb{R}\to\mathcal{X}$ and \textit{action trajectory} $\bm{a}:\mathbb{R}\to\mathcal{A}$ are both functions of time, where an individual state $\bm{x}(t)\in\mathcal{X}$ or an individual action~$\bm{a}(t)\in\mathcal{A}$ are points on these trajectories. Given a time interval~$\mathcal{I}\subseteq \mathbb{R}$, $\bm{x}_{\mathcal{I}}\in\mathcal{X}^{\mathcal{I}}$ and $\bm{a}_{\mathcal{I}}\in\mathcal{A}^{\mathcal{I}}$ we denote the partial state and action trajectories on that interval such that $\bm{x}_{\mathcal{I}}(t)=\bm{x}(t)$ and $\bm{a}_{\mathcal{I}}(t)=\bm{a}(t)$ for $t\in\mathcal{I}$.

\paragraph{Laplace Transform}
The Laplace transform of a trajectory $\bm{x}$ is defined as \citep{schiff1999laplace}
\begin{equation}
    \bm{X}(\bm{s})=\mathcal{L}\{\bm{x}\}(\bm{s})=\int_0^\infty e^{-\bm{s}t} \bm{x}(t) dt,
\label{mainlt}
\end{equation}
where $\bm{s}\in \mathbb{C}^{d_\mathcal{X}}$ is a vector of \textit{complex} numbers and $\bm{X}(\bm{s}) \in \mathbb{C}^{d_\mathcal{X}}$ is called the \textit{Laplace representation}.
The $\bm{X}(\bm{s})$ may have singularities, i.e., points where $\bm{X}(\bm{s})\to \bm{\infty}$ for one component \citep{schiff1999laplace}.
Importantly, the Laplace transform is well-defined for trajectories that are \textit{piecewise continuous}, i.e., having a finite number of isolated and finite discontinuities \citep{Poularikas2000TheTA}.
This property allows a learned Laplace representation to model a dynamics model that can have delay differential equation solutions \citep{holt2022neural}.

\paragraph{Inverse Laplace Transform}
The inverse Laplace transform (ILT) is defined as
\begin{equation}
    \hat{\bm{x}}(t) = \mathcal{L}^{-1}\{\bm{X}(\bm{s})\}(t)=\frac{1}{2\pi i} \int_{\sigma - i \infty}^{\sigma + i \infty} \bm{X}(\bm{s})e^{\bm{s}t}d\bm{s},
\label{ilt}
\end{equation}
where the integral refers to the Bromwich contour integral in $\mathbb{C}^{d_\mathcal{X}}$ with the contour $\sigma>0$ chosen such that all the singularities of $\bm{X}(\bm{s})$ are to the left of it \citep{schiff1999laplace}. 
Many algorithms have been developed to numerically evaluate Equation \ref{ilt}. On a high level, they involve two steps: \citep{10.1145/321439.321446, deHoog:1982:IMN, kuhlman2012}.
\begin{align}
\label{eq:ILT-Query}
    \mathcal{Q}(t) &= \text{ILT-Query} (t) \\
    \label{eq:ILT-Compute}
    \hat{\bm{x}}(t) &= \text{ILT-Compute}\big(\{\bm{X}(\bm{s})| \bm{s} \in \mathcal{Q}(t) \}\big)
\end{align}

To evaluate $\bm{x}(t)$ on time points $t \in \mathcal{T} \subset \mathbb{R}_+$, the algorithms first construct a set of \textit{query points} $\bm{s} \in \mathcal{Q}(\mathcal{T}) \subset \mathbb{C}$. They then compute $\hat{\bm{x}}(t)$ using the $\bm{X}(\bm{s})$ evaluated on these points.
The number of query points scales \textit{linearly} with the number of time points, i.e., $|\mathcal{Q}(\mathcal{T})| = d_{\mathcal{S}} |\mathcal{T}|$, where the constant $d_{\mathcal{S}} > 1$, denotes the number of reconstruction terms per time point and is specific to the algorithm. 
Importantly, the computation complexity of ILT only depends on the \textit{number} of time points, but not their values (e.g., ILT for $t=0$ and $t=100$ requires the same amount of computation).
The vast majority of ILT algorithms are differentiable with respect to $\bm{X}(\bm{s})$, which allows the gradients to be back propagated through the ILT transform \citep{holt2022neural}. 

Intuitively, the inverse Laplace transform (ILT) (Equation \ref{ilt}) reconstructs the dynamics model time solution with the basis functions of complex exponentials $e^{\bm{s}t}$, which exhibit a mixture of \textit{sinusoidal} and \textit{exponential} components \citep{schiff1999laplace, 10.5555/281875, kuhlman2012}.

\paragraph{Solving control of differential equations in the Laplace domain}
A key application of the Laplace transform is to solve broad classes of DEs \citep{podlubny1997laplace, yousef2018application, yi2006solution, kexue2011laplace}. Due to the Laplace derivative theorem \citep{schiff1999laplace}, the Laplace transform can convert a DE into an \textit{algebraic equation} even when the DE contains historical states $\bm{x}(t-\tau)$ (as in a delayed DE).
It also applies to coupled DEs and can allow decoupled solutions to coupled DEs for dynamical systems \citep{aastrom2010feedback}.
The resulting algebraic equation can either be solved analytically or numerically to obtain the solution of the DE, $\bm{X}(\bm{s})$, in the Laplace domain.
Finally, one can obtain the time solution $\bm{x}(t)$ by applying the ILT on $\bm{X}(\bm{s})$. 
For instance, we could use the concise \textit{Laplace transform method} to solve the (delay) differential equations to get solutions for the state trajectories conditioned on a control input trajectory \citep{yi2008controllability}. 

\paragraph{Stereographic projection}
However, the Laplace representation $\bm{X}(\bm{s})$ often involves singularities \citep{schiff1999laplace}, which are difficult for neural networks to approximate or represent \citep{DBLP:journals/rc/BakerP98}.
We instead propose to use a stereographic projection $u(s) = (\theta, \phi)$ to translate any complex number $s\in \mathbb{C}$ into a coordinate on the Riemann Sphere $(\theta, \phi) \in \mathcal{D} = (-{\pi}, {\pi}) \times (-\frac{\pi}{2}, \frac{\pi}{2})$ \citep{10.5555/26851}, i.e.,
\begin{align}
\label{phi0}
     u(s) = \left( \arctan \left( \frac{\text{Im}(s)}{\text{Re}(s)} \right),\arcsin \left( \frac{|s|^2-1}{|s|^2+1} \right) \right)
\end{align}
Where the associated inverse transform, $u^{-1}: \mathcal{D} \rightarrow \mathbb{C}$, is given as
\begin{align}
\label{polarcoords}
    s = u^{-1}(\theta, \phi) = \tan \left( \frac{\phi}{2} + \frac{\pi}{4} \right) e^{i \theta}
\end{align}
A nice example of this map is the function of $1/s$, which corresponds to a rotation of the Riemann-sphere $180^{\circ}$ about the real axis. Therefore, a representation of $1/s$ under this transformation becomes the map $\theta, \phi \mapsto - \theta, - \phi$ \citep{10.5555/26851}. 
\begin{figure}[!htb]
\begin{center}
      \includegraphics[width=0.35\textwidth]{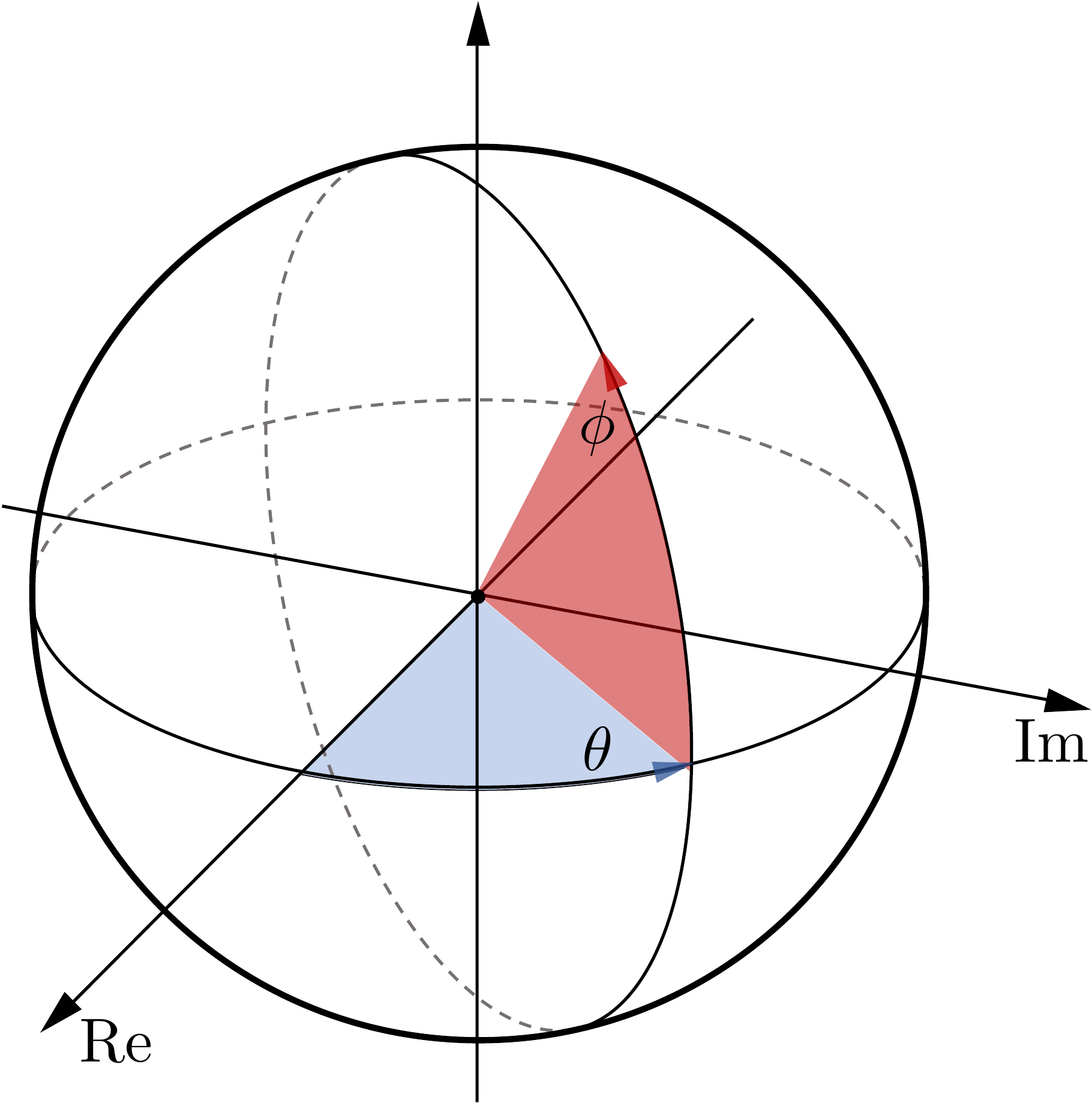}
     \end{center}
  \caption{Geometry of the Riemann sphere map for a complex number $\mathbb{C}$ into a spherical co-ordinate representation of $\theta, \phi$.}
    \label{spheregeom}
\end{figure}

\paragraph{Inverse Laplace transform}
After obtaining the Laplace representation $\bm{X}(\bm{s})$ from Equation 6 (see main paper), we compute the predicted or reconstructed state values $\hat{\bm{x}}(t)$ using the ILT. 
We highlight that we can evaluate $\hat{\bm{x}}(t)$ \textit{at any future time} $t\in \mathbb{R}_+$ as the Laplace representation is independent of time once learnt.
In practice, we use the well-known ILT Fourier series inverse algorithm (ILT-FSI), which can obtain the most general time solutions whilst remaining numerically stable \citep{10.1145/321439.321446, deHoog:1982:IMN, kuhlman2012}. We use the specific ILT Fourier series algorithm from \citet{holt2022neural} and use their code implementation of the ILT algorithm.

\section{MPC MPPI PSEUDOCODE AND PLANNER IMPLEMENTATION DETAILS}
\label{mppipseudocode}

We opt to use the model predictive controller of Model Predictive Path Integral (MPPI) \citep{williams2017information}. This uses a zeroth order particle-based trajectory optimizer method with our learned Laplace dynamics model.
Specifically, this computes a discrete action sequence up to a fixed time horizon of $H \in \mathbb{R}_+$ seconds, and then executes the first element in the planned action sequence.
Where we denote $\delta\in\mathbb{R}_+$ as the observation time interval, that is the time between two consecutive state observations.
It is natural for online control problems to be controlled at discrete-time steps of $\delta$, where $\delta$ can be varied. 
Therefore, the MPPI plans actions at discrete-time steps $\delta$, up to a fixed time horizon $H$ by planning ahead $N \in \mathbb{Z}_+$ steps into the future, thus the planning time horizon is determined by $H = \delta\cdot N$.
This leverages a number of parallel roll-outs $M \in \mathbb{Z}_+$, a hyper parameter, which can be tuned.
As MPPI is a Monte Carlo based sampler, increasing the number of roll-outs improves the input trajectory optimization, however, scales the run-time complexity as $\mathcal{O}(NM)$.

We use the standard MMPI algorithm \citep{williams2017information} with our dynamics model $F$, with the slight modification where we provide the dynamics model the current state, action and a buffer of previous actions back to $\omega$ seconds, i.e., $\bm{x}_{t+\delta}=F(\bm{x}_t, \bm{a}_{[t-\omega:t]}, \delta)$.
For simplification of notation, as MPPI plans at discrete time steps of $\delta$ seconds, we relax the $\delta$ notation to discrete time steps of $\delta$, where $\delta=\bar{\Delta}=0.05$ seconds at run-time. 
Specifically, for $\omega=4\bar{\Delta}=4\delta$ we denote the discrete multiple of $\delta$ as $\bar{\omega}=4$, the next state estimate is given by $\bm{x}_{t+1}=F(\bm{x}_t, \bm{a}_{[t-\bar{\omega}:t]})$.
Furthermore, MPPI requires us to keep in memory the global action trajectory $\bm{T}\in \mathbb{R}^{(N+\bar{\omega}) \times d_\mathcal{A}}$ buffer of the previously planned action trajectory. With slight abuse of notation, we define the global action trajectory that is used to plan ahead $N$ discrete time steps to also contain the past $\bar{\omega}$ action histories, i.e., $\bm{T}_{[0-\bar{\omega}:N]}$.
We detail the MPPI pseudocode with the high-level policy in Algorithm \ref{alg:high_level_policy} and the MPPI action trajectory optimizer in Algorithm \ref{alg:trajectory_optomiziation}.

\begin{algorithm}[H]
\caption{High-Level MPPI Policy} \label{alg:high_level_policy}
\begin{algorithmic}
\State {\bfseries Input:} Pre-trained dynamics model $F$.
\State $\bm{T}^0 \gets [\bm{0}_{0-\bar{\omega}},\dots,\bm{0}_{N-1}]$ \Comment{Initialize planned action trajectory.}
\For{$t=1\dots \infty$}
\State $\bm{x}_t \gets$ Observe $\bm{x}_t$
\State $\bm{T}^{t} \gets \text{MPPI-Trajectory-Optimization}(F,\bm{x}_t,\bm{T}^{t-1})$ \Comment{Update planned trajectory $\bm{T}^t$ starting with $\bm{T}_{[0-\bar{\omega}:0]}$.}
\State $\bm{a}_t \gets \bm{T}_{0}^t$ \Comment{Use first action $\bm{T}_0$ as $\pi(\bm{x}_t)$.}
\EndFor
\end{algorithmic}
\end{algorithm}

\begin{algorithm}[H]
\caption{MPPI-Trajectory-Optimization} \label{alg:trajectory_optomiziation}
\begin{algorithmic}
\State {\bfseries Input:} Pre-trained dynamics model $F$, starting state $\bm{x}$, previous global action trajectory $\bm{T}$, steps to plan ahead for $N$, number of parallel roll-outs $M$, noise covariance $\bm{\Sigma}$, hyper parameters $\lambda$, action max $\bm{a}_{\max}$, action min $\bm{a}_{\min}$.
\State $\bm{R}_{M} \gets \bf{0}_{M}$ \Comment{This holds our $M$ trajectory returns.}
\State $\bm{A}_{M,N+\bar{\omega}} \gets \bf{0}_{M,N+\bar{\omega}}$ \Comment{This holds our $M$ action trajectories of length $N$.}
\State $\bm{A}'_{M,[0-\bar{\omega}:N-1]} \gets \bf{0}_{M,N+\bar{\omega}}$ \Comment{This holds  $M$ action trajectories of length $N$ that are perturbed by noise.}
\State $\bm{\varepsilon}_{M,N+\bar{\omega}} \gets \bf{0}_{M,N+\bar{\omega}}$ \Comment{This holds the generated scaled action noise.}
\State
\State $\bm{A}_{M,[0-\bar{\omega}:N-1]} \gets \bf{T}_{[1-\bar{\omega}:N]} / \bm{a}_{\text{max}}$ \Comment{Broadcast previous scaled down action trajectory $\bm{T}$ to $M$ roll-outs.}
\For{$m=0\dots M-1$} \Comment{Sample $M$ trajectories over the horizon $N$.}
\State $\bm{x}_0 \gets \bm{x}$
\For{$n=0\dots N-1$}
\State $\bm{\varepsilon}_{m,n} \gets \mathcal{N}(\bm{0},\bm{\Sigma})$ \Comment{Sample action noise.}
\State $\bm{A}'_{m,n} \gets \bm{A}_{m,n} + \bm{\epsilon}_{m,n}$ \Comment{Perturb action by noise.}
\State $\bm{A}'_{m,n} \gets \min(\max(\bm{A}'_{m,n},-1),+1) $ \Comment{Clip normalized perturbed noise (to bound actions to their limits).}
\State $\bm{\varepsilon}_{m,n} \gets \bm{A}'_{m,n} - \bm{A}_{m,n}$ \Comment{Update noise after bounding, so we do not  penalize clipped noise.}
\EndFor
\For{$n=0\dots N-1$}
\State $\bm{x}_{n+1} \gets F(\bm{x}_n, \bm{A}'_{[n-\bar{\omega}:n]} \cdot \bm{a}_{\max})$ \Comment{Sample next state from pre-trained dynamics model $F$.}
\State $\bm{R}_m \gets \bm{R}_m + r(\bm{x}_n,\bm{A}_{m,n}) - \lambda \bm{A}_{m,n}^{T} \Sigma^{-1}  \bm{\varepsilon}_{m,n}$ \Comment{Accumulate the current state reward.}
\EndFor
\EndFor
\State $\kappa \gets \min_m[\bm{R}_m]$
\State $\bm{T}'_n = \bm{T}_n + \frac{\sum_{m=0}^{M-1}\exp(\frac{1}{\lambda}(\bm{R}_m - \kappa))\bm{\varepsilon}_{m,n}}{\sum_{m=0}^{M-1}(\frac{1}{\lambda}(\bm{R}_m - \kappa))} \cdot \bm{a}_{\max}, \forall n \in [0,N-1]$ \Comment{Generate the return-weighted trajectory update.}
\State {\bfseries Return:} $\bm{T}'$
\end{algorithmic}
\end{algorithm}

\section{ENVIRONMENT SELECTION AND DETAILS}
\label{EnviromentSelectionanddetails}

In the following we discuss our reasoning for why we selected the delay $\tau>0$ adapted continuous-time control environments from the ODE-RL suite \citep{yildiz2021continuous} \footnote{The ODE-RL suite of the environments used can be downloaded freely available from \url{https://github.com/cagatayyildiz/oderl}.} and the reasoning behind our choice of sampling irregularly in time $\Delta_i \neq \Delta_j$ of state-action trajectories $(\bm{x}(t+\Delta_i), \bm{a}(t+\Delta_i))$ from these environments. We first outline why existing environment offline datasets are not suitable.

\paragraph{Why we cannot use an offline dataset of agent trajectories in an un-delayed discrete-time environment}
It is straight-forward, and there exists standard datasets \citep{fu2020d4rl} of state-action trajectories of (expert) agents interacting with environments that have no delays $\tau=0$ and are sampled at regular times $\Delta_i=\Delta_j$. In the following we outline why these datasets cannot be used:
\begin{itemize}
    \item \textbf{Presence of a constant delay $\tau>0$ in the environment.} Intuitively one might suggest taking a standard dataset and shift the actions by a fixed delay. However, we note this dataset is then unrealistic, as it violates causal information---as a hypothetical action
    would know how its current action affected future states before it had even observed them. Due to this fact, this prevents us from using existing standard offline datasets \citep{fu2020d4rl}. Thus, this motivates the need to sample agents that interact within an environment that has an inherent \textit{delay} of a constant delayed action (or constant delayed state observation).
    \item \textbf{Presence of irregularly sampled in time $\Delta_i \neq \Delta_j$ state-action trajectories.} Let us hypothetically imagine that there exists a regularly-sampled $\Delta_i = \Delta_j$ in time state-action trajectory offline dataset from an environment that has an inherent constant action delay $\tau > 0$. Can we then sample them irregularly? One could suppose that we use some form of interpolation (e.g., splines or similar) to interpolate to irregular time steps between states and actions, however doing so would lead to errors in the sampled state-action trajectories in comparison to the true irregularly-sampled state-action trajectories at those non-uniform time points. These errors could compound over a dynamics model being trained on these; therefore, we highlight that this approach is unsuitable. Another approach would be to start with a regularly sampled state-action trajectory $t \in \mathcal{T} \in \{0, \bar{\Delta}, 2\bar{\Delta}, \dots, N\bar{\Delta}\}$ then sub-sample state-action times from that regular grid of collected times $t \subset \mathcal{T}$. Again we indicate this approach unsuitable, as often environments are captured at run-time with a particular observation interval $\bar{\Delta}$ seconds, and only observing multiples of this would mean gaps between observations, where the mean of the observation intervals is larger than that of the environments nominal run-time observation interval $\bar{\Delta}_{\text{Sub-sample}} > \bar{\Delta}_{\text{Original Trajectory}}$. We note that this becomes a different problem, and as there is less information in the state-action trajectories with large observation interval gaps. Instead to mitigate both of these issues we prefer to collect an offline dataset ourselves of an agent interacting with the delay environments with true irregular observation intervals, where we sample the time interval to the next observation from an exponential distribution, i.e.,  $\Delta \sim \text{Exp}(\bar{\Delta})$, with a mean of $\bar{\Delta}=0.05$ seconds.
\end{itemize}

Given the above reasoning, we take the approach to collect offline datasets of a noisy agent interacting with the delay environments, where the observations occur at irregular time intervals given by $\Delta \sim \text{Exp}(\bar{\Delta})$, with a mean of $\bar{\Delta}=0.05$ seconds. We note this approach is similar to \citet{fu2020d4rl} and provides a more realistic offline dataset to train our dynamics models on.

We use the continuous-time control environments from the ODE-RL suite \citep{yildiz2021continuous}, as they provide true irregular samples in time of state observations and are fully continuous in time, unlike discrete environments \citep{brockman2016openai}. We adapt these to incorporate an arbitrary fixed delayed action time, turning the ODE environments into delay DE environments. We do this by keeping a buffer of the previous actions that have been produced by the policy, which captures the previous actions generated in the past $\omega=4\bar{\Delta}$ seconds. In practice this buffer is 4 action elements long and is fed into the dynamics model in its entirety. The true environment then executes the action at the previous time of the constant action delay, which is one of $\tau=\{0,\bar{\Delta},2\bar{\Delta},3\bar{\Delta}\}$ and is \textit{unknown} to the dynamics model. Thus, the dynamics model sees the entire action buffer and must instead learn implicitly the delay of the environment by modeling the dynamics of the environment accurately.

The starting state for all tasks is hanging down and the goal is to swing up and stabilize the pole(s) upright \citep{yildiz2021continuous} in each environment. 
In all environments the actions are continuous and bounded to a defined range $[\bm{a}_{\min},\bm{a}_{\max}$.
Here we assume a given state $\bm{x}(t)$ is composed of the position state $\bm{q}$ and their respective velocities $\dot{\bm{q}}$, i.e., $\bm{x}(t)=\{\bm{q}(t),\bm{\dot{q}}(t)\}$. 
Here, each environment uses the reward function of the exponential of the negative distance from the current state to the goal state $\bm{q}^*$, whilst also penalizing the magnitude of action, and we assume that we are given this reward function when planning---as we often know the desired goal state $\bm{q}^*$ and our current state $\bm{q}$.
Therefore, the reward function for the environments has the following form:
\begin{align}
    r(\{\bm{q}(t),\bm{\dot{q}}(t)\},\bm{a}(t)) = \exp(-||\bm{q}(t) - \bm{q}^*||^2_2 - b||\bm{\dot{q}}(t)||^2_2) - c ||\bm{a}(t)||^2_2
\end{align}
Where $b$ and $c$ are specific environment constants \citep{yildiz2021continuous}.
Specifically, when we use our MPC planner we observe that it plans better without the exponential operator, therefore remove it, and use the following reward function throughout, $r(\{\bm{q}(t),\bm{\dot{q}}(t)\},\bm{a}(t)) = -||\bm{q}(t) - \bm{q}^*||^2_2 - b||\bm{\dot{q}}(t)||^2_2 - c ||\bm{a}(t)||^2_2$.
\citet{yildiz2021continuous} set the environments parameters of $b,c$ to penalize large values and enforce exploration from trivial states, and we use their same values which are also tabulated in Table \ref{table:env_details}.

\begin{table*}[!htb]
\centering
\caption[]{Environment specification parameters of the ODE-RL suite \citep{yildiz2021continuous}.}
\begin{tabular}{cccccc}
\toprule
Base Environment & $b$ & $c$ & $\bm{a}_{\max}$ & $\bm{x}_{\text{Init}}$ & $\bm{q}^*$ \\
\midrule
Pendulum & $1e-2$ & $1e-2$ & $[2]$ & $[0.1,0.1]$ & $[0,L]$ \\
Cartpole & $1e-2$ & $1e-2$ & $[3]$ & $[0.05, 0.05, 0.05, 0.05]$ & $[0,0,L]$ \\
Acrobot & $1e-4$ & $1e-2$ & $[4,4]$ & $[0.1, 0.1, 0.1, 0.1]$ & $[0,2L]$ \\
\bottomrule
\end{tabular}
\label{table:env_details}
\end{table*}

The goal states for all environments is when the poles, each of length $L$ are fully upright, such that their $x,y$ co-ordinates of the tip of the pole reach the goal state. Where $\bm{q}^*$ is: $[0,L]$ for the Pendulum environment, $[0,0,L]$ for the Cartpole environment---where the additional 0 is zero for the cart's $x$ location and in Acrobot is $[0,2L]$ as there are two poles connected to each other. Furthermore, upon restarting the environment the initial state $x$ is sampled from the uniform distribution of $x_0 \sim \mathcal{U}[-\bm{x}_{\text{Init}},\bm{x}_{\text{Init}}]$ \citep{yildiz2021continuous}, then the $\theta$ states are added with set angle such that the pole(s) are pointing downwards (i.e., Cartpole $\theta'=\theta_{\text{Init}}+\pi$).

We note that existing offline RL methods have only been developed for discrete-time $\Delta_i = \Delta_j$ settings and environments that do not possess a delay dynamics $\tau=0$ \citep{argenson2020model}, therefore they are not applicable---instead we opt to train an environment dynamics model and use a planner to select the next action to take.

In the following we describe each of our environments introducing the vanilla environment with a screenshot figure.

\subsection{Cartpole (swing up) Environment}

\begin{figure}[!htb]
\begin{center}
      \includegraphics[width=\textwidth]{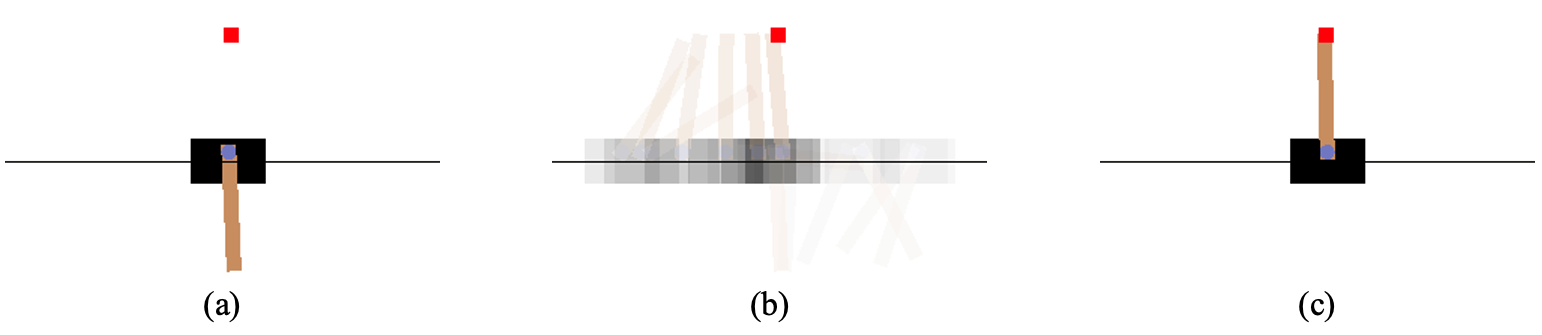}
     \end{center}
  \caption{Screen shots of the Cartpole environment. The task is to swing up a pole attached to a cart that can move horizontally along a rail. In the following we see: (a) the starting downward state with an additional small amount of perturbation, (b) the optimal trajectory solution found by a policy that scores $\mathcal{R}=100\%$ including our NLC model and an expert + MPC and (c) the final goal state that has been reached, that is, to swing up the pole and stabilize it upwards---which is a challenging control task. We note that the control actuator is bounded and limited, and the force is such that the Cartpole cannot be directly swung up---rather it must gain momentum through a swing and then stabilize this swing to not overshoot when stabilizing the pole upwards in the goal position, as indicated when the tip of the pole reaches the centre of the red target square. Furthermore, we note this environment is an underactuated system.}
    \label{CartpoleEnv}
\end{figure}

We can see in Figure \ref{CartpoleEnv}, an illustration of the starting state Figure \ref{CartpoleEnv} (a) with a small perturbed random initial start. Here a pole is attached to an un-actuated joint to a cart that moves along a frictionless track \citep{barto1983neuronlike}. The pendulum starts in the downward position Figure \ref{CartpoleEnv} (a) and the goal is to swing the pendulum upwards and then balance the pole upright by applying forces to the left or right horizontal direction of the cart. This environment has the state of $[x,\dot{x},\theta, \dot{\theta}]$ and a corresponding observation of $[x,\dot{x},\cos(\theta),\sin(\theta),\dot{\theta}]$, where $\theta \in (-\pi,\pi)$ is measured from the upward vertical of the pole. We note that this environment is an underactuated system, as it has two degrees of freedom $[x,\theta]$, however only the carts position is actuated, leaving $\theta$ indirectly controlled.

\subsection{Pendulum Environment}

\begin{figure}[!htb]
\begin{center}
      \includegraphics[width=\textwidth]{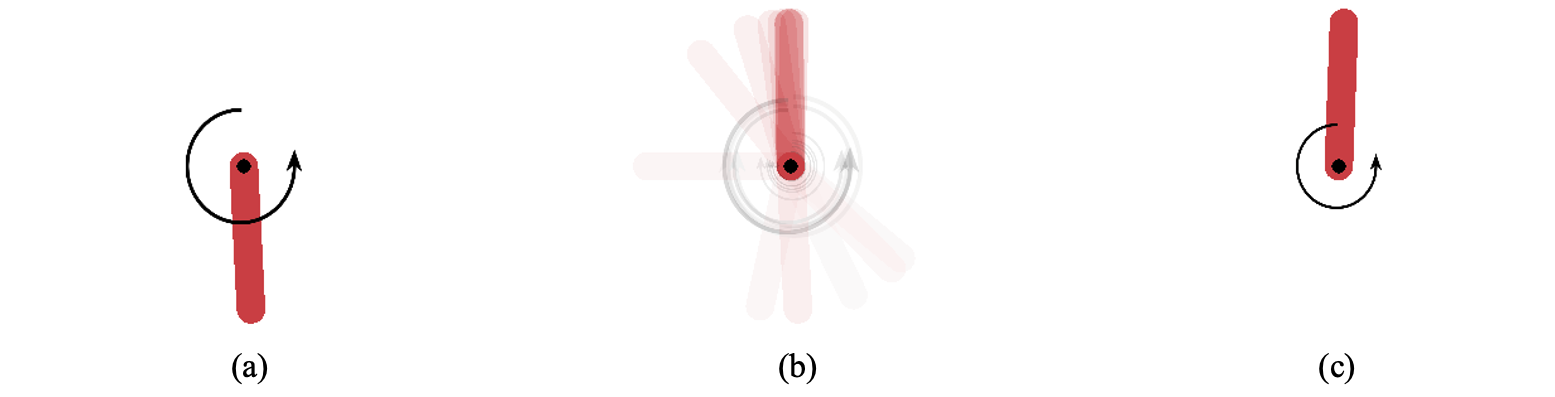}
     \end{center}
  \caption{Screen shots of the Pendulum environment. The task is to swing up the pole (pendulum). In the following we see: (a) starting downward state with an additional small amount of perturbation, (b) the optimal trajectory solution found by a policy that scores $\mathcal{R}=100\%$ including our NLC model and an expert + MPC and (c) the final goal state that has been reached, that is, to swing up the pole and stabilize it upwards. We note that the control actuator is bounded and limited, and the force is such that the Pendulum cannot be directly swung up---rather it must gain momentum through a swing and then stabilize this swing to not overshoot when stabilizing the pole upwards in the goal position.}
    \label{pendulum_env}
\end{figure}

We can see in Figure \ref{pendulum_env}, an illustration of the starting state Figure \ref{pendulum_env} (a) with a small perturbed random initial start. Here a pole (pendulum) is attached to a fixed point at one end with the other end being free \citep{barto1983neuronlike, yildiz2021continuous}. The pendulum starts in the downward position Figure \ref{pendulum_env} (a) and the goal is to swing the pendulum upwards and then balance the pole upright by applying torques about the fixed point, as indicated in the Figure \ref{pendulum_env} with a visualisation showing the torque direction and magnitude based on the size of the arrow. This environment has the state of $[\theta,\dot{\theta}]$ and a corresponding observation of $[\sin(\theta),\cos(\theta),\dot{\theta}]$.

\subsection{Acrobot Environment}

\begin{figure}[!htb]
\begin{center}
      \includegraphics[width=\textwidth]{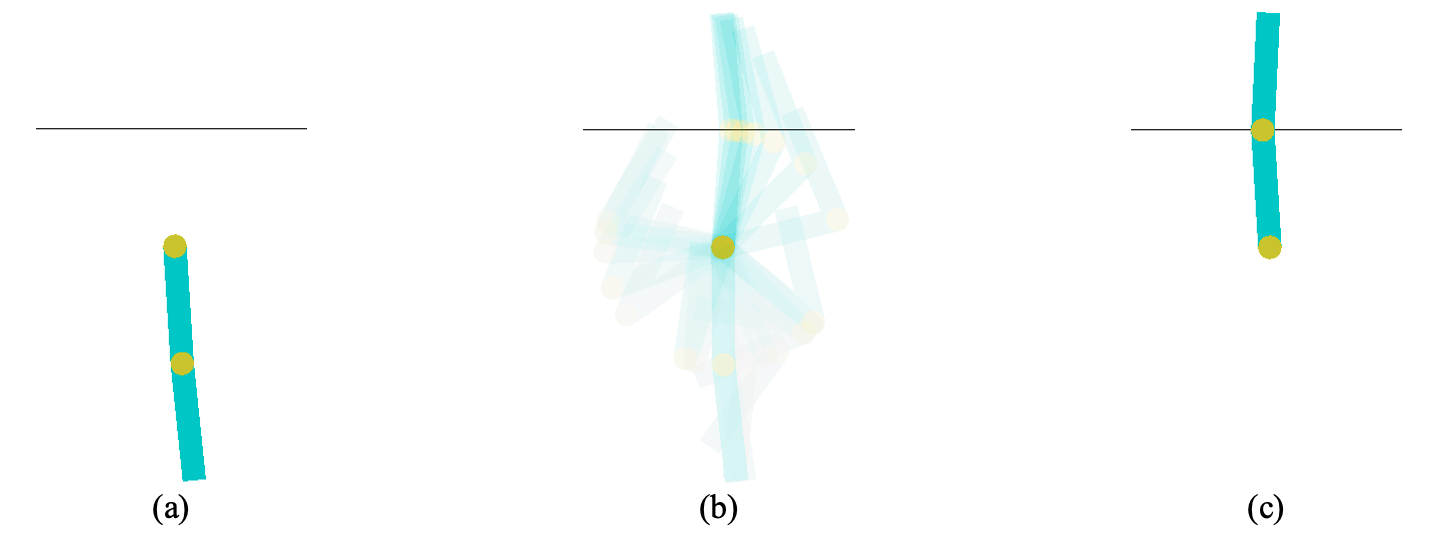}
     \end{center}
  \caption{Screen shots of the Acrobot environment. The task is to swing up the 2-link pendulum. In the following we see: (a) starting downward state with an additional small amount of perturbation, (b) the optimal trajectory solution found by a policy that scores $\mathcal{R}=100\%$ including our NLC model and an expert + MPC and (c) the final goal state that has been reached, that is, to swing up the 2-link pendulum and stabilize it upwards. We note that the control actuator is bounded and limited, and the force is such that the 2-link pendulum cannot be directly swung up---rather it must gain momentum through a 2-link swing and then stabilize this swing to not overshoot when stabilizing the 2-link pendulum upwards in the goal position.}
    \label{acrobot_env}
\end{figure}

We can see in Figure \ref{acrobot_env}, an illustration of the starting state Figure \ref{acrobot_env} (a) with a small perturbed random initial start. It is a 2-link pendulum with the individual joints actuated \citep{brockman2016openai}. The 2-link pole starts in the downward position Figure \ref{acrobot_env} (a) and the goal is to swing the 2-link pendulum upwards and then balance the pole(s) upright by applying torques about their fixed points. This environment has the state of $[\theta_1,\dot{\theta}_1,\theta_2,\dot{\theta}_2]$ and a corresponding observation of $[\sin(\theta_1),\cos(\theta_1),\dot{\theta}_1,\sin(\theta_2),\cos(\theta_2),\dot{\theta}_2]$. Here the Acrobot environment is fully actuated, as no method has been able to solve the underactuated balancing problem \citep{yildiz2021continuous, zhong2019symplectic}.

\section{BENCHMARK METHOD IMPLEMENTATION DETAILS}
\label{benchmarkmethodimplementationdetails}

\begin{table}[!htb]
\centering
  \caption[]{Benchmark dynamic models implemented and their number of parameters for each model.}
  \label{benchmarkparams}
  \begin{tabular}{l c }
    \toprule
    Dynamics model & \# Parameters \\
    \midrule
$\Delta t-$RNN	& 79,075 \\
NODE	 & 76,956 \\
Latent-ODE	& 76,453 \\
Neural Laplace Control & 81,772 \\
    \bottomrule
  \end{tabular}
\end{table}

We tuned all the baseline dynamics models to have the same approximate number of parameters, as can be seen in Table \ref{benchmarkparams}---to ensure fair comparison for any gains in modeling complexity. 
We train and evaluate all the dynamics models and all data used with double point floating precision, as is recommended when using an inverse Laplace transform (ILT) to aid the ILT stability \citep{holt2022neural, kuhlman2012}. Further all dynamics models are implemented in PyTorch \citep{NEURIPS2019_9015}, and trained with an Adam optimizer \citep{kingma2017adam} with a learning rate of 1e-4.
Each baseline dynamics model are:

\paragraph{Discrete-delay method}
We implemented the discrete-delay method similar to \citet{chen2021delay}, a RNN over the action buffer and current state, and adapt it to model continuous-time with a new input of the time increment to predict the next state for (\textbf{$\Delta t-$RNN}). We note to adapt discrete-time models to continuous-time we add an additional input parameter, that of the time difference between the current time and the next state observation to predict, i.e., $\delta$, e.g., $\bm{x}_{i+1}=\bm{x}_i+\bm{f}(\bm{x}_i,\bm{a}_i,\delta)$.
\citep{yildiz2021continuous}. Specifically, we feed the action buffer $\mathcal{H}_i$ into a gated recurrent neural (GRU) network with a hidden size of 160 features in the hidden state and feed the final hidden state and concatenate it with the current observed state and time increment to predict the next state which is all input into a linear layer to produce the output state prediction.

\paragraph{Continuous-time methods}
We implemented an augmented Neural-ODE (\textbf{NODE}) \citep{chen2018neural}, using their code and corresponding implementation provided, and set their ODE function $\bm{f}(t, \bm{x}(t), \bm{a}(t))$ to be a 3-layer multilayer perceptron (MLP), of 270 units, with $\tanh$ activation functions---with an additional augmented dimension of zeros. As neural-ODE does not have an encoder, we feed the most recent action taken at time $t$, i.e., $\bm{a}(t)$ into the MLP $\bm{f}$ function instead. Further, to allow for fair comparison we use the semi-norm trick for faster back propagation \cite{DBLP:journals/corr/abs-2009-09457}, and use the 'euler' solver throughout. We also use the reconstruction MSE for training.

We also compare with \textbf{Latent-ODE} \citep{DBLP:journals/corr/abs-1907-03907}, which uses an ODE-RNN encoder and an ODE model decoder. We feed this the action history buffer $\mathcal{H}_i$ concatenated with an equivalent state history buffer $\mathcal{H}'_i=\{(\bm{x}_j,t_j-t_i):t_j\in[t_i-\omega,t_i]\}$ for the same sample times.
We use their code provided, setting the units to be 128 for the GRU and ODE function $\bm{f}(t, \bm{x}(t), \bm{a}(t))$ net, with $\tanh$ activation functions and use the 'dopri5' solver. We also use their reconstruction variational loss function for training.

\paragraph{Neural Laplace Control}
This paper uses a GRU encoder $h_\zeta$ \citep{cho2014properties}, to encode the action buffer $\mathcal{H}_i$ with 2 layers and a hidden size of 64 features in the hidden state, with a final linear layer on the final hidden to output $\bm{p}_i^{(\mathcal{A})}$. We do not encode the state and instead feed it directly as $\bm{p}_i^{(\mathcal{X})}=\bm{x}_i$. Therefore, we encode both into a latent dimension $\bm{p}_i=(\bm{p}_i^{(\mathcal{A})},\bm{p}_i^{(\mathcal{X})})$.
For the Laplace representation model, $g_{\psi}$ we use a 3-layer MLP with 128 units, with $\tanh$ activations. As recommended by \citet{holt2022neural} we further use a $\tanh$ on the output to constrain the output domain to be $(\theta, \phi) \in \mathcal{D} = (-{\pi}, {\pi}) \times (-\frac{\pi}{2}, \frac{\pi}{2})$ for each output state prediction. For a given state prediction, we encode the action buffer and state into $\bm{p}_i$ and concatenate this with $u(\bm{s})$ as the input to $g_{\psi}$, i.e., $g_\psi \big(\bm{p}, u(\bm{s}))$, where $\bm{s}$ is given by the ILT algorithm for a future time point to predict the state for. Specifically, we use the Fourier-series inverse algorithm (ILT-FSI), with $d_{\mathcal{S}} = 17$ reconstruction terms. Where we use the specific ILT-FSI from \citet{holt2022neural} and use their code implementation of the ILT algorithm.

\paragraph{MPPI Implementation}
We use the MPPI algorithm, with pseudocode and is further described in Appendix \ref{mppipseudocode}.
Specifically, as is recommended by \citet{lutter2021learning} we optimized the MPPI hyperparameters through a grid search with the true (Oracle) dynamics model for a single environment setting, that of the Cartpole environment with a delay of $\tau=\bar{\Delta}=0.05$ seconds, and fix these for planning with all the learned dynamics models throughout. Specifically, our final optimized hyperparameter combination is $N=40, M=1,000, \lambda=1.0, \sigma=1.0$. Where $\Sigma$ the MPPI action noise is defined as:
\begin{align}
    \Sigma = \begin{dcases}
        [\sigma^2] & \text{if}~ d_{\mathcal{A}} = 1 \\
        \begin{bmatrix}
\sigma^2 & 0.5 \sigma^2 \\
0.5 \sigma^2 & \sigma^2
\end{bmatrix} & \text{if}~ d_{\mathcal{A}} = 2 \\
    \end{dcases}
\end{align}
Where the Cartpole and Pendulum environments have $d_{\mathcal{A}}=1$ and Acrobot environment has $d_{\mathcal{A}}=2$. These hyperparameters were found by searching over a grid of possible values, which is detailed in Table \ref{table:hyperparametergridsearch}.

\begin{table*}[!htb]
\centering
\caption[]{MPPI hyperparameter grid search sweep values.}
\begin{tabular}{cc}
\toprule
Hyperparameter & Grid values searched over \\
\midrule
$N$ & $\{1,2,4,8,16,20,40,50,60,70,80,90,100,128,256,512,1024,2048\}$ \\
$M$ & $\{1,2,4,8,16,20,40,50,60,70,80,90,100,128,256,500,1000,2000,4000,8000\}$ \\
$\lambda$ & $\{0.00001, 0.0001, 0.001, 0.01, 0.1, 0.5, 0.8, 1.0, 1.5, 2.0, 10.0, 100.0, 1000.0\}$ \\
$\sigma$ & $\{0.00001, 0.0001, 0.001, 0.01, 0.1, 0.5, 0.8, 1.0, 1.5, 2.0, 10.0, 100.0, 1000.0\}$ \\
\bottomrule
\end{tabular}
\label{table:hyperparametergridsearch}
\end{table*}

\section{EVALUATION METRICS}
\label{EvaluationMetrics}

For each environment, with a different delay setting we collect an offline dataset of irregularly-sampled in time trajectories, consisting of 1e6 samples from the \quotes{noisy expert} agent interacting within that environment (Section 5 \& Appendix \ref{datasetgeneationandmodeltraining}). For each benchmark dynamics model, we follow the same two step evaluation process of, firstly, training the dynamics model on that environment's collected offline dataset using a MSE error loss for the next step ahead state prediction $\hat{\bm{x}}(t_{i+1})$. Then, secondly, taking the same pre-trained model and freezing the weights, and only using it for planning with the MPPI (MPC) planner at run-time in an environment episode, that lasts for $10$ seconds.
In total, we evaluate our model-based control algorithms online in the same environment, running each one for a fixed observation interval of $\delta=\bar{\Delta} = 0.05$ seconds (as is the nominal value for these environments \citep{yildiz2021continuous, brockman2016openai}, unless specified otherwise), and take the cumulative reward value after running one episode of the planner (policy) and repeat this for 20 random seed runs for each result.
We quote the normalized score $\mathcal{R}$ \citep{yu2020mopo} of the policy in the environment, averaged over the 20 random seed run episodes, with standard deviations throughout.
The scores are un-discounted cumulative rewards normalized to lie roughly between 0 and 100, where a score of 0 corresponds to a random policy, and 100 corresponds to an expert (oracle with a MPC planner).
Specifically, when we evaluate the insights experiments in Section 5.2, where we change the planning observation interval $\delta$, this changes the number of steps taken in an episode, therefore we quote the un-discounted average rewards normalized to lie roughly between 0 and 100, where a score of 0 corresponds to a random policy, and 100 corresponds to an expert (oracle with a MPC planner). We note that a normalized cumulative reward for an episode and an average reward for an episode for the same number of steps in an episode are equivalent.
Furthermore, we also track the metric of total planning time taken to plan the next action as $\mathcal{O}$ seconds and perform all experiments using a single Intel Core i9-12900K CPU @ 3.20GHz, 64GB RAM with a Nvidia RTX3090 GPU 24GB.

\section{DATASET GENERATION AND MODEL TRAINING}
\label{datasetgeneationandmodeltraining}

For each environment we generate an offline state-action trajectory dataset by using an agent that uses an oracle dynamics model combined with MPC and has additional noise added to the agents selected action, $\bar{\pi}(t) = \pi(t) + \epsilon, \epsilon \sim \mathcal{N}(0,\bm{a}_{\max})$. This \quotes{noisy expert} agent interacts with the environment and observes observations at irregular unknown times, where we sample the time interval to the next observation from an exponential distribution, i.e.,  $\Delta \sim \text{Exp}(\bar{\Delta})$, with a mean of $\bar{\Delta}=0.05$ seconds. We note other irregular sampling types are possible, however \citet{yildiz2021continuous} has shown they are approximately equivalent.
We assume a fixed action delay $\tau$, and evaluate discrete multiples of this delay of the mean sampling time $\bar{\Delta}$, i.e., $\tau=\bar{\Delta}$ for one step delay, $\tau=2\bar{\Delta}$ for two step delay etc.
We enforce the observed action history buffer that includes past actions back to $\omega = 4\bar{\Delta}$ seconds. For each specific delay version of each base environment class (Cartpole, Pendulum and Acrobot) we collect a total of 1e6 irregular state-action samples in time (unless otherwise specified).
Using the whole collected dataset we pre-process this by a standardization step, to make each dimension of the samples have zero mean and unit variance (by taking away the mean for each dimension and then dividing by the standard deviation for each dimension)---we also use this step during run-time for each dynamics model. Furthermore, we train all the baseline models on all the samples collected in the offline dataset (all samples are training data), training the models parameters with the Adam optimizer \citep{kingma2017adam} with a learning rate of 1e-4 throughout.
Specifically, we train all the baseline dynamics models on a given offline dataset by training each model for 2 hours and 15 minutes. We also ran a further experiment where we trained all the models for a fixed number of epochs instead, detailed further in Appendix \ref{Trainingdynamicsmodelsforafixednumberofepochs}. For the insights experiments in Section 5.2 that quote a validation dataset error, we achieve this by generating a new offline dataset using the same setup described above with a different random seed to use as a validation dataset.

\section{RAW RESULTS}
\label{rawresults}

The full results from Table 2, are in Table \ref{table:main_normalized_scores_all_results_table_2} along with their un-normalized versions in Table \ref{table:main_unormalized_scores_all_results_table_2}.

\paragraph{High variance in Latent-ODE and NODE}
We detail their poor performance to: (1) both these models do not support batches of trajectories evaluated at different non-uniform time points---therefore, they are trained with a batch size of 1, (2) we only train a single dynamics model, unlike other works \citep{yildiz2021continuous} that train an ensemble of models, and (3) Latent-ODE is trained using the recommended variational loss of the next step ahead prediction.

\begin{table*}[!htb]
\caption[]{Normalized scores $\mathcal{R}$ of the offline model-based agents, where the irregularly-sampled (P1) offline dataset consists of an action delay (P2) of $\{0,1,2,3\}$ multiples of the environments observation interval time step $\bar{\Delta} = 0.05$ seconds. Averaged over 20 random seeds, with $\pm$ standard deviations.
Scores are un-discounted cumulative rewards normalized to be between 0 and 100, where 0 corresponds to the Random agent and 100 corresponds to the expert with the \textit{known} world model (oracle+MPC). Negative normalized scores, i.e., worse than random are set to zero.}
\resizebox{\textwidth}{!}{
\begin{tabular}{@{}l|ccc|ccc|ccc|ccc}
\toprule
                                &  \multicolumn{3}{c|}{Action Delay~$\tau=0$}     & \multicolumn{3}{c|}{Action Delay~$\tau=\bar{\Delta}$}      & \multicolumn{3}{c|}{Action Delay~$\tau=2\bar{\Delta}$}      &  \multicolumn{3}{c}{Action Delay~$\tau=3\bar{\Delta}$}               \\
        Dynamics Model                  &     Cartpole & Pendulum & Acrobot & Cartpole & Pendulum & Acrobot                   & Cartpole & Pendulum & Acrobot                     & Cartpole & Pendulum & Acrobot \\        
\midrule
Random              &     0.0$\pm$0.0 &      0.0$\pm$0.0 &      0.0$\pm$0.0 &     0.0$\pm$0.0 &      0.0$\pm$0.0 &      0.0$\pm$0.0 &      0.0$\pm$0.0 &        0.0$\pm$0.0 &       0.0$\pm$0.0 &     0.0$\pm$0.0 &       0.0$\pm$0.0 &      0.0$\pm$0.0 \\
Oracle              &     100.0$\pm$0.04 &     100.0$\pm$3.34 &    100.0$\pm$1.84 &     100.0$\pm$0.15 &     100.0$\pm$3.14 &    100.0$\pm$2.19 &     100.0$\pm$0.04 &     100.0$\pm$2.57 &    100.0$\pm$1.79 &     100.0$\pm$0.08 &     100.0$\pm$2.57 &    100.0$\pm$1.26 \\
$\Delta t-$RNN      &     96.76$\pm$0.34 &     32.73$\pm$7.09 &    12.61$\pm$4.65 &      95.28$\pm$0.4 &      1.14$\pm$6.31 &     18.95$\pm$7.6 &     97.01$\pm$0.31 &      9.94$\pm$2.48 &    28.39$\pm$9.73 &      97.8$\pm$0.25 &    11.81$\pm$11.93 &     3.89$\pm$6.72 \\
Latent-ODE          &        0.0$\pm$0.0 &      8.08$\pm$8.45 &     4.45$\pm$8.81 &        0.0$\pm$0.0 &        0.0$\pm$0.0 &       0.0$\pm$0.0 &        0.0$\pm$0.0 &     1.24$\pm$20.67 &    8.91$\pm$13.62 &    41.56$\pm$47.07 &     3.26$\pm$12.24 &     9.19$\pm$9.08 \\
NODE                &     70.08$\pm$2.94 &     12.89$\pm$4.19 &       0.0$\pm$0.0 &     85.09$\pm$7.95 &      0.63$\pm$5.16 &    23.07$\pm$6.94 &     90.75$\pm$1.34 &        0.0$\pm$0.0 &   10.92$\pm$10.09 &     94.55$\pm$1.08 &      1.97$\pm$4.01 &    11.78$\pm$8.33 \\
\midrule
\bf NLC \textbf{(Ours)} &      \textbf{99.87$\pm$0.1} &     \textbf{101.52$\pm$3.3} &    \textbf{99.16$\pm$1.91} &     \textbf{99.83$\pm$0.19} &     \textbf{98.31$\pm$3.51} &     \textbf{99.12$\pm$1.7} &      \textbf{99.88$\pm$0.1} &     \textbf{93.28$\pm$4.96} &   \textbf{100.44$\pm$2.13} &     \textbf{99.92$\pm$0.12} &     \textbf{98.98$\pm$1.32} &    \textbf{99.46$\pm$1.88} \\
\bottomrule
\end{tabular}
}
\label{table:main_normalized_scores_all_results_table_2}
\end{table*}

\begin{table*}[!htb]
\caption[]{Un-normalized scores $\mathcal{R}$ of the offline model-based agents, where the irregularly-sampled (P1) offline dataset consists of an action delay (P2) of $\{0,1,2,3\}$ multiples of the environments observation interval time step $\bar{\Delta} = 0.05$ seconds. Averaged over 20 random seeds, with $\pm$ standard deviations.
Scores are un-discounted cumulative rewards.}
\resizebox{\textwidth}{!}{
\begin{tabular}{@{}l|ccc|ccc|ccc|ccc}
\toprule
                                &  \multicolumn{3}{c|}{Action Delay~$\tau=0$}     & \multicolumn{3}{c|}{Action Delay~$\tau=\bar{\Delta}$}      & \multicolumn{3}{c|}{Action Delay~$\tau=2\bar{\Delta}$}      &  \multicolumn{3}{c}{Action Delay~$\tau=3\bar{\Delta}$}               \\
        Dynamics Model                  &     Cartpole & Pendulum & Acrobot & Cartpole & Pendulum & Acrobot                   & Cartpole & Pendulum & Acrobot                     & Cartpole & Pendulum & Acrobot \\        
\midrule
Random              &    -14246.3$\pm$0.0 &  -616.77$\pm$0.0 &  -2948.64$\pm$0.0 &      -9713.19$\pm$0.0 &  -575.98$\pm$0.0 &    -2910.5$\pm$0.0 &    -15097.54$\pm$0.0 &    -584.8$\pm$0.0 &  -2938.69$\pm$0.0 &  -20798.89$\pm$0.0 &  -596.38$\pm$0.0 &  -2885.99$\pm$0.0 \\
Oracle              &         -139.69$\pm$5.27 &  -121.05$\pm$16.54 &     -571.11$\pm$43.7 &         -146.26$\pm$13.95 &   -123.44$\pm$14.2 &    -558.76$\pm$51.41 &          -145.56$\pm$5.26 &  -125.08$\pm$11.79 &     -582.01$\pm$42.1 &       -153.19$\pm$16.72 &  -133.51$\pm$11.88 &    -588.61$\pm$28.84 \\
$\Delta t-$RNN      &        -597.21$\pm$47.95 &  -454.51$\pm$35.13 &  -2648.73$\pm$110.56 &         -598.27$\pm$37.96 &  -570.82$\pm$28.57 &  -2464.87$\pm$178.77 &         -592.66$\pm$46.17 &  -539.12$\pm$11.39 &   -2269.64$\pm$229.3 &       -607.96$\pm$51.11 &  -541.71$\pm$55.22 &  -2796.69$\pm$154.29 \\
Latent-ODE          &  -152488.09$\pm$89951.95 &  -576.69$\pm$41.91 &   -2842.9$\pm$209.42 &  -814206.21$\pm$292756.78 &  -696.61$\pm$71.72 &  -2911.94$\pm$263.25 &  -1539109.96$\pm$288324.7 &  -579.11$\pm$95.01 &   -2728.8$\pm$320.88 &   -12217.77$\pm$9717.18 &  -581.29$\pm$56.67 &  -2674.88$\pm$208.65 \\
NODE                &      -4359.77$\pm$415.24 &  -552.86$\pm$20.78 &   -3048.13$\pm$79.08 &       -1572.47$\pm$760.41 &  -573.12$\pm$23.33 &  -2368.03$\pm$163.25 &       -1528.85$\pm$200.28 &  -613.56$\pm$16.01 &  -2681.39$\pm$237.74 &     -1277.55$\pm$222.61 &  -587.27$\pm$18.57 &  -2615.45$\pm$191.46 \\
\midrule
\bf NLC \textbf{(Ours)} &        \textbf{-158.31$\pm$14.66} &  \textbf{-113.52$\pm$16.35} &    \textbf{-591.16$\pm$45.46} &         \textbf{-162.47$\pm$18.31} &   \textbf{-131.07$\pm$15.9} &    \textbf{-579.39$\pm$39.92} &          \textbf{-163.0$\pm$14.67} &  \textbf{-155.96$\pm$22.82} &     \textbf{-571.64$\pm$50.1} &       \textbf{-170.17$\pm$25.59} &   \textbf{-138.23$\pm$6.13} &    \textbf{-601.04$\pm$43.17} \\
\bottomrule
\end{tabular}
}
\label{table:main_unormalized_scores_all_results_table_2}
\end{table*}

\section{INSIGHT EXPERIMENTS}
\label{insightexperiments}

In this section we seek to gain further insight into \textit{how} Neural Laplace Control outperforms the benchmarks. In the following we seek to understand if NLC is able to learn from irregularly-sampled state-action offline datasets (P1), whilst learning the delayed dynamics of the environment (P2). Furthermore, we also explore the benefits of the NLC approach for planning at longer time horizons with a fixed amount of compute and being sample efficient.

\subsection{Can the baseline dynamics models learn a good model?}

\begin{figure}[!htb]
\begin{center}
    \centering
      \includegraphics[width=\textwidth]{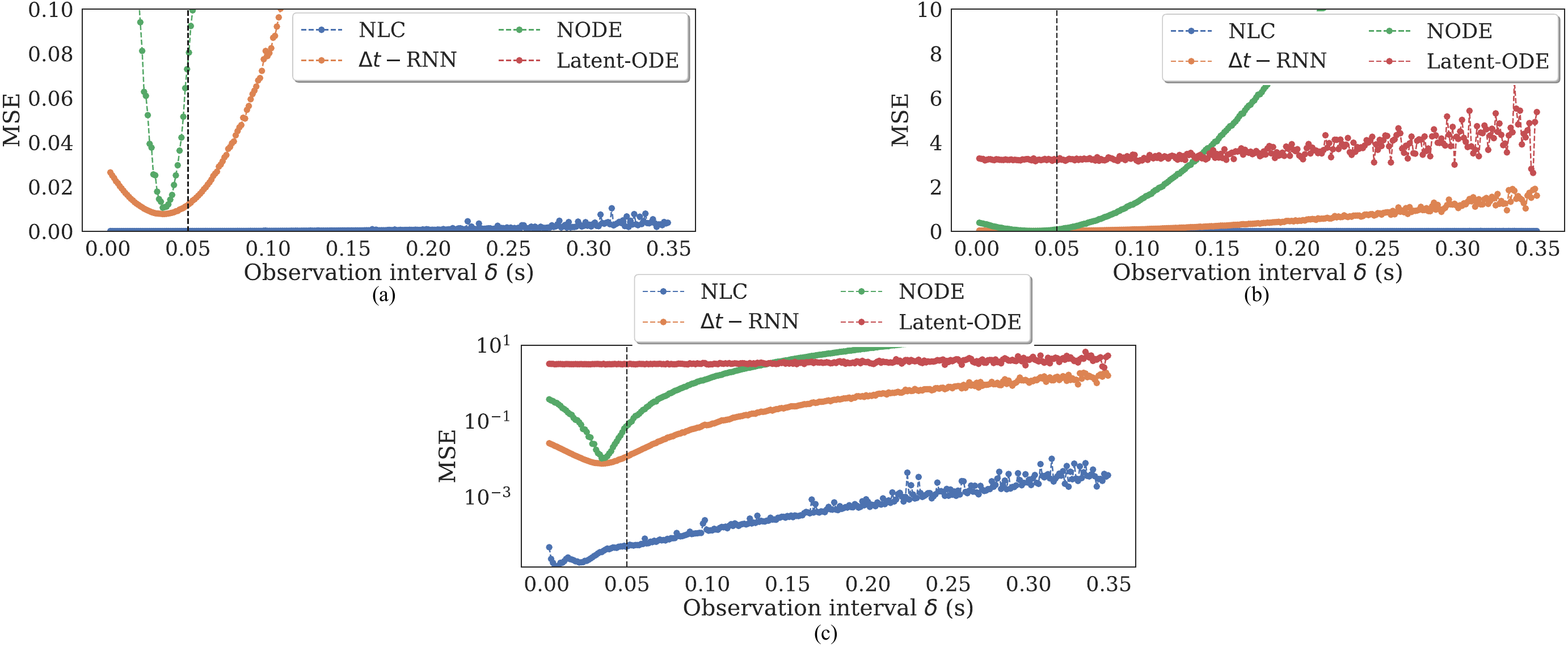}
     \end{center}
  \caption{Next step ahead validation error (MSE) at a variable time step of an observation interval $\delta$ of the learnt baseline dynamics models, for the irregularly-sampled Cartpole environment with a fixed action delay of $\tau=\bar{\Delta}$. Where in each sub-figure we have the same results plotted at: (a) a zoomed-in y-limit, (b) a zoomed-out y-limit and (c) a log scale plot. The black dotted line indicates the environments run-time observation interval $\delta=\bar{\Delta}=0.05$ s. Here, we observe Neural Laplace Control learns a good dynamics model over a wide range of observation intervals $\delta$, correctly learning from the irregularly-sampled offline dataset (P1).}
    \label{model_comp_app_plots}
\end{figure}

To explore if NLC is able to learn a suitable dynamics model, we plot the trained dynamics models next step ahead prediction error with that of the ground truth for a varying observation interval $\delta$ for the Cartpole environment with a delay of $\bar{\Delta}$, as shown in Figure \ref{model_comp_app_plots}. Empirically we observe that NLC using its Laplace-based dynamics model is able to better approximate a wider range of observation intervals $\delta$ and achieve a good \textit{global} approximation compared to the recurrent neural network and ODE based models.
We note that due to the offline dataset being sampled with trajectories that have irregular sampling times (P1), where the sampling times are defined by an exponential distribution with a mean of $\bar{\Delta}=0.05$ seconds; the other competing methods seem to over-fit purely to the median sample time of the exponential distribution, i.e., $0.05 \cdot \ln(2) = 0.034$ s.
Other works have shown a more accurate next step prediction model correlates to a higher environment episode reward \citep{williams2017information}.

\subsection{Can the baseline dynamics models learn delay environment dynamics?}

\begin{figure}[!htb]
\begin{center}
    \centering
      \includegraphics[width=\textwidth]{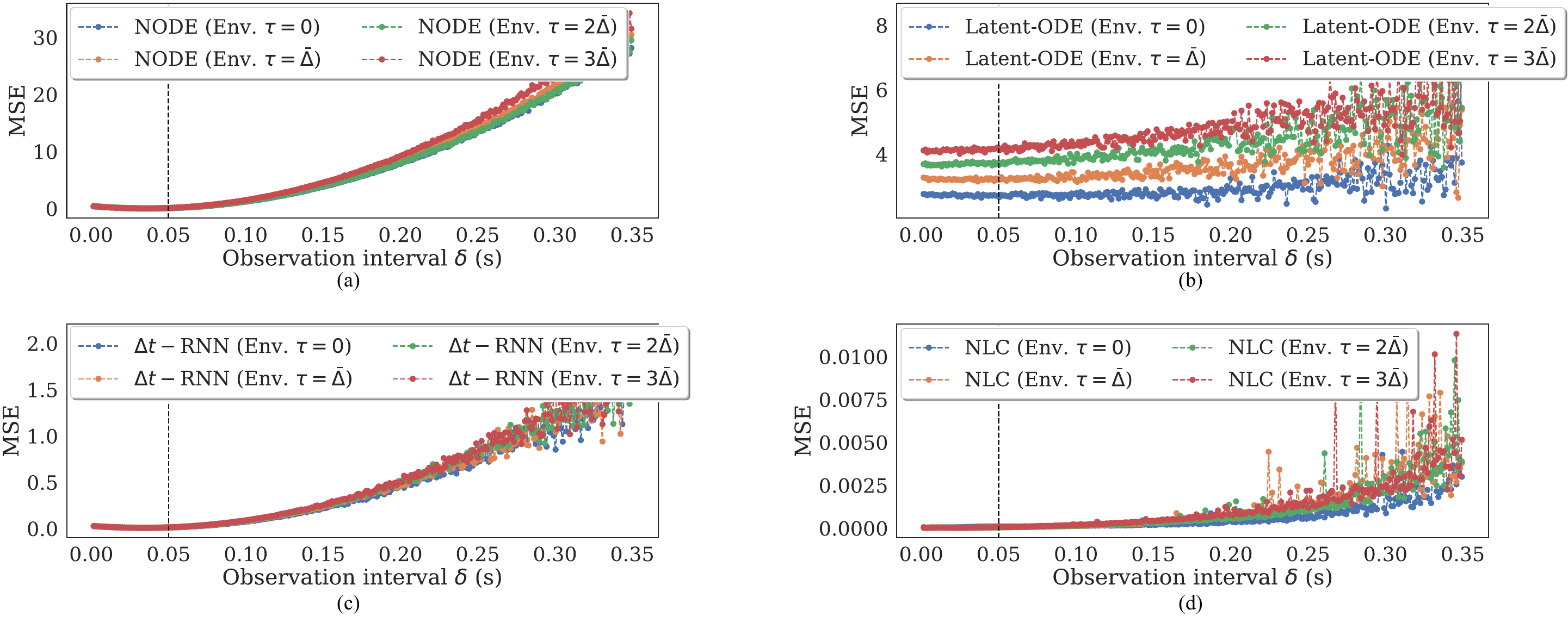}
     \end{center}
  \caption{Next step ahead validation error (MSE) at a variable time step of an observation interval $\delta$ of the learnt baseline dynamics models, for each delayed environment versions $\tau=\{0,\bar{\Delta},2\bar{\Delta},3\bar{\Delta}\}$ of the specific Cartpole environment. Where in each sub-figure we have: (a) NODE, (b) Latent-ODE, (c) $\Delta t-$RNN and (d) NLC. The black dotted line indicates the environments run-time observation interval $\delta=\bar{\Delta}=0.05$ s. Here, Neural Laplace Control is able to correctly learn and capture the delayed dynamics (P2), as the forward MSE errors are low and similar---whereas neural-ODE methods (a) \& (b) have a greater increasing forward MSE.}
    \label{app_delay_plots}
\end{figure}

To investigate this, we similarly plot the trained NLC dynamics models next step ahead prediction error with that of the ground truth for a varying observation interval $\delta$, for each of the delayed environment versions of the specific Cartpole environment, as show in Figure \ref{app_delay_plots} (d). Empirically we observe that the NLC dynamics models correctly learnt the delay dynamics (P2) of each individual environment, as they each have a similar low forward MSE error for the varying levels of inherent delay.
In contrast, neural-ODE models are unable to model the delay dynamics correctly, and we observe that they have a higher rate of increasing forward MSE error Figure \ref{app_delay_plots} (a) \& (b), that can also increase for an increasing environment delay and is shown further in Figure \ref{app_delay_plots} for the Latent-ODE model---intuitively this may occur as increasing the delay in a delay differential equation driven environment dynamics becomes less like that of an ordinary differential equation driven environment dynamics, which neural-ODE methods implicitly assume.

\subsection{Can NLC plan with a longer time horizon using a fixed amount of compute?}

\begin{figure}[!htb]
  \centering
\includegraphics[width=\columnwidth]{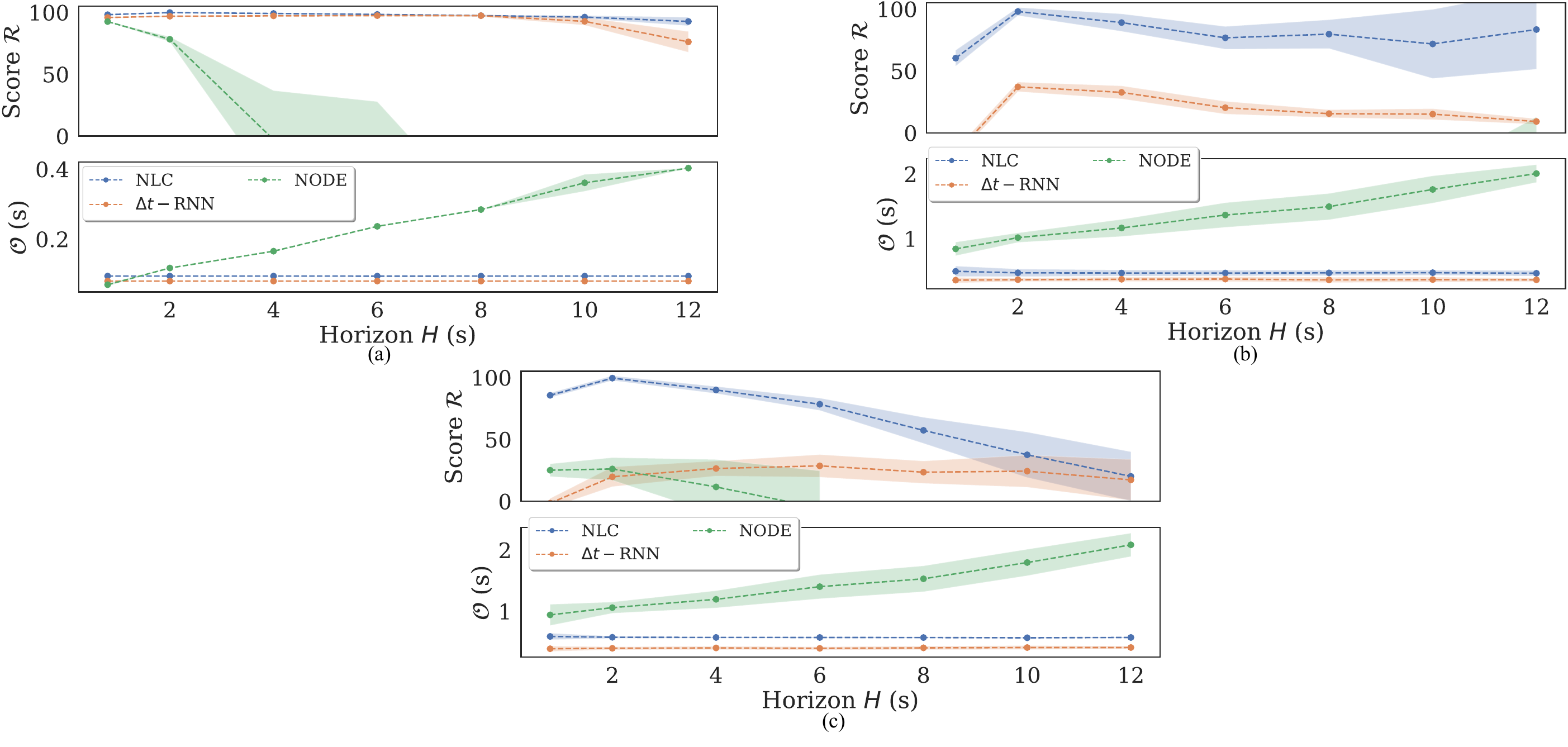}
\caption{Normalized score $\mathcal{R}$ of the baseline methods on the three environments with an action delay of $\tau=\bar{\Delta}=0.05$ seconds, plotted against an increasing time horizon $H$, by increasing the observation interval $\delta$. Specifically in: (a) the Cartpole environment, (b) the Pendulum environment and (c) the Acrobot environment. NLC, maintains a high performing policy at a longer time horizon---whilst using the same amount of \textit{constant} planning time per action $\mathcal{O}$ as a $\Delta t-$RNN.}
\label{Fig:fixed_amount_of_compute_appendix}
\end{figure}

We investigate this by planning with a MPC planner, increasing the observation interval $\delta$ and keeping $N$ fixed, therefore the time horizon $H$ increases---as shown in Figure \ref{Fig:fixed_amount_of_compute}. Here we measure the total planning time taken to plan the next action as $\mathcal{O}$ seconds \footnote{We perform all results using a Intel Core i9-12900K CPU @ 3.20GHz, 64GB RAM with a Nvidia RTX3090 GPU 24GB.} and observe that planning with the NLC dynamics model takes the same amount of planning time, and hence a \textit{fixed amount compute} for planning at a greater time horizon $H$---which is the same as a $\Delta t-$RNN.
This is achieved by the Laplace-based dynamics model that can predict a future state at \textit{any} future time interval using the same number of forward model evaluations, and hence the same amount of compute.
In contrast, this is \textit{not} readily achievable with neural-ODE continuous-time methods that use a larger number of numerical forward steps with a numerical ODE step-wise solver for an increasing time horizon---leading to an increasing planning time for an increasing time horizon, i.e., $\mathcal{O} \propto H$. Furthermore, we highlight, that there exists a trade-off of the time horizon $H$ to plan at---as we wish to use a large \quotes{enough} horizon that captures sufficient future dynamics, whilst minimizing compounded model inaccuracies at a larger planning time horizon. Therefore, these two opposing factors, give rise to the maxima of the normalized score $\mathcal{R}$ at a time horizon $H=2$ seconds, as seen in Figure \ref{Fig:fixed_amount_of_compute_appendix}. The numeric values for each environment are tabulated in Tables \ref{table:longer_time_horizon_same_compute_cartpole}, \ref{table:longer_time_horizon_same_compute_pendulum} \& \ref{table:longer_time_horizon_same_compute_acrobot}.

\begin{table*}[!htb]
\caption[]{Numerical results of the Cartpole environment, of the planning time taken $\mathcal{O}$ to generate the next action, and normalized scores $\mathcal{R}$ of the baseline methods with an environment action delay of $\tau=\bar{\Delta}=0.05$s, varying against an increasing time horizon $H$---by increasing the observation interval $\delta$. NLC, maintains a high performing policy at a longer time horizon---whilst using the same amount of \textit{constant} planning time per action $\mathcal{O}$ as a $\Delta t-$RNN.}
\resizebox{\textwidth}{!}{
\begin{tabular}{ccccccccccccc}
\toprule
&&$H$=0.8 s&$H$=2.0 s&$H$=4.0 s&$H$=6.0 s&$H$=8.0 s&$H$=10.0 s&$H$=12.0 s\\
Dynamics Model&&$N$=40, $\delta$=0.02 s&$N$=40, $\delta$=0.05 s&$N$=40, $\delta$=0.1 s&$N$=40, $\delta$=0.15 s&$N$=40, $\delta$=0.2 s&$N$=40, $\delta$=0.25 s&$N$=40, $\delta$=0.3 s\\
\midrule
$\Delta t-$RNN & $\mathcal{O}$ &      0.08$\pm$0.0 &      0.08$\pm$0.0 &      0.08$\pm$0.0 &      0.08$\pm$0.0 &      0.08$\pm$0.0 &      0.08$\pm$0.0 &      0.08$\pm$0.0 \\
     & $\mathcal{R}$ &  (95.85$\pm$0.34) &  (96.88$\pm$0.34) &  (97.18$\pm$0.35) &  (97.37$\pm$0.26) &  (97.32$\pm$0.84) &  (92.75$\pm$3.09) &  (76.16$\pm$8.36) \\
NODE & $\mathcal{O}$ &     0.069$\pm$0.0 &     0.117$\pm$0.0 &     0.165$\pm$0.0 &     0.236$\pm$0.0 &     0.284$\pm$0.0 &    0.361$\pm$0.02 &     0.403$\pm$0.0 \\
     & $\mathcal{R}$ &  (92.56$\pm$0.16) &  (78.24$\pm$1.96) &     (0.0$\pm$0.0) &     (0.0$\pm$0.0) &     (0.0$\pm$0.0) &     (0.0$\pm$0.0) &     (0.0$\pm$0.0) \\
     \midrule
NLC \textbf{(Ours)} & $\mathcal{O}$ &     0.094$\pm$0.0 &     0.094$\pm$0.0 &     0.094$\pm$0.0 &     0.094$\pm$0.0 &     0.094$\pm$0.0 &     0.094$\pm$0.0 &     0.094$\pm$0.0 \\
     & $\mathcal{R}$ &  (98.13$\pm$0.16) &  (99.83$\pm$0.14) &  (99.08$\pm$0.29) &  (98.34$\pm$0.43) &  (97.38$\pm$0.53) &   (96.2$\pm$1.19) &  (92.65$\pm$3.45) \\
\bottomrule
\end{tabular}
}
\label{table:longer_time_horizon_same_compute_cartpole}
\end{table*}

\begin{table*}[!htb]
\caption[]{Numerical results of the Pendulum environment, of the planning time taken $\mathcal{O}$ to generate the next action, and normalized scores $\mathcal{R}$ of the baseline methods with an environment action delay of $\tau=\bar{\Delta}=0.05$s, varying against an increasing time horizon $H$---by increasing the observation interval $\delta$. NLC, maintains a high performing policy at a longer time horizon---whilst using the same amount of \textit{constant} planning time per action $\mathcal{O}$ as a $\Delta t-$RNN.}
\resizebox{\textwidth}{!}{
\begin{tabular}{ccccccccccccc}
\toprule
&&$H$=0.8 s&$H$=2.0 s&$H$=4.0 s&$H$=6.0 s&$H$=8.0 s&$H$=10.0 s&$H$=12.0 s\\
Dynamics Model&&$N$=40, $\delta$=0.02 s&$N$=40, $\delta$=0.05 s&$N$=40, $\delta$=0.1 s&$N$=40, $\delta$=0.15 s&$N$=40, $\delta$=0.2 s&$N$=40, $\delta$=0.25 s&$N$=40, $\delta$=0.3 s\\
\midrule
$\Delta t-$RNN & $\mathcal{O}$ &    0.355$\pm$0.04 &    0.362$\pm$0.03 &    0.369$\pm$0.03 &    0.372$\pm$0.04 &    0.359$\pm$0.04 &    0.364$\pm$0.04 &     0.361$\pm$0.03 \\
     & $\mathcal{R}$ &     (0.0$\pm$0.0) &  (37.11$\pm$3.79) &  (32.67$\pm$5.23) &  (20.28$\pm$5.13) &  (15.46$\pm$3.15) &   (15.07$\pm$4.4) &    (9.12$\pm$2.22) \\
     NODE & $\mathcal{O}$ &      0.84$\pm$0.1 &    1.014$\pm$0.07 &    1.164$\pm$0.13 &    1.364$\pm$0.19 &     1.494$\pm$0.2 &     1.76$\pm$0.21 &     2.007$\pm$0.14 \\
     & $\mathcal{R}$ &     (0.0$\pm$0.0) &     (0.0$\pm$0.0) &     (0.0$\pm$0.0) &     (0.0$\pm$0.0) &     (0.0$\pm$0.0) &     (0.0$\pm$0.0) &      (0.0$\pm$0.0) \\
     \midrule
     \bf NLC \textbf{(Ours)} & $\mathcal{O}$ &    0.493$\pm$0.08 &    0.469$\pm$0.06 &    0.466$\pm$0.05 &    0.466$\pm$0.04 &    0.468$\pm$0.04 &    0.471$\pm$0.04 &     0.462$\pm$0.04 \\
          & $\mathcal{R}$ &  (60.31$\pm$6.39) &  (97.98$\pm$3.21) &  (89.04$\pm$7.04) &  (76.73$\pm$9.15) &  (79.7$\pm$11.59) &  (71.8$\pm$27.86) &  (83.44$\pm$32.02) \\
\bottomrule
\end{tabular}
}
\label{table:longer_time_horizon_same_compute_pendulum}
\end{table*}

\begin{table*}[!htb]
\caption[]{Numerical results of the Acrobot environment, of the planning time taken $\mathcal{O}$ to generate the next action, and normalized scores $\mathcal{R}$ of the baseline methods with an environment action delay of $\tau=\bar{\Delta}=0.05$s, varying against an increasing time horizon $H$---by increasing the observation interval $\delta$. NLC, maintains a high performing policy at a longer time horizon---whilst using the same amount of \textit{constant} planning time per action $\mathcal{O}$ as a $\Delta t-$RNN.}
\resizebox{\textwidth}{!}{
\begin{tabular}{ccccccccccccc}
    \toprule
    &&$H$=0.8 s&$H$=2.0 s&$H$=4.0 s&$H$=6.0 s&$H$=8.0 s&$H$=10.0 s&$H$=12.0 s\\
    Dynamics Model&&$N$=40, $\delta$=0.02 s&$N$=40, $\delta$=0.05 s&$N$=40, $\delta$=0.1 s&$N$=40, $\delta$=0.15 s&$N$=40, $\delta$=0.2 s&$N$=40, $\delta$=0.25 s&$N$=40, $\delta$=0.3 s\\
    \midrule
    $\Delta t-$RNN & $\mathcal{O}$ &    0.387$\pm$0.04 &    0.394$\pm$0.02 &       0.4$\pm$0.03 &   0.393$\pm$0.02 &    0.401$\pm$0.03 &     0.406$\pm$0.03 &     0.407$\pm$0.02 \\
         & $\mathcal{R}$ &     (0.0$\pm$0.0) &  (19.78$\pm$7.95) &    (26.49$\pm$6.0) &  (28.62$\pm$9.1) &  (23.51$\pm$9.02) &  (24.34$\pm$12.93) &  (17.28$\pm$16.53) \\
         NODE & $\mathcal{O}$ &     0.94$\pm$0.17 &    1.058$\pm$0.09 &     1.194$\pm$0.14 &    1.401$\pm$0.2 &    1.529$\pm$0.21 &     1.795$\pm$0.22 &     2.084$\pm$0.19 \\
         & $\mathcal{R}$ &   (25.1$\pm$5.12) &  (26.13$\pm$9.13) &  (11.59$\pm$22.12) &    (0.0$\pm$0.0) &              NA &               NA &               NA \\
         \midrule
         \bf NLC \textbf{(Ours)} & $\mathcal{O}$ &    0.588$\pm$0.05 &    0.574$\pm$0.02 &     0.571$\pm$0.01 &   0.571$\pm$0.01 &    0.569$\pm$0.01 &     0.566$\pm$0.01 &     0.571$\pm$0.01 \\
              & $\mathcal{R}$ &  (85.62$\pm$1.85) &  (99.49$\pm$1.87) &    (89.93$\pm$2.8) &  (78.4$\pm$4.96) &  (57.32$\pm$10.5) &  (37.56$\pm$18.32) &  (20.16$\pm$19.83) \\
    \bottomrule
    \end{tabular}
}
\label{table:longer_time_horizon_same_compute_acrobot}
\end{table*}

\subsection{Can NLC use less compute to plan with the same time horizon?}

\begin{figure}[!htb]
  \centering
\includegraphics[width=\columnwidth]{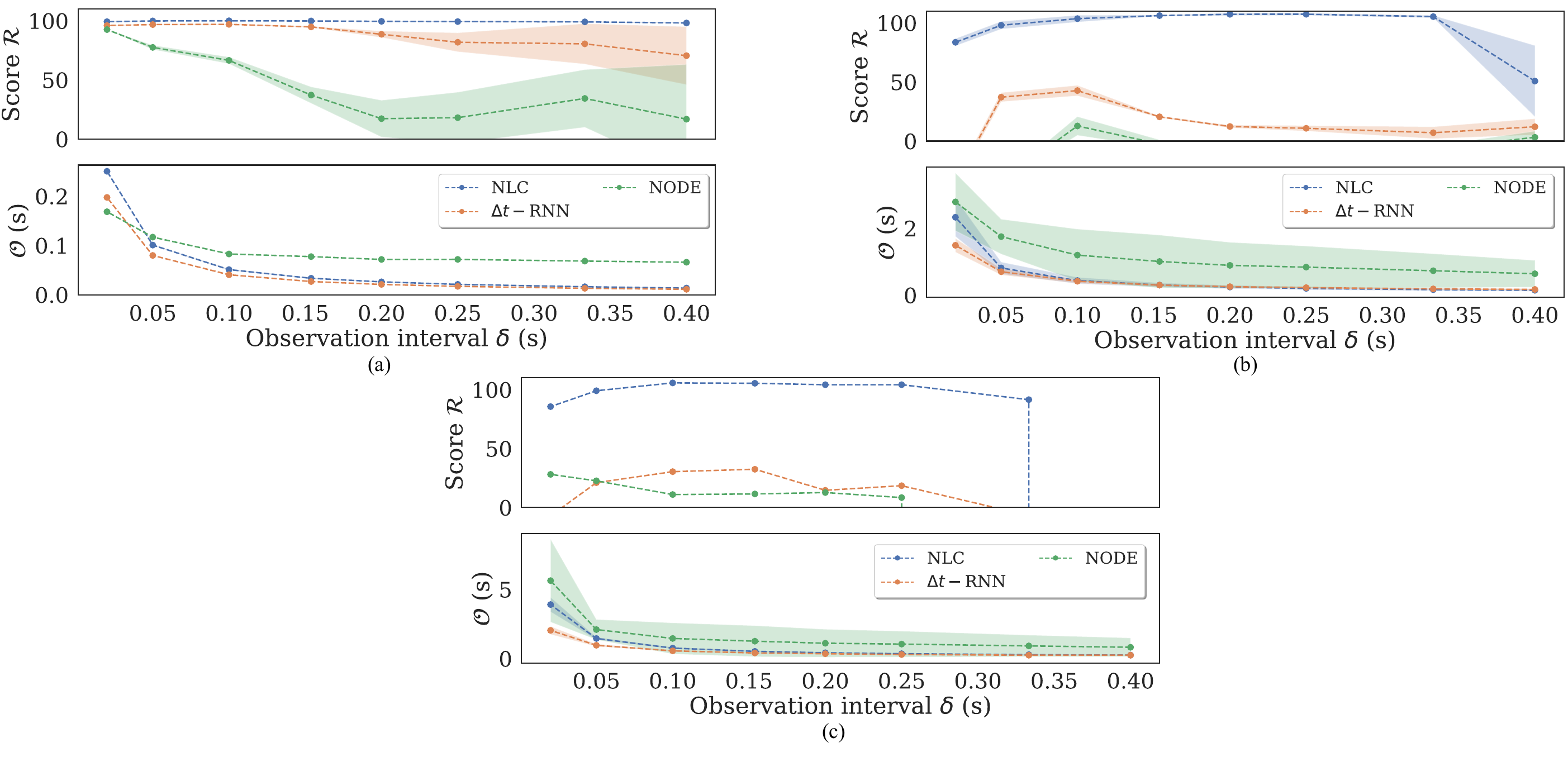}
\caption{Normalized score $\mathcal{R}$ of the baseline methods on the three environments in each sub-figure with an action delay of $\tau=\bar{\Delta}=0.05$s, plotted against an increasing observation interval $\delta$. Specifically in: (a) the Cartpole environment, (b) the Pendulum environment and (c) the Acrobot environment. Here, the time horizon is fixed at $H=2$s, thus increasing the observation interval $\delta$ decreases the number of MPC forward planning steps needed (i.e., $N=\frac{H}{\delta}$). NLC demonstrates that it can still outperform the baselines, achieving a near optimal policy---whilst reducing the planning time taken $\mathcal{O}$ needed to generate the next action.}
\label{Fig:DifferentFreqsAllApp}
\end{figure}

We further investigate an alternative setup in Figure \ref{Fig:DifferentFreqsAllApp}, and keep the time horizon fixed at $H=2$ seconds and increase the observation interval $\delta$---allowing us to reduce $N$ the number of MPC forward planning steps (i.e., $N=\frac{H}{\delta}$). Importantly, this \textit{reduces the planning time} $\mathcal{O}$ needed to generate the next action, enabling a method to use a higher frequency of executing actions to control the dynamics---whilst still planning at the \textit{same fixed time horizon} $H$. NLC is able to still outperform the baselines, achieving a high performing policy---even when using a lesser amount of planning compute per action.
The numeric values for each environment are tabulated in Tables \ref{table:less_compute_to_plan_same_horizon_cartpole}, \ref{table:less_compute_to_plan_same_horizon_pendulum} \& \ref{table:less_compute_to_plan_same_horizon_acrobot}.

\begin{table*}[!htb]
\caption[]{Numerical results of the Cartpole environment, of the planning time taken $\mathcal{O}$ to generate the next action, and normalized scores $\mathcal{R}$ of the baseline methods with an environment action delay of $\tau=\bar{\Delta}=0.05$s, varying against an increasing observation interval $\delta$. Here, the time horizon is fixed at $H=2$s, thus increasing the observation interval $\delta$ decreases the number of MPC forward planning steps needed (i.e., $N=\frac{H}{\delta}$). NLC demonstrates that it can still outperform the baselines, achieving a near optimal policy---whilst reducing the planning time taken $\mathcal{O}$ needed to generate the next action.}
\resizebox{\textwidth}{!}{
\begin{tabular}{ccccccccccccc}
\toprule
&&$H$=2.0 s&$H$=2.0 s&$H$=2.0 s&$H$=2.0 s&$H$=2.0 s&$H$=2.0 s&$H$=2.0 s&$H$=2.0 s\\
Dynamics Model&&$N$=100, $\delta$=0.02 s&$N$=40, $\delta$=0.05 s&$N$=20, $\delta$=0.1 s&$N$=13, $\delta$=0.15 s&$N$=10, $\delta$=0.2 s&$N$=8, $\delta$=0.25 s&$N$=6, $\delta$=0.33 s&$N$=5, $\delta$=0.4 s\\
\midrule
$\Delta t-$RNN & $\mathcal{O}$ &     0.197$\pm$0.0 &      0.08$\pm$0.0 &      0.04$\pm$0.0 &     0.027$\pm$0.0 &      0.021$\pm$0.0 &      0.017$\pm$0.0 &      0.013$\pm$0.0 &      0.011$\pm$0.0 \\
     & $\mathcal{R}$ &   (95.95$\pm$0.4) &  (96.79$\pm$0.25) &  (96.95$\pm$0.27) &  (94.79$\pm$0.49) &   (88.59$\pm$2.68) &   (81.81$\pm$7.97) &  (80.47$\pm$17.05) &  (70.48$\pm$23.23) \\
NODE & $\mathcal{O}$ &     0.168$\pm$0.0 &     0.117$\pm$0.0 &     0.083$\pm$0.0 &     0.077$\pm$0.0 &      0.072$\pm$0.0 &      0.072$\pm$0.0 &      0.068$\pm$0.0 &      0.066$\pm$0.0 \\
     & $\mathcal{R}$ &  (92.57$\pm$0.13) &  (77.49$\pm$1.77) &   (66.49$\pm$2.8) &   (37.18$\pm$6.9) &  (17.23$\pm$15.49) &  (18.08$\pm$21.49) &  (34.32$\pm$24.34) &  (16.81$\pm$38.05) \\
     \midrule
\bf NLC \textbf{(Ours)} & $\mathcal{O}$ &      0.25$\pm$0.0 &     0.101$\pm$0.0 &     0.051$\pm$0.0 &     0.033$\pm$0.0 &      0.026$\pm$0.0 &      0.021$\pm$0.0 &      0.016$\pm$0.0 &      0.013$\pm$0.0 \\
     & $\mathcal{R}$ &   (99.23$\pm$0.3) &  (99.85$\pm$0.13) &  (99.95$\pm$0.12) &  (99.82$\pm$0.15) &   (99.51$\pm$0.18) &   (99.32$\pm$0.27) &   (99.06$\pm$0.44) &   (98.13$\pm$0.16) \\
\bottomrule
\end{tabular}
}
\label{table:less_compute_to_plan_same_horizon_cartpole}
\end{table*}

\begin{table*}[!htb]
\caption[]{Numerical results of the Pendulum environment, of the planning time taken $\mathcal{O}$ to generate the next action, and normalized scores $\mathcal{R}$ of the baseline methods with an environment action delay of $\tau=\bar{\Delta}=0.05$s, varying against an increasing observation interval $\delta$. Here, the time horizon is fixed at $H=2$s, thus increasing the observation interval $\delta$ decreases the number of MPC forward planning steps needed (i.e., $N=\frac{H}{\delta}$). NLC demonstrates that it can still outperform the baselines, achieving a near optimal policy---whilst reducing the planning time taken $\mathcal{O}$ needed to generate the next action.}
\resizebox{\textwidth}{!}{
\begin{tabular}{ccccccccccccc}
\toprule
&&$H$=2.0 s&$H$=2.0 s&$H$=2.0 s&$H$=2.0 s&$H$=2.0 s&$H$=2.0 s&$H$=2.0 s&$H$=2.0 s\\
Dynamics Model&&$N$=100, $\delta$=0.02 s&$N$=40, $\delta$=0.05 s&$N$=20, $\delta$=0.1 s&$N$=13, $\delta$=0.15 s&$N$=10, $\delta$=0.2 s&$N$=8, $\delta$=0.25 s&$N$=6, $\delta$=0.33 s&$N$=5, $\delta$=0.4 s\\
\midrule
$\Delta t-$RNN & $\mathcal{O}$ &      1.49$\pm$0.2 &     0.699$\pm$0.1 &     0.414$\pm$0.06 &     0.299$\pm$0.04 &     0.251$\pm$0.04 &     0.223$\pm$0.03 &     0.186$\pm$0.03 &     0.169$\pm$0.02 \\
     & $\mathcal{R}$ &     (0.0$\pm$0.0) &  (37.11$\pm$3.79) &    (42.7$\pm$4.37) &    (20.51$\pm$0.6) &   (12.32$\pm$1.09) &   (10.81$\pm$2.27) &    (7.07$\pm$4.99) &   (12.12$\pm$6.86) \\
     NODE & $\mathcal{O}$ &    2.788$\pm$0.86 &    1.749$\pm$0.52 &     1.196$\pm$0.78 &     1.005$\pm$0.79 &     0.889$\pm$0.69 &     0.836$\pm$0.63 &     0.728$\pm$0.51 &      0.638$\pm$0.4 \\
     & $\mathcal{R}$ &     (0.0$\pm$0.0) &     (0.0$\pm$0.0) &   (12.78$\pm$7.86) &      (0.0$\pm$0.0) &      (0.0$\pm$0.0) &      (0.0$\pm$0.0) &      (0.0$\pm$0.0) &    (3.22$\pm$5.19) \\
     \midrule
     \bf NLC \textbf{(Ours)} & $\mathcal{O}$ &    2.329$\pm$0.57 &    0.813$\pm$0.18 &      0.434$\pm$0.1 &     0.301$\pm$0.06 &     0.241$\pm$0.05 &       0.2$\pm$0.04 &     0.163$\pm$0.03 &     0.145$\pm$0.03 \\
          & $\mathcal{R}$ &  (83.55$\pm$3.02) &  (97.98$\pm$3.21) &  (103.55$\pm$2.84) &  (106.13$\pm$0.41) &  (107.22$\pm$1.03) &  (107.28$\pm$0.93) &  (105.27$\pm$0.98) &  (50.73$\pm$30.43) \\
\bottomrule
\end{tabular}
}
\label{table:less_compute_to_plan_same_horizon_pendulum}
\end{table*}

\begin{table*}[!htb]
\caption[]{Numerical results of the Acrobot environment, of the planning time taken $\mathcal{O}$ to generate the next action, and normalized scores $\mathcal{R}$ of the baseline methods with an environment action delay of $\tau=\bar{\Delta}=0.05$s, varying against an increasing observation interval $\delta$. Here, the time horizon is fixed at $H=2$s, thus increasing the observation interval $\delta$ decreases the number of MPC forward planning steps needed (i.e., $N=\frac{H}{\delta}$). NLC demonstrates that it can still outperform the baselines, achieving a near optimal policy---whilst reducing the planning time taken $\mathcal{O}$ needed to generate the next action.}
\resizebox{\textwidth}{!}{
\begin{tabular}{ccccccccccccc}
    \toprule
    &&$H$=2.0 s&$H$=2.0 s&$H$=2.0 s&$H$=2.0 s&$H$=2.0 s&$H$=2.0 s&$H$=2.0 s&$H$=2.0 s\\
    Dynamics Model&&$N$=100, $\delta$=0.02 s&$N$=40, $\delta$=0.05 s&$N$=20, $\delta$=0.1 s&$N$=13, $\delta$=0.15 s&$N$=10, $\delta$=0.2 s&$N$=8, $\delta$=0.25 s&$N$=6, $\delta$=0.33 s&$N$=5, $\delta$=0.4 s\\
    \midrule
    $\Delta t-$RNN & $\mathcal{O}$ &    2.037$\pm$0.29 &     0.96$\pm$0.08 &     0.558$\pm$0.06 &     0.405$\pm$0.04 &     0.348$\pm$0.03 &     0.292$\pm$0.03 &     0.243$\pm$0.02 &  0.243$\pm$0.02 \\
         & $\mathcal{R}$ &     (0.0$\pm$0.0) &  (21.04$\pm$7.06) &    (30.4$\pm$9.03) &   (32.44$\pm$9.69) &    (14.59$\pm$5.5) &    (18.49$\pm$7.6) &      (0.0$\pm$0.0) &   (0.0$\pm$0.0) \\
         NODE & $\mathcal{O}$ &    5.631$\pm$2.99 &    2.095$\pm$0.73 &     1.455$\pm$1.12 &     1.256$\pm$1.12 &     1.108$\pm$1.01 &     1.045$\pm$0.93 &     0.915$\pm$0.78 &  0.815$\pm$0.67 \\
         & $\mathcal{R}$ &  (28.09$\pm$6.18) &  (22.63$\pm$7.97) &  (10.96$\pm$22.24) &  (11.48$\pm$24.95) &  (12.67$\pm$17.63) &   (8.41$\pm$16.82) &      (0.0$\pm$0.0) &   (0.0$\pm$0.0) \\
         \midrule
         \bf NLC \textbf{(Ours)} & $\mathcal{O}$ &    3.902$\pm$0.52 &    1.452$\pm$0.09 &     0.755$\pm$0.06 &     0.521$\pm$0.05 &     0.419$\pm$0.04 &     0.349$\pm$0.04 &     0.283$\pm$0.03 &  0.251$\pm$0.03 \\
              & $\mathcal{R}$ &   (85.6$\pm$4.98) &  (99.05$\pm$2.06) &  (105.71$\pm$0.92) &  (105.39$\pm$1.24) &   (104.15$\pm$0.9) &  (104.18$\pm$2.61) &  (91.49$\pm$20.89) &   (0.0$\pm$0.0) \\
    \bottomrule
    \end{tabular}
}
\label{table:less_compute_to_plan_same_horizon_acrobot}
\end{table*}

\subsection{Can NLC learn from few samples?}
\label{sampleeffieicny}

\begin{figure}[!htb]
\begin{center}
    \centering
      \includegraphics[width=\textwidth]{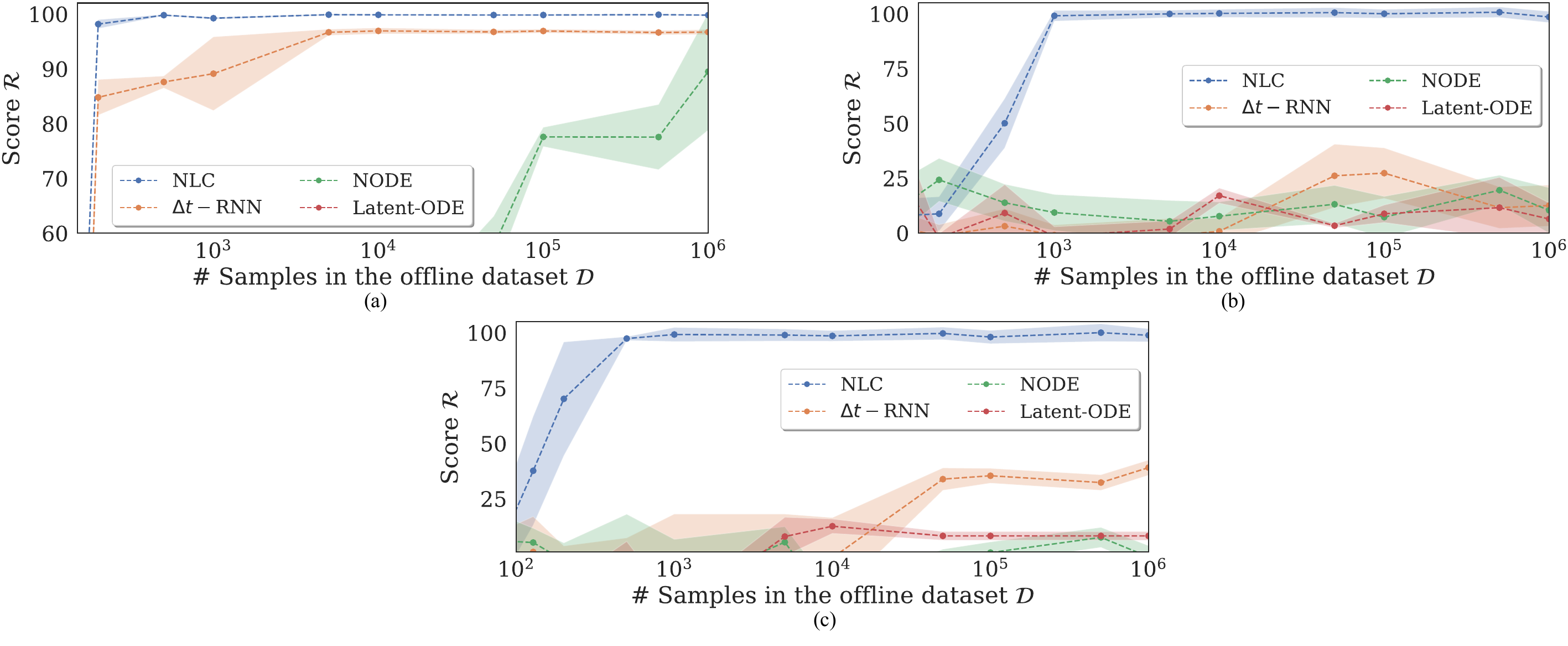}
     \end{center}
  \caption{Normalized score $\mathcal{R}$ of the baseline methods on the three environments in each sub-figure with an action delay of $\tau=\bar{\Delta}=0.05$s, plotted against the number of samples in the irregularly-sampled offline dataset used to train the dynamics model of each method. Specifically, in: (a) the Cartpole environment, (b) the Pendulum environment and (c) the Acrobot environment. NLC can maintain a high performing policy on the Cartpole and Pendulum environments---even from the challenging case of only learning a dynamics model from 200 samples from an irregularly-sampled in time offline dataset $\mathcal{D}$.}
    \label{model_sample_less_plots}
\end{figure}

We observe in Figure \ref{model_sample_less_plots} that NLC can still learn a suitable dynamics model and perform well across the environments with a delay of $\tau=\bar{\Delta}=0.05$ seconds, when trained with an offline irregularly-sampled in time dataset that contains a limited number of samples. Specifically, it is able to learn with only 200 random samples on the Cartpole and Pendulum environments---which corresponds 10 seconds of interaction time of a noisy expert (expert with random action noise) agent from the delayed environment. Also, on the Acrobot environment, a more challenging environment it is able to learn a sufficient dynamics model from only 1,000 random samples.

\subsection{Can NLC incorporate state-based constraints?}

\begin{figure}[!htb]
  \centering
\includegraphics[width=\columnwidth]{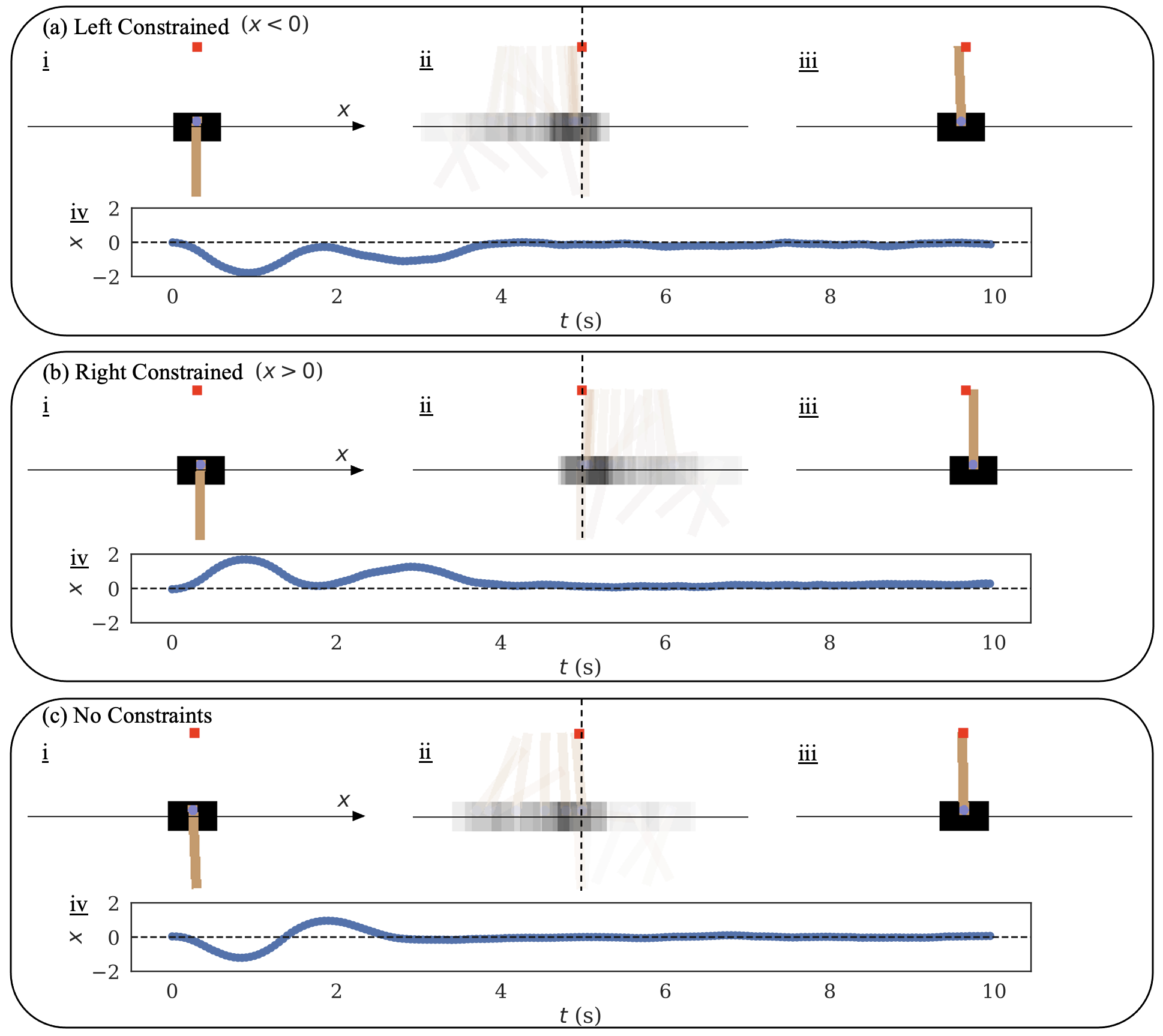}
\caption{Screen shots of the Cartpole environment, with an action delay of $\tau=\bar{\Delta}=0.05$s using the NLC dynamics model. We see additional new state-based constraints given at run-time, and the MPC planner is able to incorporate these such that the action trajectories generated and executed satisfy the state constraint, here on the horizontal $x$ position of the cart. Specifically, in sub-figures we observe: (a) a \quotes{left constrained} ($x<0$) version, (b) a \quotes{right constrained} ($x>0$) version and in (c) a standard version with no state constraints. Specifically in each sub-figure, we have further sub-figures where: (i) shows the starting screenshot, (ii) a visualization of the trajectory followed (with past screenshots superimposed, where the more faded image is oldest), (iii) the final state reached and in (iv) the state $x$ trajectory plotted over the entire episode's length of 10 seconds. Therefore, using a MPC planner we can easily incorporate new additional unseen state constraints at run-time, whereas using a learnt policy (q-learner) would be unable to do this and would have to re-train a new policy for each new state constraint.}
\label{Fig:StateBasedConstraint}
\end{figure}

We show that using a MPC planner, we can easily incorporate a new state-based constraint at test time (run-time) in Figure \ref{Fig:StateBasedConstraint}. Here, using the Cartpole environment we add an additional constraint on the horizontal $x$ position of the cart. Specifically, we illustrate that a \quotes{left constrained} ($x<0$) version at run-time can still be solved by our NLC method and similarly for a \quotes{right constrained} ($x>0$) version can be solved---this is made possible as the planner only generates feasible trajectories where these additional state-based constraints are satisfied, which is further illustrated in Figure \ref{Fig:StateBasedConstraint}. Therefore, using a MPC planner we can easily incorporate new additional unseen state constraints at run-time, whereas using a learnt policy (q-learner) would be unable to do this and would have to re-train a new policy for each new state constraint. We note that this benefit of using a MPC planner to incorporate additional state-based constraints has been shown by others \citep{lutter2021learning}.

\section{ADDITIONAL EXPERIMENTS}
\label{additonalexperiments}

In this section we seek to gain further insight into how the baseline dynamics models compare when trained using the same number of fixed training epochs on each dataset, and an ablation by training the dynamics models on regularly-sampled offline datasets.

\subsection{Training dynamics models for a fixed number of epochs}
\label{Trainingdynamicsmodelsforafixednumberofepochs}

Here we changed the training setup to train each of the dynamics models for the same number of 10 epochs, with the same batch size, that of 1 (As NODE and Latent-ODE are only able to support a batch size of 1 using our offline dataset) on the irregularly sampled offline dataset consisting of only 10,000 random state-action samples. The benchmark methods against each environment, which consists of a continuous-time environment with a specific delay---with normalized scores $\mathcal{R}$ are tabulated in Table \ref{table:main_normalized_scores_all_results_training_fixed_epochs}.
We observe that the results are consistent and similar to the original results reported in Table 2 in the paper---where each dynamics model received the same number of training iterations, and this is less training than our original results tabulated in Table 2 in the paper. Given less training iterations, NLC is still able to outperform the baseline dynamics models.

\begin{table*}[!htb]
\caption[]{Normalized scores $\mathcal{R}$ of the offline model-based agents, where the irregularly-sampled (P1) offline dataset consists of an action delay (P2) of $\{0,1,2,3\}$ multiples of the environment's observation interval time step $\bar{\Delta} = 0.05$ seconds. Averaged over 20 random seeds, with $\pm$ standard deviations.
Scores are un-discounted cumulative rewards normalized to be between 0 and 100, where 0 corresponds to the Random agent and 100 corresponds to the expert with the \textit{known} world model (oracle+MPC). Negative normalized scores, i.e., worse than random are set to zero. Specifically, we trained each dynamics model for the same number of 10 epochs, with a batch size of 1 from an offline dataset consisting of only 10,000 random state-action samples---we note that each dynamics model received the same number of training iterations, and this is less training than our original results tabulated in Table 2 in the paper. Given less training iterations, NLC is still able to outperform the baseline dynamics models.}
\resizebox{\textwidth}{!}{
\begin{tabular}{@{}l|ccc|ccc|ccc|ccc}
\toprule
                                &  \multicolumn{3}{c|}{Action Delay~$\tau=0$}     & \multicolumn{3}{c|}{Action Delay~$\tau=\bar{\Delta}$}      & \multicolumn{3}{c|}{Action Delay~$\tau=2\bar{\Delta}$}      &  \multicolumn{3}{c}{Action Delay~$\tau=3\bar{\Delta}$}               \\
        Dynamics Model                  &     Cartpole & Pendulum & Acrobot & Cartpole & Pendulum & Acrobot                   & Cartpole & Pendulum & Acrobot                     & Cartpole & Pendulum & Acrobot \\        
\midrule
Random              &       0.0$\pm$0.0 &        0.0$\pm$0.0 &      0.0$\pm$0.0 &        0.0$\pm$0.0 &      0.0$\pm$0.0 &      0.0$\pm$0.0 &     0.0$\pm$0.0 &      0.0$\pm$0.0 &       0.0$\pm$0.0 &        0.0$\pm$0.0 &      0.0$\pm$0.0 &       0.0$\pm$0.0 \\
Oracle              &     100.0$\pm$0.04 &     100.0$\pm$3.59 &    100.0$\pm$2.61 &     100.0$\pm$0.02 &     100.0$\pm$3.15 &     100.0$\pm$1.8 &     100.0$\pm$0.03 &     100.0$\pm$2.42 &    100.0$\pm$1.64 &     100.0$\pm$0.04 &      100.0$\pm$2.6 &    100.0$\pm$1.42 \\
$\Delta t-$RNN      &     97.13$\pm$0.27 &      19.86$\pm$5.2 &     30.9$\pm$8.19 &     98.59$\pm$0.13 &     13.02$\pm$11.0 &    25.16$\pm$9.86 &     98.02$\pm$0.17 &     19.64$\pm$6.21 &    33.67$\pm$7.71 &     96.89$\pm$0.29 &      13.18$\pm$4.4 &    26.27$\pm$6.03 \\
Latent-ODE          &        0.0$\pm$0.0 &        0.0$\pm$0.0 &     5.99$\pm$9.64 &        0.0$\pm$0.0 &     5.26$\pm$11.63 &    6.45$\pm$13.78 &        0.0$\pm$0.0 &     2.86$\pm$14.97 &     7.97$\pm$9.31 &    35.81$\pm$55.95 &     5.29$\pm$12.05 &    12.22$\pm$6.08 \\
NODE                &     95.63$\pm$0.19 &        0.0$\pm$0.0 &   16.04$\pm$10.68 &      93.43$\pm$4.1 &        0.0$\pm$0.0 &      1.7$\pm$6.73 &        0.0$\pm$0.0 &      6.19$\pm$3.16 &     5.5$\pm$10.52 &     95.49$\pm$0.12 &       4.4$\pm$5.13 &    23.98$\pm$8.18 \\
\midrule
\bf NLC \textbf{(Ours)} &     \textbf{99.54$\pm$0.12} &     \textbf{92.65$\pm$4.43} &    \textbf{57.1$\pm$22.72} &     \textbf{99.39$\pm$0.02} &     \textbf{95.78$\pm$2.35} &    \textbf{66.55$\pm$6.59} &     \textbf{99.71$\pm$0.07} &     \textbf{99.01$\pm$4.13} &   \textbf{57.73$\pm$20.73} &     \textbf{97.16$\pm$0.24} &     \textbf{91.56$\pm$4.73} &   \textbf{80.48$\pm$14.03} \\
\bottomrule
\end{tabular}
}
\label{table:main_normalized_scores_all_results_training_fixed_epochs}
\end{table*}

\subsection{Ablation by training dynamics models on regularly-sampled offline datasets}
\label{Ablationbytrainingdynamicsmodelsonregularlysampledofflinedatasets}

We further investigate how the baseline models perform by training on regularly-sampled offline datasets, i.e., datasets that are collected with a noisy agent that observes the next observation $\bm{x}(t+\Delta_i)$ with the same discrete time observation interval $\Delta_i=\Delta_j$. These results are tabulated in Table \ref{table:main_normalized_scores_all_results_table_regularly-sampled-data}. We observe a similar pattern, where NLC is still able to learn a \textit{good} dynamics model and outperform the baselines.

\begin{table*}[!htb]
\caption[]{Normalized scores $\mathcal{R}$ of the offline model-based agents, where we use a regularly-sampled offline dataset---that consists of an action delay (P2) of $\{0,1,2,3\}$ multiples of the environments observation interval time step $\bar{\Delta} = 0.05$ seconds. Averaged over 20 random seeds, with $\pm$ standard deviations.
Scores are un-discounted cumulative rewards normalized to be between 0 and 100, where 0 corresponds to the Random agent and 100 corresponds to the expert with the \textit{known} world model (oracle+MPC). Negative normalized scores, i.e., worse than random are set to zero.}
\resizebox{\textwidth}{!}{
\begin{tabular}{@{}l|ccc|ccc|ccc|ccc}
\toprule
                                &  \multicolumn{3}{c|}{Action Delay~$\tau=0$}     & \multicolumn{3}{c|}{Action Delay~$\tau=\bar{\Delta}$}      & \multicolumn{3}{c|}{Action Delay~$\tau=2\bar{\Delta}$}      &  \multicolumn{3}{c}{Action Delay~$\tau=3\bar{\Delta}$}               \\
        Dynamics Model                  &     Cartpole & Pendulum & Acrobot & Cartpole & Pendulum & Acrobot                   & Cartpole & Pendulum & Acrobot                     & Cartpole & Pendulum & Acrobot \\        
\midrule
Random              &      0.0$\pm$0.0 &      0.0$\pm$0.0 &       0.0$\pm$0.0 &     0.0$\pm$0.0 &      0.0$\pm$0.0 &      0.0$\pm$0.0 &     0.0$\pm$0.0 &        0.0$\pm$0.0 &       0.0$\pm$0.0 &     0.0$\pm$0.0 &       0.0$\pm$0.0 &      0.0$\pm$0.0 \\
Oracle              &     100.0$\pm$0.06 &     100.0$\pm$3.59 &    100.0$\pm$1.98 &     100.0$\pm$0.03 &     100.0$\pm$3.05 &    100.0$\pm$1.18 &      100.0$\pm$0.1 &     100.0$\pm$2.57 &    100.0$\pm$2.27 &     100.0$\pm$0.03 &      100.0$\pm$2.7 &     100.0$\pm$1.7 \\
$\Delta t-$RNN      &     97.17$\pm$0.42 &        0.0$\pm$0.0 &     3.07$\pm$6.09 &     97.18$\pm$0.32 &        0.0$\pm$0.0 &     7.03$\pm$7.53 &     96.54$\pm$0.35 &        0.0$\pm$0.0 &      5.39$\pm$6.0 &     98.01$\pm$0.31 &        0.0$\pm$0.0 &    10.47$\pm$7.09 \\
Latent-ODE          &        0.0$\pm$0.0 &        0.0$\pm$0.0 &    12.26$\pm$9.47 &        0.0$\pm$0.0 &        0.0$\pm$0.0 &    3.34$\pm$25.49 &        0.0$\pm$0.0 &        0.0$\pm$0.0 &   12.81$\pm$14.84 &        0.0$\pm$0.0 &        0.0$\pm$0.0 &    19.47$\pm$10.4 \\
NODE                &        0.0$\pm$0.0 &      49.8$\pm$7.27 &   27.18$\pm$10.36 &        0.0$\pm$0.0 &     39.12$\pm$5.54 &   32.06$\pm$10.84 &    54.19$\pm$51.77 &        0.0$\pm$0.0 &   24.31$\pm$10.38 &        0.0$\pm$0.0 &        0.0$\pm$0.0 &   25.49$\pm$11.89 \\
\midrule
\bf NLC \textbf{(Ours)} &    \textbf{100.01$\pm$0.04} &     \textbf{99.95$\pm$3.74} &    \textbf{99.39$\pm$2.31} &      \textbf{99.97$\pm$0.1} &     \textbf{101.14$\pm$3.7} &   \textbf{101.38$\pm$2.14} &    \textbf{100.03$\pm$0.05} &     \textbf{99.45$\pm$2.98} &    \textbf{99.74$\pm$1.79} &     \textbf{99.93$\pm$0.08} &     \textbf{99.81$\pm$3.17} &   \textbf{100.11$\pm$1.92} \\
\bottomrule
\end{tabular}
}
\label{table:main_normalized_scores_all_results_table_regularly-sampled-data}
\end{table*}

\subsection{Environments with observational noise}

\begin{figure}[!htb]
  \begin{center}
      \centering
        \includegraphics[width=0.5\textwidth]{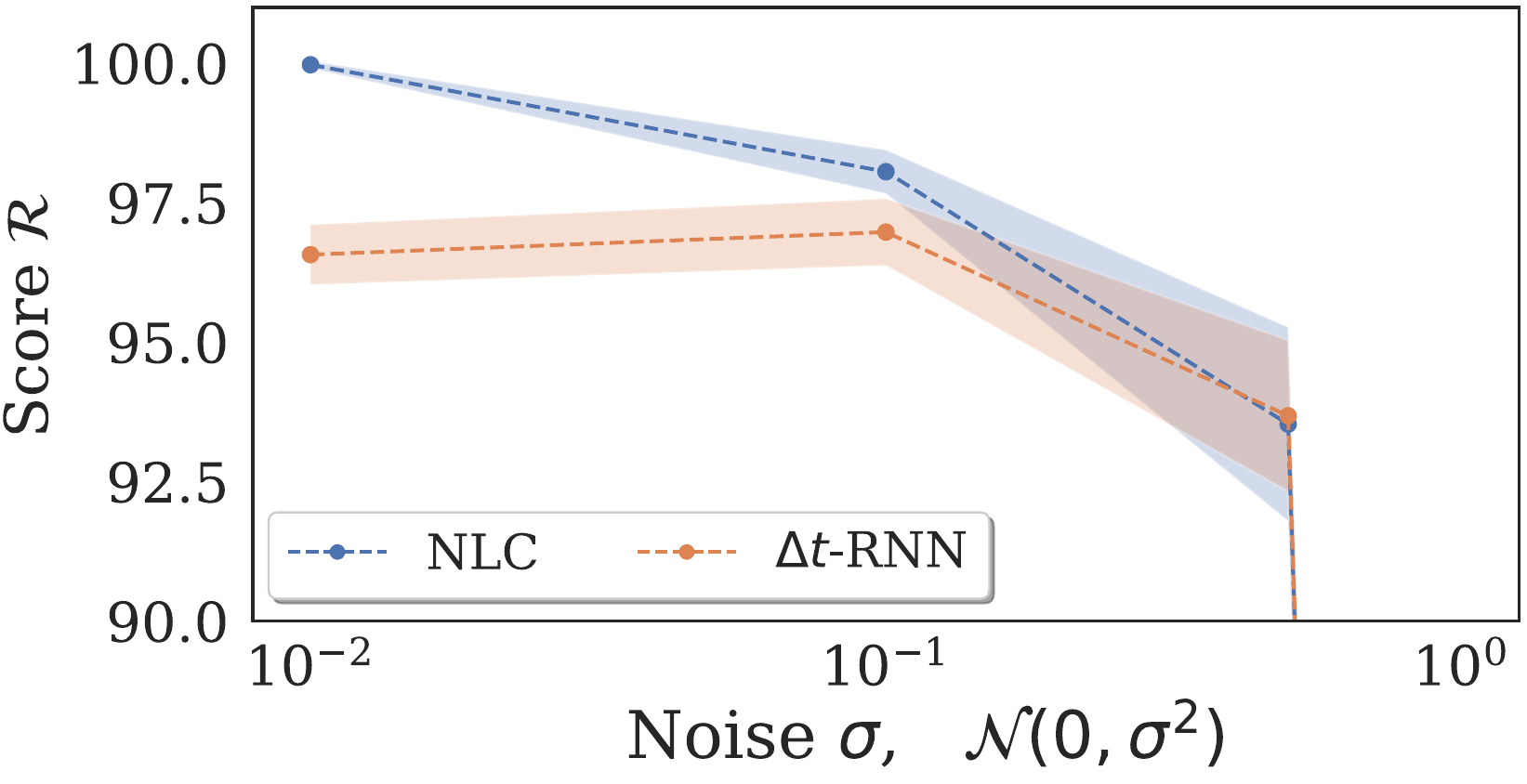}
       \end{center}
    \caption{Normalized score $\mathcal{R}$ of the baseline methods on the Cartpole environment with an action delay of $\tau=\bar{\Delta}=0.05$s. Here adding observation noise, specifically, perturbing the state with Gaussian noise $\mathcal{N}(0,\sigma^2)$, plotted against noise $\sigma$. NLC can maintain a high performing policy on the Cartpole with noise---however degrades like the other closest performing policy with increasing noise, as there becomes less signal to noise in the enviroment observations.}
      \label{fig:increasing_noise_cartpole}
  \end{figure}

We performed an additional experiment of adding observation noise, specifically, perturbing the state with Gaussian noise $\mathcal{N}(0,0.01^2)$. We observe in Table \ref{table:additional_observation_noise} that NLC is still performant under this observation noise. Here, the noisy expert datasets are collected with this observation noise, and the same observation noise is used at run-time evaluation for each method. We also see this same effect for increasing noise, in Figure \ref{fig:increasing_noise_cartpole}.

\begin{table*}[!htb]
\centering
\caption[]{Normalized scores $\mathcal{R}$ of the offline model-based agents, where the irregularly-sampled (P1) offline dataset consists of an action delay (P2) of $\{0,1,2,3\}$ multiples of the environments observation interval time step $\bar{\Delta} = 0.05$ seconds. Averaged over 20 random seeds, with $\pm$ standard deviations. Here we add observation noise---whereby the state is perturbed with Gaussian noise $\mathcal{N}(0,0.01^2)$.
Scores are un-discounted cumulative rewards normalized to be between 0 and 100, where 0 corresponds to the Random agent and 100 corresponds to the expert with the \textit{known} world model (Oracle+MPC). Negative normalized scores, i.e., worse than random are set to zero.}
\begin{tabular}{@{}l|cccc}
\toprule
                                &  Action Delay~$\tau=0$     & Action Delay~$\tau=\bar{\Delta}$     & Action Delay~$\tau=2\bar{\Delta}$      &  Action Delay~$\tau=3\bar{\Delta}$               \\
    Dynamics Model                  &     Cartpole &  Cartpole & Cartpole & Cartpole  \\
\midrule
Random                  &     0.0$\pm$0.0 &        0.0$\pm$0.0 &        0.0$\pm$0.0 &        0.0$\pm$0.0 \\
Oracle              &     100.0$\pm$0.04 &     100.0$\pm$0.02 &     100.0$\pm$0.02 &     100.0$\pm$0.02 \\
$\Delta t-$RNN      &     98.33$\pm$0.26 &     97.96$\pm$0.24 &      97.8$\pm$0.25 &     97.99$\pm$0.27 \\
Latent-ODE          &        0.0$\pm$0.0 &        0.0$\pm$0.0 &        0.0$\pm$0.0 &        0.0$\pm$0.0 \\
NODE                &        0.0$\pm$0.0 &        0.0$\pm$0.0 &        0.0$\pm$0.0 &     9.86$\pm$56.04 \\
\midrule
\bf NLC \textbf{(Ours)} &     \textbf{99.99$\pm$0.03} &     \textbf{99.97$\pm$0.04} &     \textbf{99.91$\pm$0.08} &      \textbf{99.9$\pm$0.03} \\
\bottomrule
\end{tabular}
\label{table:additional_observation_noise}
\end{table*}

\subsection{Environment with friction}

\begin{figure}[!htb]
  \begin{center}
      \centering
        \includegraphics[width=0.5\textwidth]{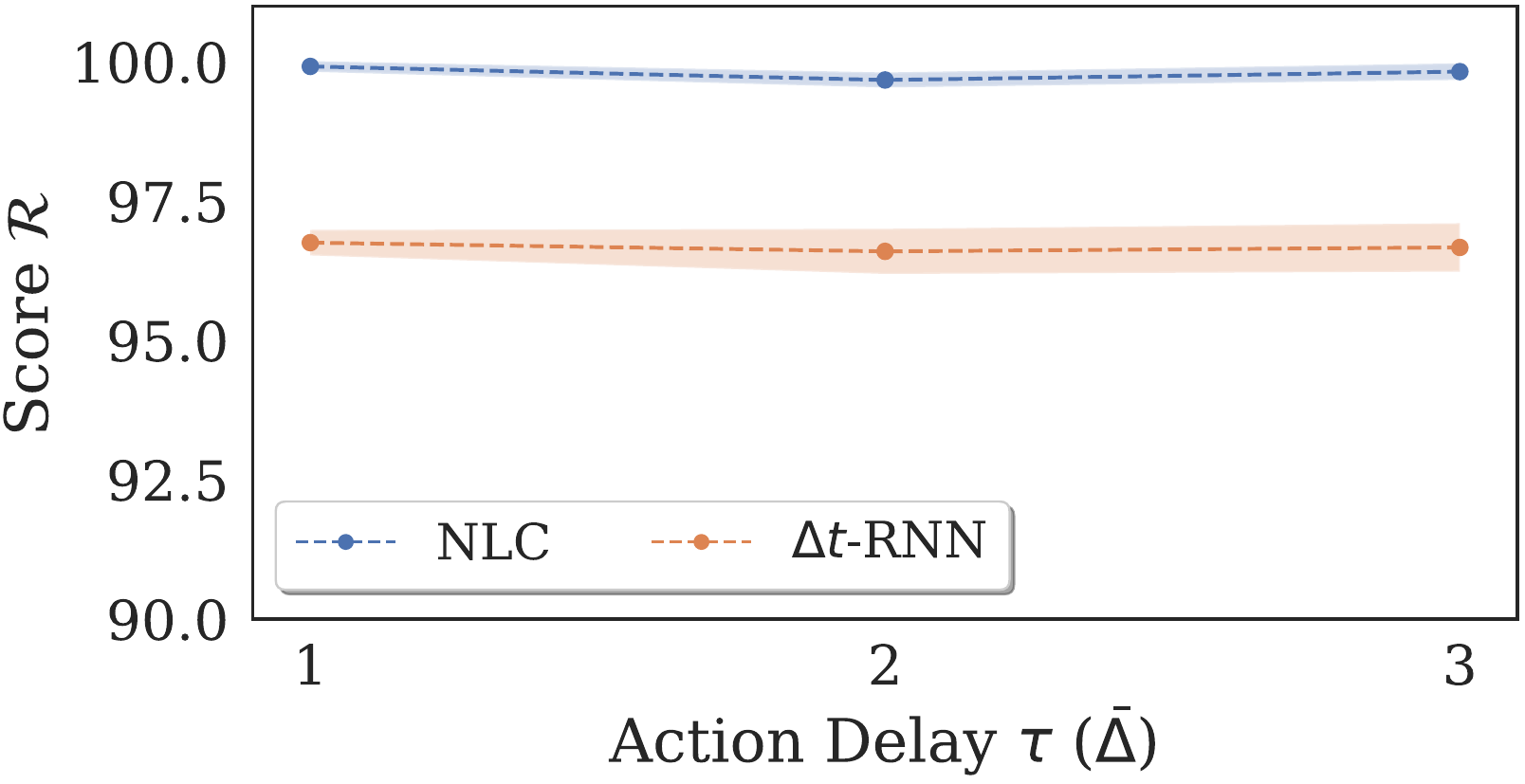}
       \end{center}
    \caption{Normalized score $\mathcal{R}$ of the baseline methods on the Cartpole environment against an action delay of $\{0,1,2,3\}$ multiples of the environments observation interval time step $\bar{\Delta} = 0.05$ seconds. Here the Cartpole environment has friction added to it. NLC can maintain a high performing policy on the Cartpole with friction.}
      \label{fig:cartpole_with_friction}
  \end{figure}

We also performed an additional experiment of adding friction to the environment. We observe in Figure \ref{fig:cartpole_with_friction} that NLC is still performant under this friction for the Cartpole environment. Here, the expert datasets are collected with this friction, and the same friction is used at run-time evaluation for each method. Specifically, we used a fiction coefficient of 5e-4 for the cart and a friction coefficient of 2e-6 for the pole.

\end{document}